\documentclass[format=acmsmall, review=false, screen=true]{acmart}
\usepackage{booktabs} 
\usepackage[ruled]{algorithm2e} 
\usepackage{graphicx}
\usepackage{psfrag}
\usepackage{subfigure}
\usepackage{multirow}
\usepackage{array}
\usepackage{tabu}
\usepackage{color,xcolor}
\usepackage{algorithmic} 

\usepackage{amsmath,amsfonts,amssymb,amsbsy,bm,paralist,ifthen}

\SetAlFnt{\small}
\SetAlCapFnt{\small}
\SetAlCapNameFnt{\small}
\SetAlCapHSkip{0pt}
\IncMargin{-\parindent}

\setcopyright{acmlicensed}

\begin{document}
\title{HERA: Partial Label Learning by Combining Heterogeneous Loss with Sparse and Low-Rank Regularization}

\author{Gengyu Lyu}
\orcid{1234-5678-9012-3456}
\affiliation{%
  \institution{Beijing Jiaotong University}
  \streetaddress{No.3 Shangyuan Cun}
  \city{Haidian District}
   \state{Beijing}
  \postcode{100044}
  \country{China}}
\email{18112030@bjtu.edu.cn}

\author{Songhe Feng}
\affiliation{%
  \institution{Beijing Jiaotong University}
\state{Beijing}
  \postcode{100044}
  \country{China}}
\email{shfeng@bjtu.edu.cn}

\author{Yi Jin *}
\affiliation{%
 \institution{Beijing Jiaotong University}
\state{Beijing}
  \postcode{100044}
  \country{China}}
\email{yjin@bjtu.edu.cn}
\thanks{*Corresponding author: yjin@bjtu.edu.cn}

\author{Guojun Dai}
\affiliation{%
 \institution{Hangzhou Dianzi University}
 \city{Hangzhou}
 \state{Zhejiang}
 \country{China}}
\email{daigj@hdu.edu.cn}

\author{Congyan Lang}
\affiliation{%
 \institution{Beijing Jiaotong University}
 \streetaddress{No.3 Shangyuan Cun}
\city{Haidian District}
 \state{Beijing}
 \country{China}}
\email{cylang@bjtu.edu.cn}

\author{Yidong Li}
\affiliation{%
 \institution{Beijing Jiaotong University}
 \streetaddress{No.3 Shangyuan Cun}
\city{Haidian District}
 \state{Beijing}
 \country{China}}
\email{ydli@bjtu.edu.cn}

\begin{abstract}
Partial Label Learning (PLL) aims to learn from the data where each training instance is associated with a set of candidate labels, among which only one is correct. Most existing methods deal with such problem by either treating each candidate label equally or identifying the ground-truth label iteratively. In this paper, we propose a novel PLL approach called \textbf{HERA}, which simultaneously incorporates the \emph{\textbf{H}eterog\textbf{E}neous Loss} and the \emph{Spa\textbf{R}se and Low-r\textbf{A}nk} procedure to estimate the labeling confidence for each instance while training the model. Specifically, the heterogeneous loss integrates the strengths of both the pairwise ranking loss and the pointwise reconstruction loss to provide informative label ranking and reconstruction information for label identification, while the embedded sparse and low-rank scheme constrains the sparsity of ground-truth label matrix and the low rank of noise label matrix to explore the global label relevance among the whole training data for improving the learning model. Extensive experiments on both artificial and real-world data sets demonstrate that our method can achieve superior or comparable performance against the state-of-the-art methods.
\end{abstract}

%
%



\maketitle

\renewcommand{\shortauthors}{G. Lyu et al.}

\section{Introduction}
As a weakly-supervised machine learning framework, partial label learning \footnote{In some literatures, partial label learning is also called as superset label learning \cite{Liu:ICML2014}, soft label learning \cite{Oukhellou2009Learning} or ambiguous label learning \cite{Chen:IEEET2014}.} learns from the ambiguous data where the ground-truth label is concealed in its corresponding candidate label set and does not directly accessible to the learning algorithm \cite{Cour:lfpl-JMLR2011} \cite{wang2018towards}  \cite{wu2019disambiguation} \cite{zhang2016partial}.

\begin{figure}[H]
\centering
\includegraphics[width = 3in,height=1.5in]{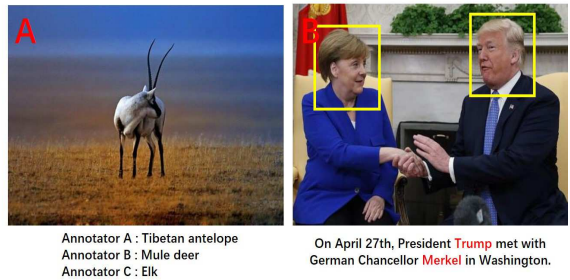}
\vspace{-1mm}
\caption{Exemplar applications of partial-label learning. (A) The candidate annotations are provided by the web users, while the actual correspondence between image and its ground-truth annotation is unknown. (B) The candidate names are extracted from the corresponding text description, while the actual correspondence between each face and its ground-truth name is not available.}
\label{fig00}
\end{figure}

In recent years, owing to its excellent performance on learning from the data with partial labels, PLL has been widely used in many real world scenarios. For instance, in online object annotation (Figure \ref{fig00} A), given the object annotations from varying users, one can treat the objects as instances and annotations as candidate labels, while the correct correspondence between instances and ground-truth labels are unknown \cite{Liu:acmmmfsll-NIPS2012} \cite{Yu2017Geometric}. In automatic face naming (Figure \ref{fig00} B), given multi-figure images and the corresponding text descriptions, each face detected from images can be regarded as an instance and names extracted from the text description can be regarded as candidate labels, similarly the actual correspondence between faces and names are also not available \cite{chen2018learning}. Partial label learning techniques can precisely identify the actual instance-label correspondence and make prediction for unseen examples. In addition, partial label learning has also widely used in other real-world applications, including web mining \cite{Feng:IJCAI2019} \cite{Luo:lfcls-NIPS2010}, multimedia content analysis \cite{Chen2015Matrix} \cite{Cour:lfpl-JMLR2011} \cite{wang2019adaptive} \cite{Zeng:lbaali-CVPR2013}, facial age estimation \cite{feng2019partial} \cite{Wang:IJCAI2019}, Eco informatics \cite{Zhou2016Partial}, etc.

To accomplish the task of learning from partial label data, an intuitive strategy is disambiguation, i.e. trying to identify the ground-truth label from its corresponding candidate labels. Following such strategy, existing PLL learning framework can be roughly grouped into two categories: average-based framework and identification-based framework. For the average-based framework, each candidate label is assumed to have equal contribution to the learning model and the ground-truth label is obtained by averaging the outputs from all candidate labels \cite{Chen2017A} \cite{Cour:lfpl-JMLR2011} \cite{Huller:LNCS2005} \cite{Zhang:IJCAI2015}. For the identification-based framework, the ground-truth label is considered as a latent variable, and it is often refined in an iteration manner \cite{jin2003learning} \cite{Nguyen:KDDM2008} \cite{Yu2015ACML} \cite{Yu:ML2015}. Most of these methods mainly focus on identifying the unique ground-truth label, while the different contribution of other false candidate labels are not taken into consideration. Correspondingly, the pairwise ranking information of varying candidate labels and the candidate label relevance of the whole training data set are also regrettably ignored, which may lead these methods to be suboptimal.

In light of this observation, in this paper, we proposed a novel PLL approach called \textbf{HERA}, which simultaneously incorporates the \emph{\textbf{H}eterog\textbf{E}neous Loss} and the \emph{Spa\textbf{R}se and Low-r\textbf{A}nk  Scheme} to estimate the different label confidence for each instance while training the model. Specifically, we formalize the different labeling confidence levels as a latent label confidence vector, and then the confidence vector is estimated by minimizing the heterogeneous loss, where the heterogeneous loss integrates the strengths of both the pairwise ranking loss and the pointwise reconstruction loss to provide informative identification information for model training. Note that, different from existing label ranking scheme, the employed pairwise label ranking does not just refer to the difference between the unique ground-truth label and all other false candidate labels, but covers the labeling confidence ranking of any two candidate labels, which makes it clear to distinguish the different contributions of varying candidate labels.

\begin{figure}
\centering
\setlength{\abovecaptionskip}{0.cm}
\setlength{\belowcaptionskip}{0.cm}
\includegraphics[width = 4in,height=2in]{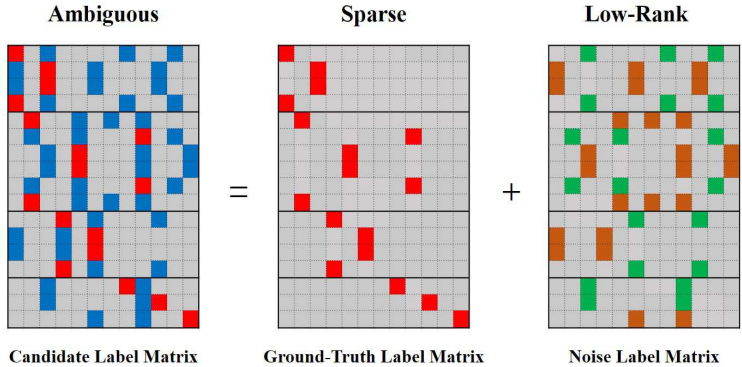}
\vspace{1mm}
\caption{The observed candidate label matrix $\textbf{Y}^{\top}$ can be decomposed into a ground-truth $\textbf{P}^{\top}$ and a noisy label matrix $\textbf{E}^{\top}$. Here, the ground-truth label matrix is encouraged to be sparse due to the nature of PLL that each instance can only correspond to a unique ground-truth label, while the remaining noisy label matrix tends to be of low-rank since instances with the same false candidate labels often recur statistically in many real-world scenarios of PLL.}
\label{fig0}
\vspace{-5mm}
\end{figure}

Moreover, in order to explore the global candidate label relevance among the whole training data sets, motivated by \cite{chiang2018using}, \cite{liurrsslrr2013}, \cite{xu2013speedup} and \cite{yu2014large} , we incorporate the sparse and low-rank scheme into our framework and assume that the observed candidate label matrix can be well approximated by decomposing into an ideal and a noisy label matrix. Different from existing multi-label learning approaches that encourage the ideal label matrix to be low-rank and the noisy label matrix to be sparse, in our framework, we pursue the ground-truth label matrix with sparse structure and constrain the false candidate label matrix with low-rank property. The sparse structure for ground-truth label matrix originates from the nature of PLL that each instance can only correspond to a unique ground-truth label, while the remaining noisy label matrix tends to be of low-rank since instances with the same false candidate labels often recur statistically in many real-world scenarios of PLL (Figure \ref{fig0}). Following this, we formalize our framework with the matrix decomposition constraint and separately regularize the ground-truth matrix and false candidate label matrix via sparse and low-rank procedure, which can not only intuitively exploit the candidate label relevance information of the whole training data but also theoretically avoid the algorithm falling into the trivial solution. Afterwards, we optimize the label confidence matrix and learning model in an alternating iteration manner. Experimental results on artificial as well as real-world PL data sets empirically show the effectiveness of the proposed method.

In summary, our main contributions lie in the following aspects:

\begin{itemize}
\item To the best of our knowledge, it is the first time to integrate the pairwise ranking loss and pointwise reconstruction loss simultaneously (as a heterogeneous loss) into partial label learning framework, which provides informative label identification information for model training.
\item Different from the widely used margin-based PLL methods that only considers the margin between the unique ground-truth label and all other false labels, the employed pairwise ranking loss can make full use of the difference of any two different candidate labels and improve the label identification ability of learning model.
\item Different from existing multi-label learning approaches that encourage the ideal label matrix to be low-rank and the noisy label matrix to be sparse, in our framework, we pursue the ground-truth label matrix with sparse structure and constrain the false candidate label matrix with low-rank property. In fact, such regularization operation for the label matrix can well satisfy the statistical constraint of PLL problem. (Figure \ref{fig0})
\end{itemize}

The rest of this paper is organized as follows. We first give a brief introduction about partial label learning in Section 2. Then, we present technical details of the proposed HERA algorithm and describe the optimization procedure for HERA in Section 3. Afterwards, in Section 4, we compare our proposed method with existing state-of-the-art methods on both artificial and real-world data sets. Finally, we conduct experimental analysis and conclude the whole paper.

\section{Related Work}

As a weakly supervised learning framework, partial label learning aims to learn from the training data with implicit labeling information. An intuitive strategy to formalize such framework is disambiguation, i.e. identifying the ground-truth label from the candidate label set. Generally speaking, existing PLL methods based on such strategy can be roughly grouped into the following three strategies:

\subsection{Average Disambiguation Strategy (ADS)}
ADS-based PLL methods often treat each candidate label equally and they make prediction for unseen instances by averaging the modeling outputs from all candidate labels. Following such strategy, Hullermeier et al. \cite{Huller:LNCS2005} adopts an instance-based model following $\arg\max_{y\in\mathcal{Y}}\sum_{i\in\mathcal{N}_{({\textbf{x}}^{*})}}\!\mathbb{I}(y\!\in\!S_i)$ to predict labels for unseen instances. Cour et al. \cite{Cour:lfpl-JMLR2011} identifies the ground-truth label from the ambiguous label set by averaging the modeling outputs from all candidate labels, i.e. $\frac{1}{|S_i|}\!\sum_{y_i\in S_i} F(\textbf{x},\bm{\Theta},y)$. Zhang et al.\cite{Zhang:IJCAI2015} and Chen et al. \cite{Chen2017A} also adopt an instance-based model and they make prediction via $k$-nearest neighbors weighted voting and minimum error reconstruction criterion. Recently, Tang and Zhang \cite{tang:AAAI2017} utilizes the boosting learning technique and improves the disambiguation model by simultaneously adapting the weights of training examples and the confidences of candidate labels. Intuitively, the above methods are clear and easy to implement, but it suffers from some common shortcomings that the output of the correct label is overwhelmed by the outputs of the other false positive labels, which may decrease the robustness of learning models.

\subsection{Identification Disambiguation Strategy (IDS)}
\label{Section2-1}
Different from averaging the output from all candidate labels, IDS-based PLL methods aims to directly identifying the ground-truth label from its corresponding candidate label set. Following such strategy, most existing PLL methods often regard the ground-truth label as a latent variable first, identified as $\arg\max_{y\in{S_i}}F(\textbf{x},{\bm{\Theta}},y)$, and then refine the model parameter $\bm{\Theta}$ iteratively by utilizing some specific criterions, such as maximum likelihood criterion  \cite{Grandvalet:CWP2004} \cite{jin2003learning} \cite{Liu:acmmmfsll-NIPS2012} , i.e. $\sum_{i=1}^{n}\log (\sum_{y\in{S_i}}F(\textbf{x},{\bm{\Theta}},y))$ and maximum margin criterion \cite{Nguyen:KDDM2008} \cite{Yu:ML2015}, i.e.  $\sum_{i=1}^{n}(\max_{y\in{S_i}}F(\textbf{x},{\bm{\Theta}},y)\!-\!\max_{y\not\in{S_i}}F(\textbf{x},{\bm{\Theta}},y))$. However, these methods mainly focus on the difference between the unique ground-truth label and all other false candidate labels, while the confidence of each candidate label being the ground-truth label tend to be regrettably ignored. In recent year, considering such shortcoming, some attempts try to improve the modeling disambiguation ability by exploiting the labeling confidence of different candidate labels, where either label distribution or label enhancement procedure is often employed into its corresponding PLL framework for improving the learning model \cite{feng2018leveraging} \cite{xu2019partial}.

\subsection{Disambiguation-Free Strategy (DFS)}
More recently, different from the above two disambiguation-based strategies, some methods aim to learn from PL data by fitting these data to existing learning techniques instead of disambiguation. Zhang et al. \cite{zhang:IEEET2017} proposes a disambiguation-free algorithm named \emph{PL-ECOC}, which utilizes \emph{Error-Correcting Output Codes} (ECOC) coding matrix \cite{d1994solving} and transfers the PLL problem into binary learning problem. Xuan et al. \cite{TEBDfPLL-IJCAI2018} proposes another disambiguation-free algorithm called \emph{PALOC}, which enables binary decomposition for PLL data in a more concise manner without relying on extra manipulations. The experiments have empirically demonstrated that such DFS-based methods can achieve competitive performance against the above two kinds of disambiguation-based methods.

\section{The Proposed Method}

Formally speaking, we denote the $d$-dimensional input space as $\mathcal{X}\!\in\!\mathbb{R}^{d}$, and the output space as $\mathcal{Y}\!=\! \{1,2,\ldots,q\}$ with $q$ class labels. PLL aims to learn a classifier $\textbf{f}:\!\mathcal{X}\!\mapsto\!\mathcal{Y}$ from the PL training data $\mathcal{D}\!=\!\{(\textbf{x}_i,S_i)\} (1\!\leq\!{i}\!\leq\!{n})$ with $n$ instances, where the ground-truth label $y_i\!\in\! S_i\!\subseteq\!\mathcal{Y}$ of each instance $\textbf{x}_{i}\!\in\!\mathcal{X}$ is not directly accessible during the training phase. For the convenience of description, we denote $\textbf{X}\!=\![\textbf{x}_1, \textbf{x}_2, \ldots, \textbf{x}_n]\!\in\!\mathbb{R}^{d\times n}$ as the training instance matrix, $\textbf{Y}\!=\!{\{0,1\}}^{q\times n}$ as the candidate label matrix, where $\textbf{Y}_{ij}\!=\!1$ (i.e. $\textbf{Y}_{ij}\!\in\! S_j$) represents label $i$ is the candidate label of instance $j$, while $\textbf{Y}_{ij}\!=\!0$ (i.e. $\textbf{Y}_{ij}\!\in\!(\mathcal{Y}\backslash S_j)$), otherwise. Besides, $\textbf{P}\!=\![\textbf{p}_1, \textbf{p}_2, \ldots, \textbf{p}_n]\!\in\!{[0,1]}^{q\times n}$ is denoted as the labeling confidence matrix, where $\textbf{p}_i\!=\![\textbf{p}_{i1}, \textbf{p}_{i2}, \ldots, \textbf{p}_{iq}]$ represents the labeling confidence of $q$ labels for $\textbf{x}_i$.

\subsection{Formulation}

As is described in Section 2, disambiguation strategy has been widely employed in many PLL learning frameworks. However, existing methods suffer from a common weakness, i.e. the learning framework is sensitive to the false positive labels that co-occur with the ground-truth label and the different labeling confidence levels of candidate labels are regrettably ignored during the training process. In this paper, to alliviate the above weakness, we consider the ground-truth label matrix as a latent labeling confidence matrix and propose a novel unified framework to estimate such latent labeling confidence matrix while training the desired model simultaneously, which is defined as \textbf{OP (1)}:
\begin{flalign*}
\min_{\textbf{P},\textbf{f}}L(\textbf{X};\textbf{f},\textbf{P})+ \beta\Omega(\textbf{f}) + \mu\Psi(\textbf{P})
\end{flalign*}where $L(\cdot)$ is the loss function, $\Omega(\textbf{f})$ controls the complexity of the model \textbf{f}, $\Psi(\textbf{P})$ serves as the regularization to guarantee an optimal estimation of labeling confidence matrix $\textbf{P}$, and $\beta, \mu$ are the parameters to control the balance of three terms.

To satisfy the different structure of PLL framework, most existing methods either utilize the \emph{max-margin} hinge loss function as ranking loss to distinguish varying candidate labels following $\sum_{i=1}^{n}(1-(F(\textbf{x},\bm{\Theta}, \bar{y})-\max_{y\neq \bar{y}}F(\textbf{x},\bm{\Theta}, y)))$ \cite{Yu:ML2015}, or employ the \emph{least square} loss function as reconstruction loss to conduct model training following ${\Vert\textbf{Y}-\textbf{W}^{\top}\textbf{X}\Vert}^{2}$ \cite{feng2018leveraging}, where single loss can only exploit less disambiguation information from training data and decrease the robustness of the learning model. Being aware of this, in order to leverage more informative disambiguation information and improve the robustness of learning model, we integrate the above two kinds of loss function into a unified heterogeneous loss  $L(\textbf{X};\textbf{f},\textbf{P})$ as

\begin{flalign}
\sum\limits_{i,j,k} \frac{1}{{\gamma}} \tilde{\textbf{P}}_{ijk}\!\cdot\!{\mathcal{L}_{rank}[((\textbf{w}_j\!-\!\textbf{w}_k)^\top\!\!\cdot\!\textbf{x}_i)^2]}
+\frac{\alpha}{2}{\Vert\textbf{P}\!-\!\textbf{W}^{\top}\!\textbf{X}\Vert}_{F}^{2}
\end{flalign}where $i\in[n]$, $j,k\in[q]$, and the two parts separately represent the pairwise ranking loss and the pointwise reconstruction loss. Specifically, $\tilde{\textbf{P}}_{ijk}\!\!=\!\! (\textbf{p}_{ij}\!-\!\textbf{p}_{ik})^2$ measures the label confidence consistence between $j$-th candidate label and $k$-th candidate label of $\textbf{x}_i$, while $\mathcal{L}_{rank}[x]\!=\!\log[1\!+\!\exp(-x)]$ describes the modeling outputs consistence. Intuitively, if the label confidence gap between $j$-th label and $k$-th label is large, $\tilde{\textbf{P}}_{ijk}$ tends to be large, which encourages $\mathcal{L}_{rank}$ to be small. In contrast, if the gap is small, $\tilde{\textbf{P}}_{ijk}$ tends to be small, which can result in a large value of the $\mathcal{L}_{rank}$. Besides, $\textbf{W}=[\textbf{w}_1, \textbf{w}_2, \ldots, \textbf{w}_q]\in\mathbb{R}^{d\times q}$ is the weight matrix of the classifier \textbf{f}. ${\gamma} = q^{2}$ is a constant for normalization on each instance and $\alpha$ is the trade-off parameter to balance the two loss functions. Note that, such cost-sensitive pairwise ranking loss can clearly distinguish the difference between any two labels, which is different from the \emph{max-margin} hinge ranking loss that only aims to identify the unique correct label.

Moreover, to optimally estimate the labeling confidence matrix \textbf{P}, the global correlations among the candidate labels are incorporated into the framework. To achieve this, we treat the ground-truth labels as outliers and decompose the observed candidate label matrix to $\textbf{Y} = \textbf{P} + \textbf{E}$, where $\textbf{P}$ is the sparse component, which captures the constraint of PLL that one instance can only correspond to a unique ground-truth label, and $\textbf{E}$ is of low-rank, which depicts the statistical label correlations among different training instances.
\begin{flalign}
\Psi(\textbf{P}) = \mu{\|\textbf{P}\|}_1+\nu{\|\textbf{E}\|}_{*}, \quad \rm{where} \quad \textbf{Y}=\textbf{P}+\textbf{E}.
\end{flalign}
In addition, in order to control the model complexity, we adopt the widely used squared Frobenius norm of \textbf{W}, i.e. $\Omega(\textbf{f})={\Vert\textbf{W}\Vert}^{2}_{F}$. And the final framework of HERA can be formulated as the following optimization problem \textbf{OP (2)}:
\begin{flalign*}
\min\limits_{\textbf{W},\textbf{P},\textbf{E}}&\sum\limits_{i,j,k} \frac{1}{\gamma}\tilde{\textbf{P}}_{ijk}\!\cdot\!{\mathcal{L}_{rank}[((\textbf{w}_j\!-\!\textbf{w}_k)^\top\!\!\cdot\!\textbf{x}_i)^2]}
+\frac{\alpha}{2}{\Vert\textbf{P}\!-\!\textbf{W}^{\top}\!\textbf{X}\Vert}_{F}^{2}\\
&+\beta{\Vert\textbf{W}\Vert}_{F}^{2}+\mu{\Vert\textbf{P}\Vert}_{1}+\nu{\Vert\textbf{E}\Vert}_{*} \\ \\
&s.t. \quad \textbf{Y} = \textbf{P} + \textbf{E}; \,\textbf{P}\geq \textbf{0},\,\textbf{E}\geq \textbf{0}.
\end{flalign*}where $\Vert\cdot\Vert_{1}$ is the ${\ell}_{1}$ norm and $\Vert\cdot\Vert_{*}$ is the nuclear norm, which separately provides a good surrogate for sparse representation and low-rank representation.

In summary, in the objective function of \textbf{OP(2)}, the first term is called \emph{heterogeneous loss} that integrates both \emph{pairwise ranking loss} and \emph{pointwise reconstruction loss} into a unified loss function, which can provide informative identification information for model training. The second term is named as \emph{model complexity} that is utilize to control the complexity of learning model, which can avoid training model tends to be overfitting during the training process. The third term serves as \emph{regularization} that decomposes the candidate label matrix into a ground-truth label matrix and a noisy label matrix, and separately regularizes them with sparse and low-rank constraints, which can lead the model to obtain an optimal estimation of labeling confidence matrix. Overall, the whole framework of HERA can utilize much more global and local structural information from both feature and label space, and encourage the learning model to be effective and robust.

Afterwards, considering that \textbf{OP(2)} is an optimization problem with multiple variables, which is difficult to solve directly, we employ an alternating optimization procedure to update \textbf{W}, \textbf{P} and \textbf{E} iteratively. The details of optimization are exhibited in the following subsection.

\subsection{Optimization}
To optimize the target function conveniently, we convert \textbf{OP (2)} to the following equivalent problem \textbf{OP (3)}:
\begin{flalign*}
\min\limits_{\textbf{W},\textbf{P},\textbf{J},\textbf{E}}&\sum\limits_{i,j,k} \frac{1}{\gamma}\tilde{\textbf{P}}_{ijk}\!\cdot\!{\mathcal{L}_{rank}[((\textbf{w}_j\!-\!\textbf{w}_k)^\top\!\!\cdot\!\textbf{x}_i)^2]}
+\frac{\alpha}{2}{\Vert\textbf{P}\!-\!\textbf{W}^{\top}\!\textbf{X}\Vert}_{F}^{2}\\
&+\beta{\Vert\textbf{W}\Vert}_{F}^{2}+\mu{\Vert\textbf{J}\Vert}_{1}+\nu{\Vert\textbf{E}\Vert}_{*} \\
&s.t. \quad \textbf{Y} = \textbf{P} + \textbf{E}, \quad \textbf{P} = \textbf{J}; \quad \textbf{P}\geq \textbf{0}, \quad \textbf{E}\geq \textbf{0}. \nonumber
\end{flalign*}\quad Intuitively, the optimization problem of \textbf{OP(3)} is a constrained optimization problem, which can be solved by using an Augmented Lagrange Multiplier (ALM) technique. And we write its ALM form as follows \textbf{OP(4)}:
\begin{flalign*}
\min\limits_{\textbf{W},\textbf{P},\textbf{J},\textbf{E}}&\sum\limits_{i,j,k} \frac{1}{\gamma}\tilde{\textbf{P}}_{ijk}\!\cdot\!{\mathcal{L}_{rank}[((\textbf{w}_j\!-\!\textbf{w}_k)^\top\!\!\cdot\!\textbf{x}_i)^2]}
+\frac{\alpha}{2}{\Vert\textbf{P}\!-\!\textbf{W}^{\top}\!\textbf{X}\Vert}_{F}^{2}\\
&+\beta{\Vert\textbf{W}\Vert}_{F}^{2}+\mu{\Vert\textbf{J}\Vert}_{1}+\nu{\Vert\textbf{E}\Vert}_{*}
+tr(\textbf{M}\!^{\top}\!\cdot\!{(\textbf{Y}\!-\!\textbf{P}\!-\!\textbf{E})})\\
&+tr(\textbf{N}^{\top}\!\cdot\!{(\textbf{P}\!-\!\textbf{J})})
+\frac{\lambda}{2}{\Vert\textbf{Y}\!-\!\textbf{P}\!-\!\textbf{E}\Vert}_{F}^{2}
+\frac{\rho}{2}{\Vert\textbf{P}\!-\!\textbf{J}\Vert}_{F}^{2}
\end{flalign*}where $tr(\cdot)$ is the trace of matrix. $\textbf{M},\textbf{N}\in\mathbb{R}^{q\times n}$ are Lagrange multiplier matrices, $\lambda$ and $\rho$ are the penalty parameters.

Obviously, for each of the four matrices $\textbf{W}, \textbf{P}, \textbf{J}, \textbf{E}$ to be solved in particularly \textbf{OP(4)}, the objective function is convex if the remaining three matrices are kept fixed. Thus, \textbf{OP(4)} can be solved iteratively via the following steps:

\textbf{Step 1: Calculate $\textbf{W}$.} Fixing the other variables, we can calculate $\textbf{W}$ by minimizing the following objective function:
\begin{flalign}
\begin{split}\label{Eq-W}
\textbf{W}^* = \arg\min\limits_{\textbf{W}}&\sum\limits_{i,j,k} \frac{1}{\gamma}\tilde{\textbf{P}}_{ijk}\!\cdot\!{\mathcal{L}_{rank}[((\textbf{w}_j\!-\!\textbf{w}_k)^\top\!\!\cdot\!\textbf{x}_i)^2]}\\
&+\frac{\alpha}{2}{\Vert\textbf{P}\!-\!\textbf{W}^{\top}\!\textbf{X}\Vert}_{F}^{2}+\beta{\Vert\textbf{W}\Vert}_{F}^{2}
\end{split}
\end{flalign}where the loss function of Eq (\ref{Eq-W}) is differentiable, thus $\textbf{W}$ can be optimized via the standard gradient descent algorithm. The $\textbf{W}$ in each iteration is updated via:
\begin{flalign}
\label{Eq-W-up}
\textbf{W}^{{t+1}}& = \textbf{W}^{{t}} - \\ \nonumber
&\eta\cdot\left(\sum_{i,j,k}\frac{-2(\textbf{P}_{ji}-\textbf{P}_{ki})^2\cdot{e^{-{\left((\textbf{W}^t_{\cdot j}-\textbf{W}^t_{\cdot k})\cdot \textbf{X}_{\cdot i}\right)}^2}}\cdot \mathcal{F}(\textbf{X}_i,\textbf{H}_{jk})}{ln2\cdot\gamma\left(1+e^{-{\left((\textbf{W}^t_{\cdot j}-\textbf{W}^t_{\cdot k})\cdot \textbf{X}_{\cdot i}\right)}^2}\right)}-\alpha(\textbf{X}\cdot(\textbf{P}^{\top}-\textbf{X}^{\top}\textbf{W}^t))+2\beta\textbf{W}^t\right)
\end{flalign}where $\eta$ is the stepsize of gradient descent, $\textbf{H}$ is an indicator matrix with equal size of matrix $\textbf{W}$. $\mathcal{F}(\textbf{X}_i,\textbf{H}_{jk})$ is a function to obtain the indicator matrix $\textbf{H}$, where $\textbf{H}_{\cdot j}={\textbf{X}_{\cdot i}}$, $\textbf{H}_{\cdot k}=-{\textbf{X}_{\cdot i}}$, otherwise $\textbf{H}_{\tilde{j}\tilde{k}}=0$.

\textbf{Step 2: Calculate $\textbf{P}$.} Fixing the other variables, the subproblem to variable $\textbf{P}$ is simplified as follows:
\begin{flalign}
\begin{split}\label{Eq-P}
\textbf{P}^* =
\arg\min\limits_{\textbf{P}}&\sum\limits_{i,j,k} \frac{1}{\gamma}\tilde{\textbf{P}}_{ijk}\!\cdot\!{\mathcal{L}_{rank}[((\textbf{w}_j\!-\!\textbf{w}_k)^\top\!\!\cdot\!\textbf{x}_i)^2]}
+\frac{\alpha}{2}{\Vert\textbf{P}\!-\!\textbf{W}^{\top}\!\textbf{X}\Vert}_{F}^{2}\\
&+tr(\textbf{M}\!^{\top}\!\cdot\!{(\textbf{Y}\!-\!\textbf{P}\!-\!\textbf{E})})
+tr(\textbf{N}^{\top}\!\cdot\!{(\textbf{P}\!-\!\textbf{J})})\\
&+\frac{\lambda}{2}{\Vert\textbf{Y}\!-\!\textbf{P}\!-\!\textbf{E}\Vert}_{F}^{2}
+\frac{\rho}{2}{\Vert\textbf{P}\!-\!\textbf{J}\Vert}_{F}^{2}
\end{split}
\end{flalign}\quad Similar to Eq (\ref{Eq-W}), the subgradient of Eq (\ref{Eq-P}) is easy to obtain, and we also employ the gradient descent procedure to solve it. Specifically,
\begin{flalign}
\label{Eq-P-up}
\textbf{P}^{t+1} = \textbf{P}^{{t}}&-
\eta\cdot\bigg(\sum_{i,j,k}\frac{2}{\gamma}(\textbf{P}_{ji}-\textbf{P}_{ki})\cdot \log\left(1+e^{-{\left((\textbf{W}_{\cdot j}-\textbf{W}_{\cdot k})\cdot\textbf{X}_{\cdot i}\right)}^2}\right)\cdot{\mathcal{F}(\textbf{1},\textbf{I}_{jk})}\\ \nonumber &-\alpha\cdot(\textbf{P}-\textbf{W}^\top\textbf{X})
-\textbf{M}+\textbf{N}-\lambda\cdot(\textbf{Y}-\textbf{P}-\textbf{E})+\rho\cdot(\textbf{P}-\textbf{J})\bigg)
\end{flalign}
Here, $\textbf{1}\in\mathbb{R}^{n\times 1}$ is an $n$-dimensional all-ones vector, $\textbf{I}_{jk}\in\mathbb{R}^{q\times n}$ is an indicator matrix, and $\mathcal{F}(\textbf{1},\textbf{I}_{jk})$ indicates that $\textbf{I}_{\cdot j}={\textbf{1}}$, $\textbf{I}_{\cdot k}=-{\textbf{1}}$, otherwise $\textbf{I}_{\tilde{j}\tilde{k}}=0$. In addition, to satisfy the constraint of $\textbf{P}\geq\textbf{0}$, we project the updated $\textbf{P}^*$ into the feasible set, i.e. $\textbf{P}_{ij}^* = \max{(\textbf{P}_{ij}^*,\textbf{0})}$.

\textbf{Step 3: Calculate $\textbf{J}$.} By fixing the other variables, the sub-problem to variable $\textbf{J}$ is reformulated as:
\begin{flalign}
\label{Eq-J}
\textbf{J}^* = \arg\min\limits_{\textbf{J}}\, &\mu{\Vert\textbf{J}\Vert}_{1}+tr(\textbf{N}^{\top}\!\cdot\!{(\textbf{P}\!-\!\textbf{J})})
+\frac{\rho}{2}{\Vert\textbf{P}\!-\!\textbf{J}\Vert}_{F}^{2}
\end{flalign} \quad According to \cite{zhu2010image}, Eq (\ref{Eq-J}) has the closed form solution, and the variable $\textbf{J}$ can be updated following $\textbf{J}=\mathcal{S}_{\frac{\mu}{\rho}}[\textbf{P}-\frac{1}{\rho}\textbf{N}]$, where
\begin{flalign}
\label{sol-J}
\mathcal{S}_{\varepsilon}[\textbf{G}] =
\begin{cases}
\textbf{G}-\varepsilon&,\quad\textrm{if}\quad \textbf{G} > \varepsilon\\
\textbf{G}+\varepsilon&,\quad\textrm{if}\quad \textbf{G} < -\varepsilon\\
\quad 0 &,\quad \textrm{otherwise}\\
\end{cases}
\end{flalign}
\textbf{Step 4: Calculate $\textbf{E}$.} When fixing all the other variables that irrelevant to $\textbf{E}$, we have:
\begin{flalign}
\label{Eq-E}
\textbf{E}^*\!=\! \arg\min\limits_{\textbf{E}}\nu{\Vert\textbf{E}\Vert}_{*}\!+\!tr(\textbf{M}\!^{\top}\!\cdot\!{(\textbf{Y}\!-\!\textbf{P}\!-\!\textbf{E})})
\!+\!\frac{\lambda}{2}{\Vert\textbf{Y}\!-\!\textbf{P}\!-\!\textbf{E}\Vert}_{F}^{2}
\end{flalign}\quad Similarly, Eq (\ref{Eq-E}) also has the closed form, and the variable $\textbf{E}$ can be optimized following $\textbf{E}=\textbf{U}\mathcal{S}_{\frac{\nu}{\lambda}}[\bm{\Sigma}]\textbf{V}^{\top}$, where $\textbf{U}\bm{\Sigma}\textbf{V}^{\top}$ is the singular value decomposition of $(\textbf{Y}-\textbf{P}+\frac{1}{\lambda}\textbf{M})$. Meanwhile, the updated $\textbf{E}^*$ is also projected into the feasible set to satisfy the constraint of $\textbf{E}\geq\textbf{0}$.

Finally, the Lagrange multiplier matrices $\textbf{M}$, $\textbf{N}$ and the regularization terms $\lambda$, $\rho$ are updated based on Linearized Alternating Direction Method (LADM).
\begin{flalign}
\label{sol-lam-rho}
\begin{split}
\textbf{M}^{t+1} &= \textbf{M}^{t} + \lambda(\textbf{Y}^{t}-\textbf{P}^{t}-\textbf{E}^{t})\\
\textbf{N}^{t+1} &= \textbf{N}^{t} + \rho(\textbf{P}^{t}-\textbf{J}^{t})\\
\lambda^{t+1}    &= \min({\lambda}_{max}, \tau {\lambda}^{t})\\
\rho^{t+1}       &= \min({\rho}_{max}, \tau {\rho}^{t})
\end{split}
\end{flalign}\quad During the entire optimization process, we first initialize the required variables, and then repeat the above steps until the algorithm converges. Finally, we make prediction for unseen instances following the steps of Section \ref{section-Prediction}. Algorithm \ref{algorithm} summarizes the whole optimization process of HERA.

\begin{algorithm}[tb]
\caption{The Algorithm of HERA}
\label{algorithm}
\textbf{Input}:\\
$\mathcal{D}$ :the PL training date set $(\textbf{x}_i,S_i)$\\
$\textbf{x}^*$:the unseen instance \\
\textbf{Parameter}: $\alpha, \beta, \mu, \nu, \lambda_{max}, \rho_{max}$\\
\textbf{Output}: \textbf{W}, \textbf{P}, $y^*$
\begin{algorithmic}[1]
\STATE Initialize \textbf{W}, \textbf{P}, \textbf{J}, \textbf{E} and $\alpha, \beta, \mu, \nu, k, \lambda_{max}, \rho_{max}$.
\WHILE{$t \leq Iter_{max}$}
\STATE Update \textbf{W} by solving Eq (\ref{Eq-W}).
\STATE Update \textbf{P} by solving Eq (\ref{Eq-P}).
\STATE Update \textbf{J} by solving Eq (\ref{Eq-J}).
\STATE Update \textbf{E} by solving Eq (\ref{Eq-E}).
\STATE Update $\textbf{M}, \textbf{N}, \lambda, \rho$ following Eq (\ref{sol-lam-rho}).
\STATE Calculate the $loss$ following \textbf{OP (4)}.
\IF {$\|loss^{(t+1)}-loss^{(t)}\| \le loss_{max}$}
\STATE break.
\ENDIF
\ENDWHILE
\STATE Prediction: $y^*\!=\!\textbf{S}^{(k+1)\times (k+1)}\!\cdot\![\textbf{P}^{k\times q};\textbf{W}^{\top}\!\cdot\!\textbf{x}^*]$
\STATE \textbf{return} $y^*$
\end{algorithmic}
\end{algorithm}

\subsection{Prediction}
\label{section-Prediction}
In this section, we utilize a hybrid label prediction algorithm that is based on the $k$-nearest neighbor scheme to make prediction for unseen instances, where the prior instance-similarity information and the modeling outputs information jointly contribute to increasing the classification accuracy.

Specifically, we construct the similarity matrix $\textbf{S}={[s_{ij}]}_{(k+1)\times(k+1)}$ to characterize the instance similarity between the unseen instance and its $k$-nearest neighbors, where $s_{ij} = \exp{(-{\Vert\textbf{x}_i-\textbf{x}_j\Vert}^2_2/{\sigma}^2)}$, $\sigma = \sum_{i=1}^{k+1}{\Vert\textbf{x}_i-\textbf{x}_{im}\Vert}_2/(k+1)$ and $\textbf{x}_{im}$ is the $m$-th nearest neighbor of $\textbf{x}_i$. Meanwhile, we concatenate the labeling confidence matrix $\textbf{P}$ and the modeling output $\textbf{W}^{\top}\textbf{x}^{*}$ into a unified matrix $\tilde{Y}$, where $\tilde{Y}=[\textbf{P};\textbf{W}^{\top}\textbf{x}^{*}] \in \mathbb{R}^{(k+1)\times q}$. Thereafter, we can obtain the final predicted label via the following label propagation scheme:
\begin{flalign}
\label{predictlabel}
\begin{split}
y^{*} = \max_{\tilde{Y}_{k+1}} \textbf{S}\!\cdot\!{[\textbf{P};\textbf{W}^{\top}\textbf{x}^{*}]}
\end{split}
\end{flalign}where $\textbf{S}\cdot[\textbf{P};\textbf{W}^\top\textbf{x}^{*}]$ is the updated labeling matrix and the final predicted label is obtained by maximizing the outputs of prediction model, i.e. maximizing the $(k+1)$-th row of  $\textbf{S}\cdot[\textbf{P};\textbf{W}^\top\textbf{x}^{*}]$.

\section{Experiments}

\subsection{Experimental Setup}
\label{section-exp-setup}
To evaluate the performance of the proposed HERA method, we implement experiments on nine controlled UCI data sets and six real-world data sets: \textbf{(1) Controlled UCI data sets}. Under different configurations of three controlling parameters (i.e. $\emph{p}$, $\emph{r}$ and $\epsilon$), the four UCI data sets generate 252 $(9\times 4\times 7)$ artificial partial-label data sets \cite{Chen:IEEET2014}\cite{Cour:lfpl-JMLR2011}. Here, $\emph{p}\!\in\!{\{0.1,0.2,\ldots,0.7\}}$ is the proportion of partial-label examples, $\emph{r}\!\in\!{\{1,2,3\}}$ is the number of false candidate labels, and $\epsilon\!\in\!\{0.1,0.2,\ldots,0.7\}$ is the co-occurring probability between one coupling candidate label and the ground-truth label. \textbf{(2) Real-World (RW) data sets }. These data sets are collected from the following four task domains: (A) {\emph{Facial Age Estimation}} [FG-NET]; (B) {\emph{Image Classification}} [MSRCv2]; (C) {\emph{Bird Sound Classification}} [BirdSong]; (D) {\emph{Automatic Face Naming}} [Lost] [Soccer Player] [Yahoo! News]. Table \ref{table1} and \ref{table2} separately summarize the characteristics of UCI and Real World data sets, including the number of examples (\textbf{EXP*}), the number of the feature (\textbf{FEA*}), the whole number of class labels (\textbf{CL*}) and the average number of class labels (\textbf{AVG-CL*}).


\begin{table}
\centering
\caption{Characteristics of the Controlled UCI data sets}
\vspace{-1mm}
\label{table1}
\small
\resizebox{13cm}{!}{
\begin{tabular}{cccc|c}
\hline \hline
Controlled UCI data sets  & EXP*     & FEA*     & CL*        &CONFIGURATIONS          \\ \hline
Glass         & 214      & 9        & 6           &          \\
Ecoli         & 336      & 7        & 8           &  \\
Dermatology   & 366      & 34       & 6           & $r = 1,  p\in\{0.1, 0.2, \ldots,0.7\}$  \\
Vehicle       & 846      & 18       & 4           & \\
Segment       & 2310     & 18       & 7           & $r = 2,  p\in\{0.1, 0.2, \ldots,0.7\}$\\
Abalone       & 4177     & 7        & 29          & \\
Letter        & 5000     & 16       & 26          & $r = 3,  p\in\{0.1, 0.2, \ldots,0.7\}$\\
Satimage      & 6345     & 36       & 7           &\\
Pendigits     & 10992    & 16       & 10          &\\ \hline \hline
\end{tabular}}
\vspace{-1mm}
\end{table}

\begin{table*}
\centering
\caption{Characteristics of the Real-World data sets}
\vspace{-1mm}
\label{table2}
\resizebox{13cm}{!}{
\begin{tabular}{cccccc}
\hline \hline
RW data sets   & EXP*     & FEA*     & CL*         & AVG-CL*   &TASK DOMAIN     \\ \hline

Lost          & 1122     & 108      & 16          & 2.33      &\emph{Automatic Face Naming} \cite{Cour:lfpl-JMLR2011}      \\
MSRCv2       & 1758     & 48       & 23          & 3.16      &\emph{Image Classification} \cite{Briggs:rlsimfmia-KDDM2012}  \\
FG-NET        & 1002     & 262      & 99          & 7.48      &\emph{Facial Age Estimation} \cite{Panis:FG-NET-JAH2015} \\
BirdSong      & 4998     & 38      & 13          & 2.18      &\emph{Bird Song Classification} \cite{Briggs:rlsimfmia-KDDM2012} \\
Soccer Player & 17472    & 279      & 171         & 2.09      &\emph{Automatic Face Naming} \cite{Zeng:lbaali-CVPR2013} \\
Yahoo! News & 22991    & 163      & 219         & 1.91      &\emph{Automatic Face Naming} \cite{Guill:mimlfalbof-ECCV2010} \\ \hline \hline
\end{tabular}}
\vspace{-1mm}
\end{table*}

Meanwhile, we employ six state-of-the-art PLL methods from three categories for comparative studies: \textbf{PL-SVM}, \textbf{PL-KNN}, \textbf{CLPL}, \textbf{LSB-CMM}, \textbf{PL-ECOC} and \textbf{PALOC}, where the configured parameters are utilized according to the suggestions in respective literatures.
\begin{itemize}
\item \textbf{PL-SVM} \cite{Nguyen:KDDM2008}: An identification-based disambiguation partial label learning algorithm, which learns from the partial label data by utilizing maximum-margin strategy. [suggested configuration: $\lambda\!\in\!{\{10^{-3}, \ldots, 10^{3}\}}$] ;
\item \textbf{PL-KNN} \cite{Huller:LNCS2005}: Based on $k$-nearest neighbor strategy, it makes prediction by averaging the outputs of the learning model. [suggested configuration: $k=10$];
\item \textbf{CLPL} \cite{Cour:lfpl-JMLR2011}: A convex optimization partial label learning method, which also makes prediction for unseen instances by adopting the averaging-based disambiguation strategy [suggested configuration: SVM with square hinge loss];
\item \textbf{LSB-CMM} \cite{Liu:acmmmfsll-NIPS2012}: A maximum-likelihood based partial label learning method by utilizing the identification-based disambiguation strategy. [suggested configuration: \emph{q} mixture components];
\item \textbf{PL-ECOC} \cite{zhang:IEEET2017}: Based on a coding-decoding procedure, it learns from partial label data in a disambiguation-free manner [suggested configuration: $L = \lceil\log_{2}(q)\rceil$];
\item \textbf{PALOC} \cite{TEBDfPLL-IJCAI2018}: An approach that adapts one-vs-one decomposition strategy to learn from PL examples [suggested configuration: $\mu = 10$];
\end{itemize}

Before conducting the experiments, we pre-introduce the values of parameters employed in our framework. Specifically, we set $\alpha$ among $\{2*10^{-3},2*10^{-2},\ldots,2*10^{0}\}$ via cross-validation. And the initial values of $\beta, \mu, \nu$ are empirically set among $\{10^{-3},10^{-2},\ldots,10^{3}\}$. Furthermore, the other variables are set as $k=10$, $loss_{max}=10^{-6}$, $\lambda = \rho = 10^{-6}$ and $\tau=1.05$. After initializing the above variables, we adopt ten-fold cross-validation to train the model and obtain the classification accuracy on each data set.

\begin{figure*}
\centering
\begin{tabular}{ccc}
\includegraphics[width = 1.77in,height=1.3in]{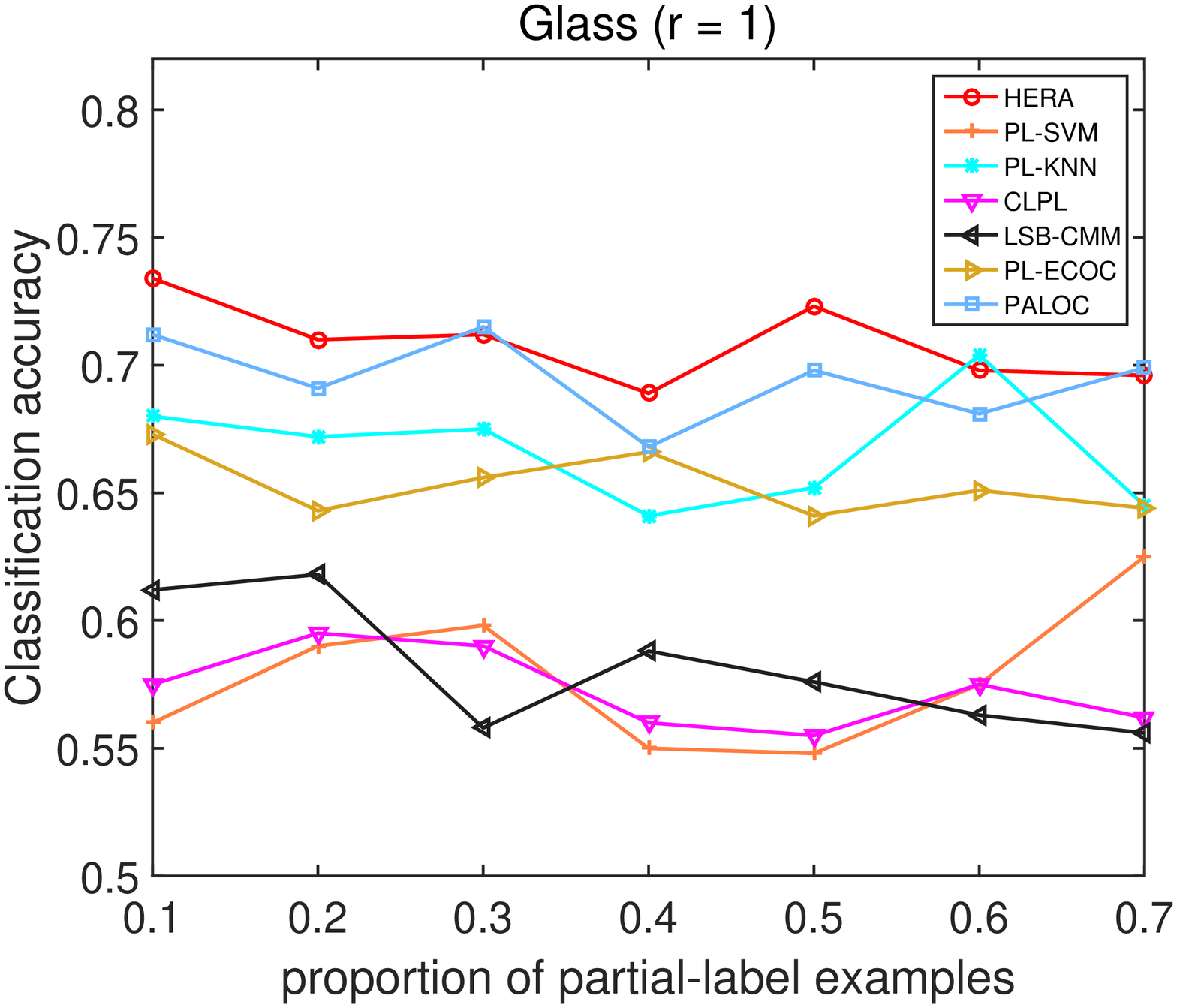}&\includegraphics[width = 1.77in,height=1.3in]{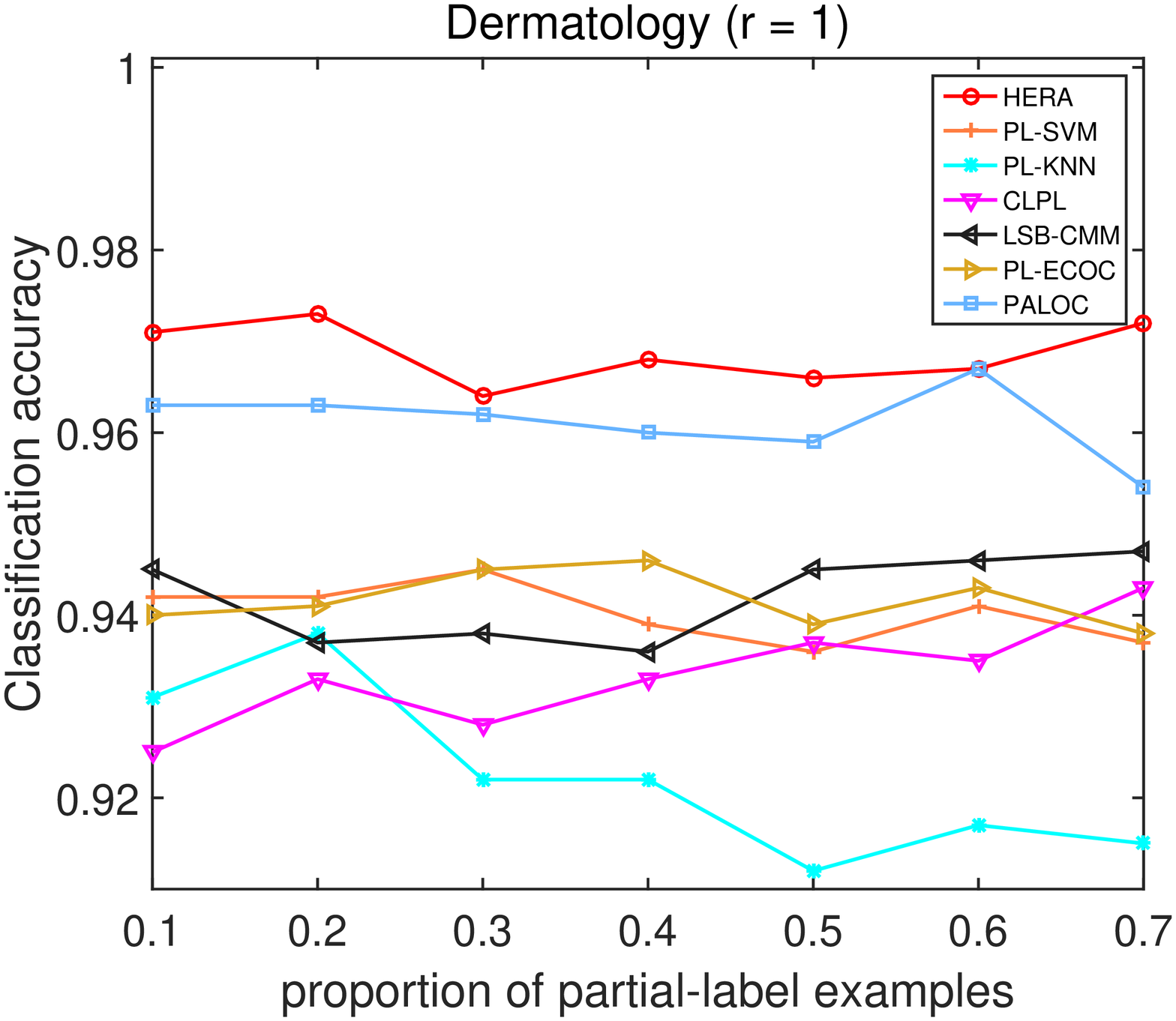}&\includegraphics[width = 1.77in,height=1.3in]{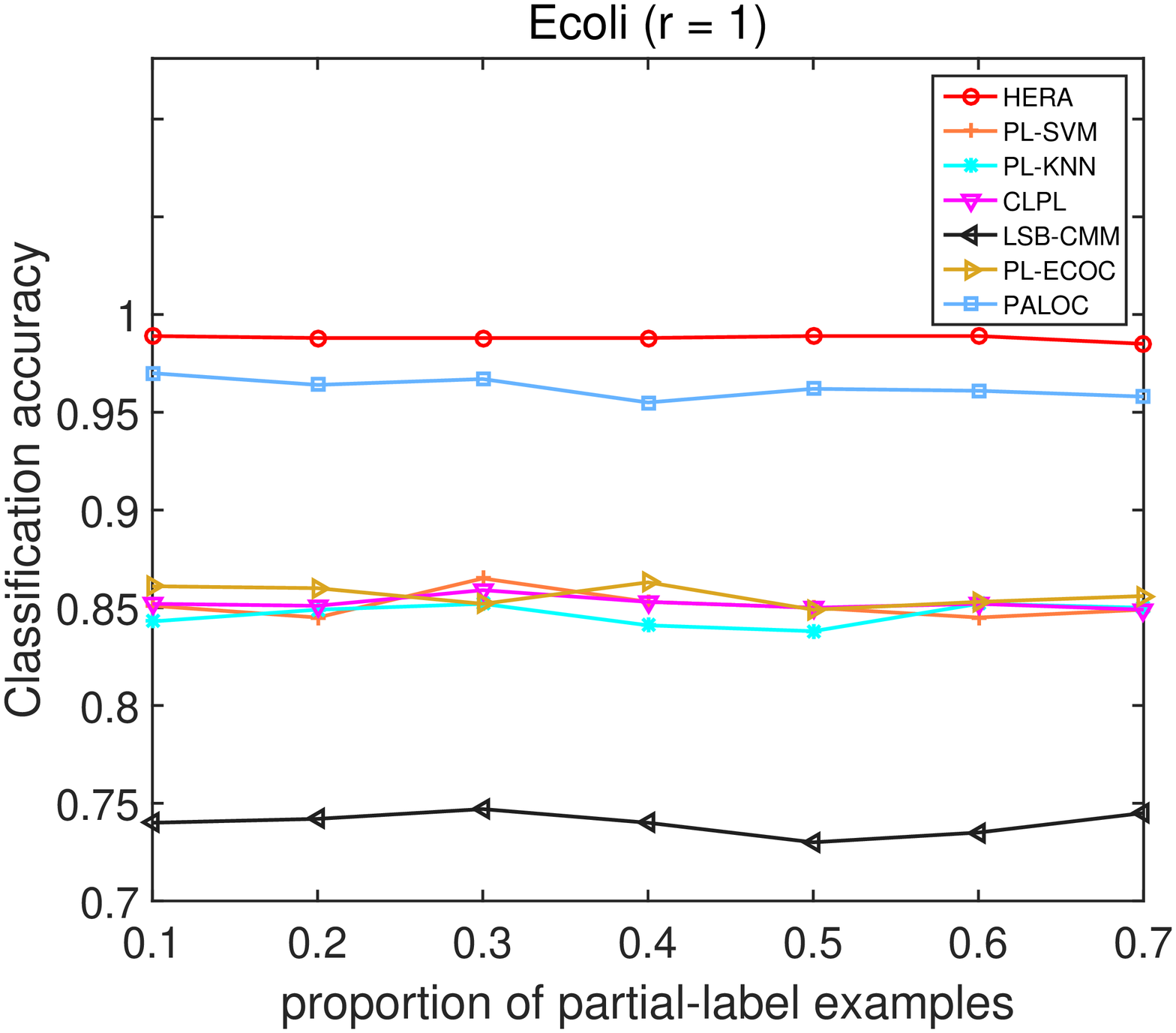}\\
\includegraphics[width = 1.77in,height=1.3in]{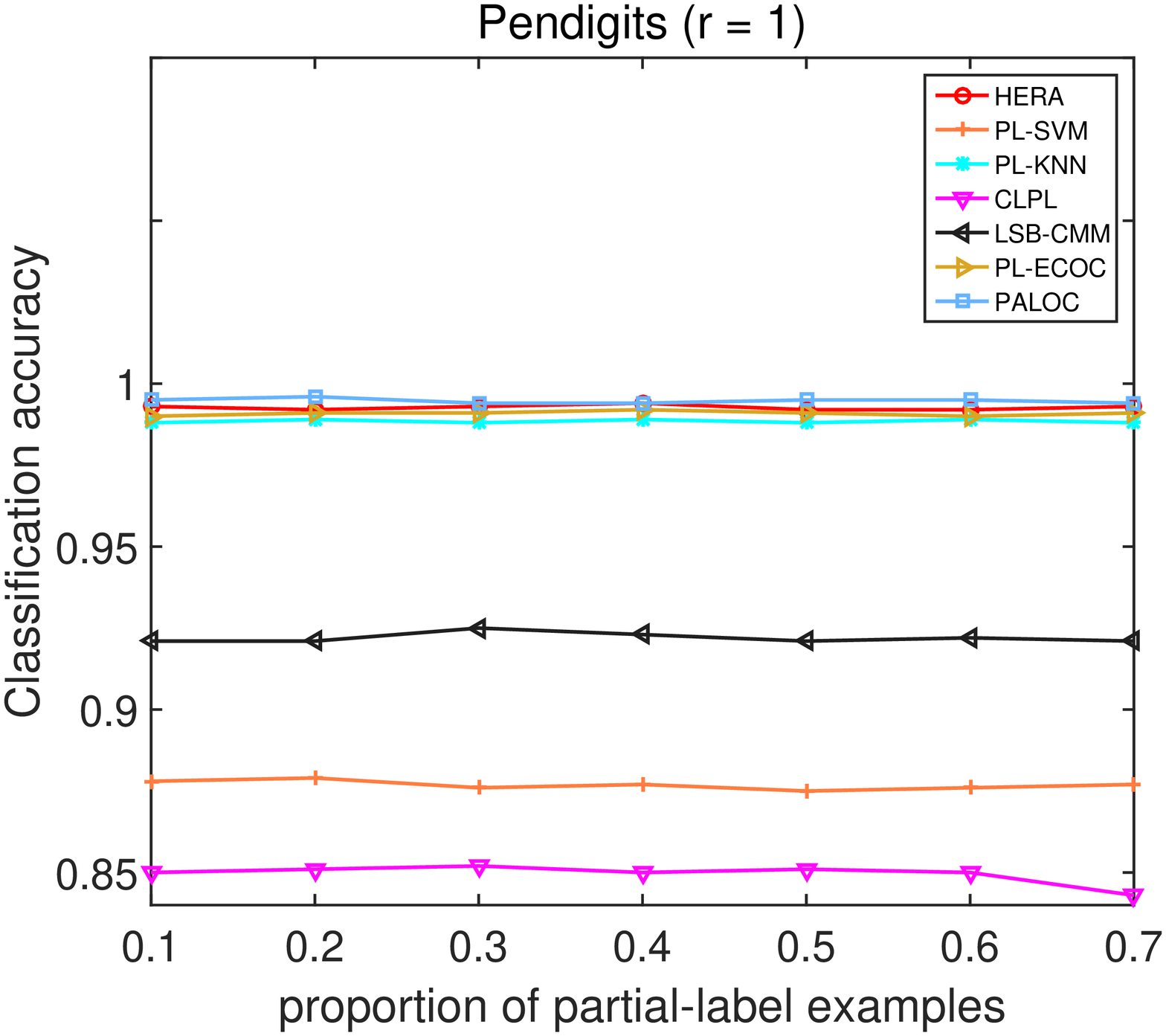}&\includegraphics[width = 1.77in,height=1.3in]{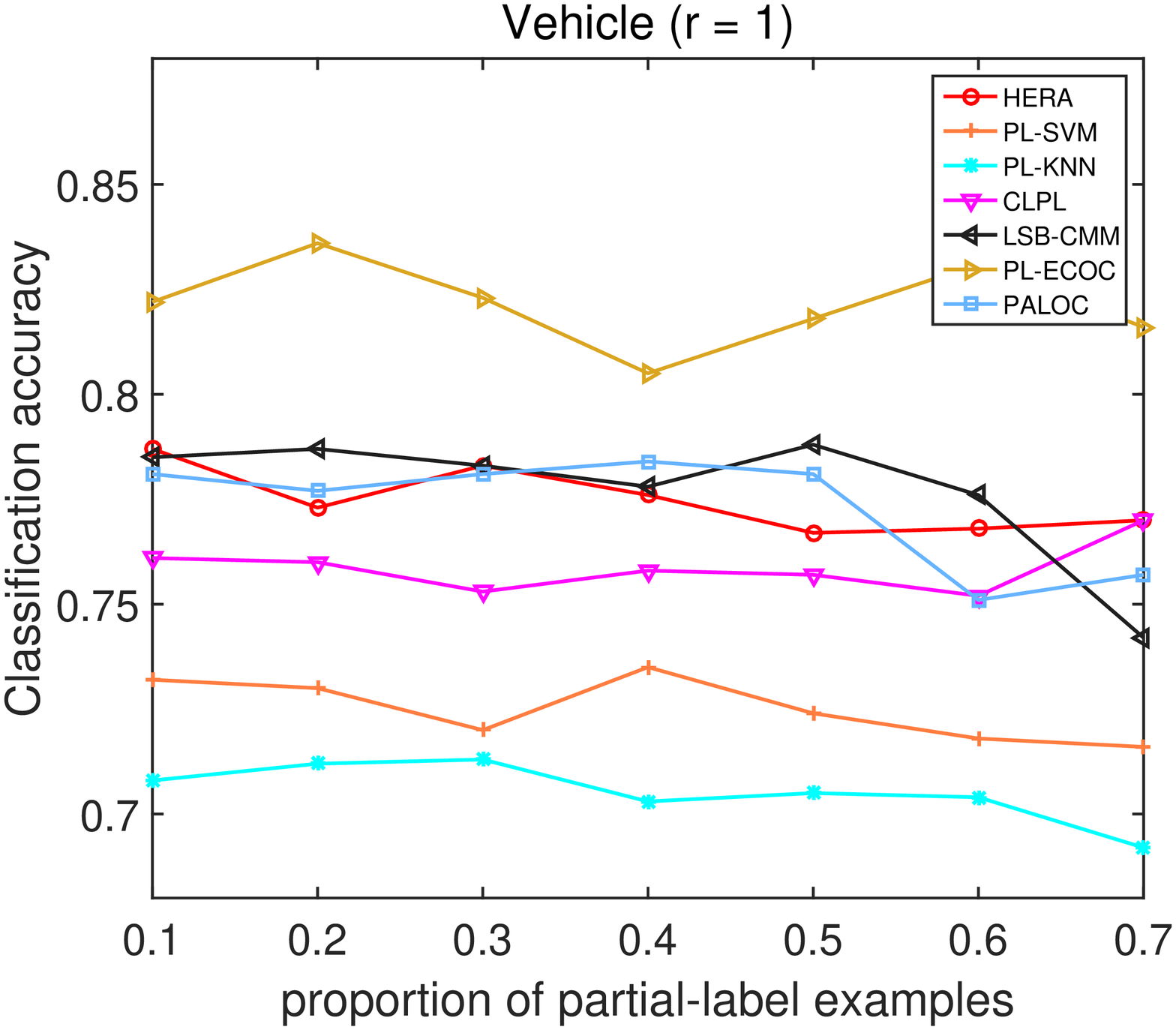}&\includegraphics[width = 1.778in,height=1.3in]{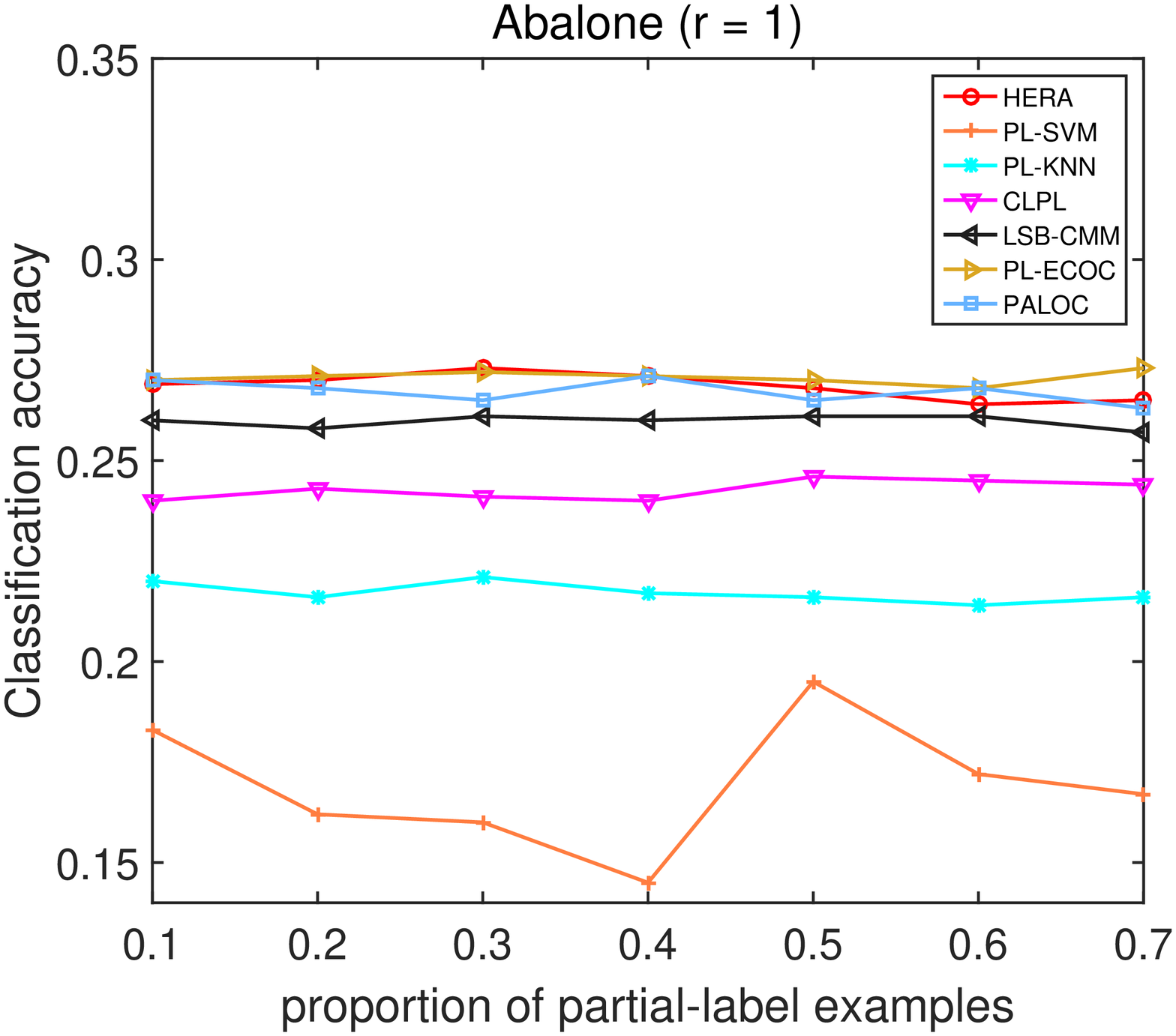}\\
\includegraphics[width = 1.77in,height=1.3in]{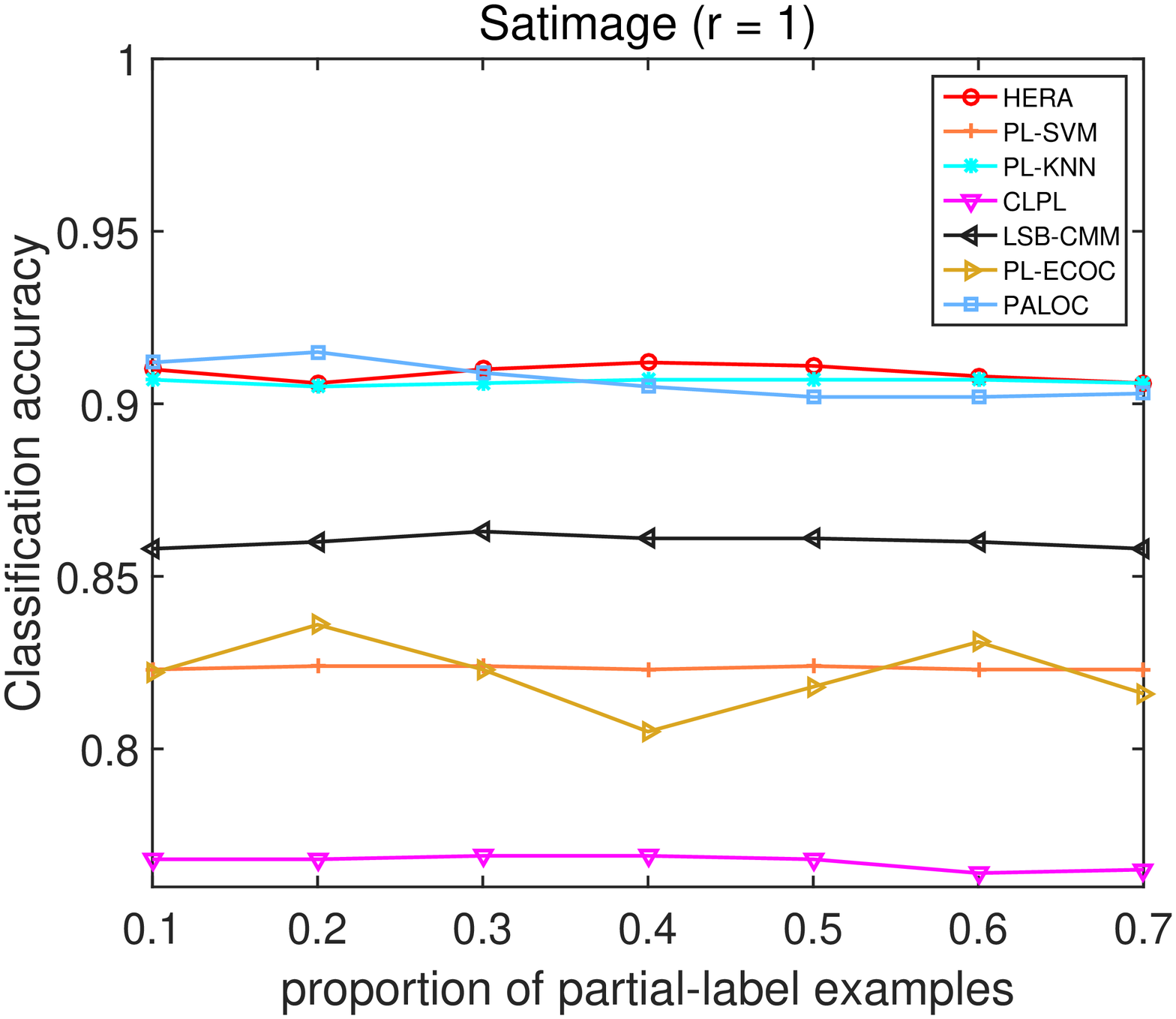}&\includegraphics[width = 1.77in,height=1.3in]{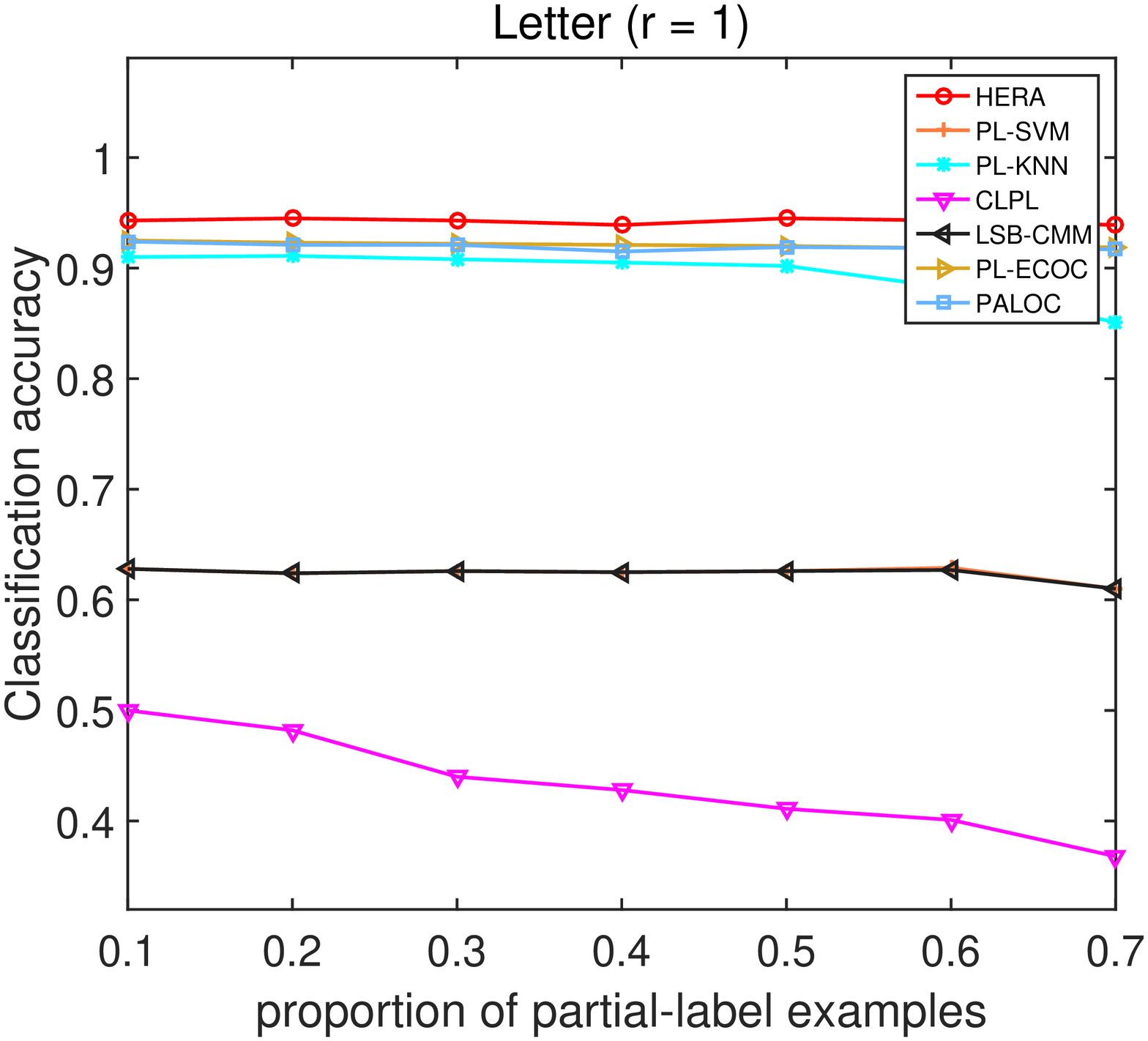}&\includegraphics[width = 1.77in,height=1.3in]{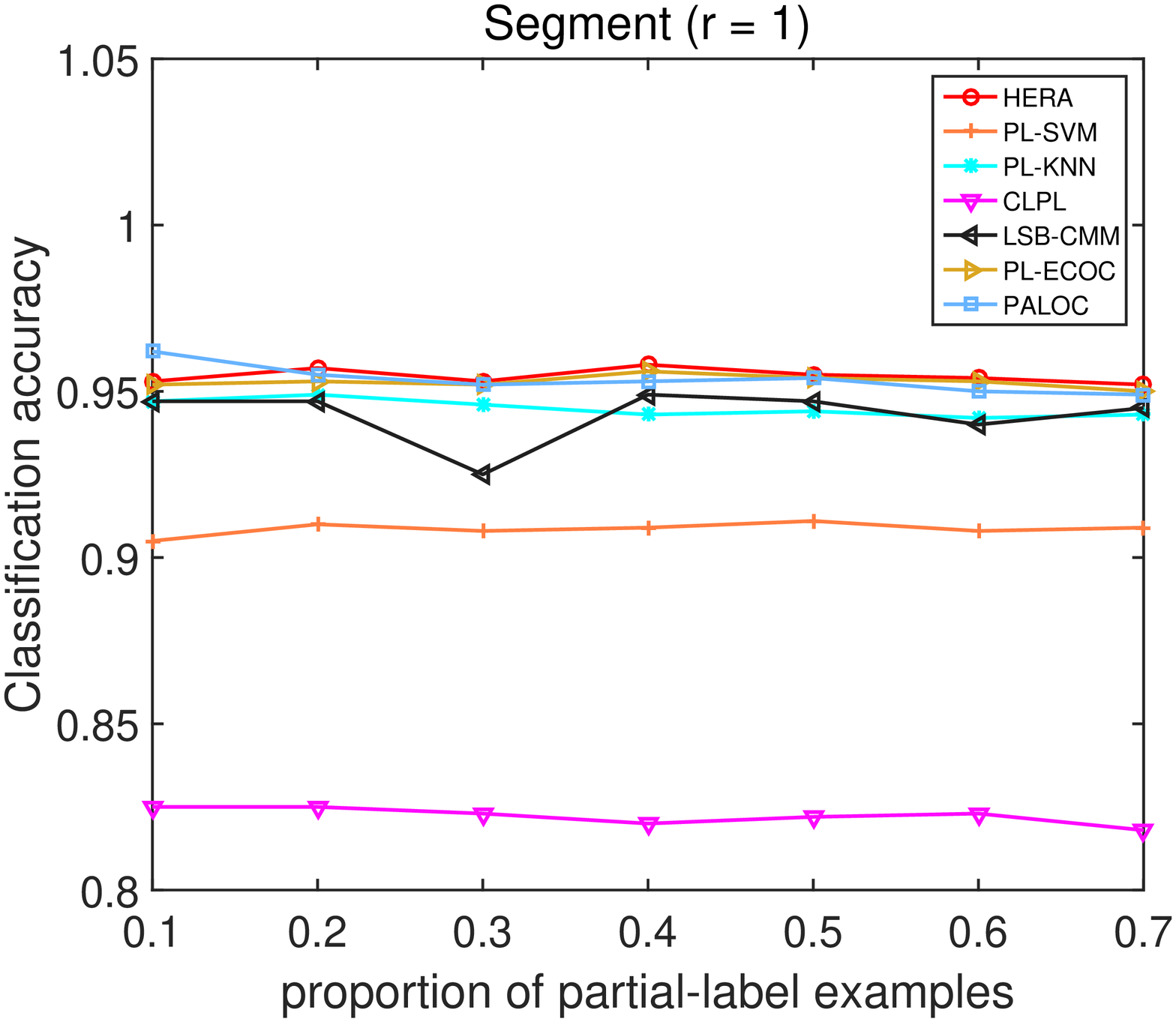}\\
\end{tabular}
\caption{The classification accuracy of several comparing methods on nine controlled UCI data sets changes as $p$ (proportion of partially labeled examples) increases from 0.1 to 0.7 (with one false candidate label [r = 1]).}
\label{fig-uci-1}
\vspace{0mm}
\end{figure*}

\begin{figure*}
\centering
\begin{tabular}{ccc}
\includegraphics[width = 1.77in,height=1.3in]{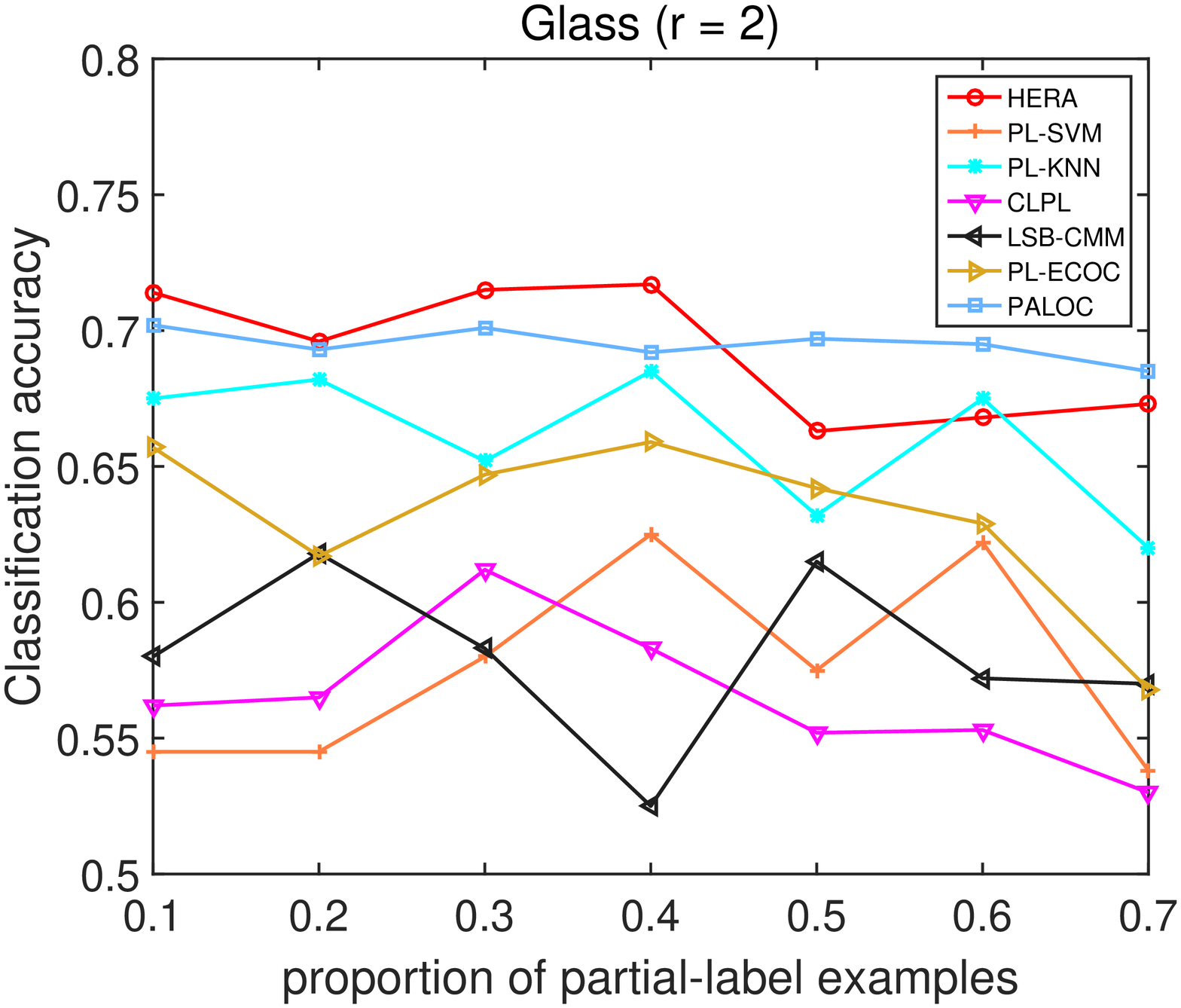}&\includegraphics[width = 1.77in,height=1.3in]{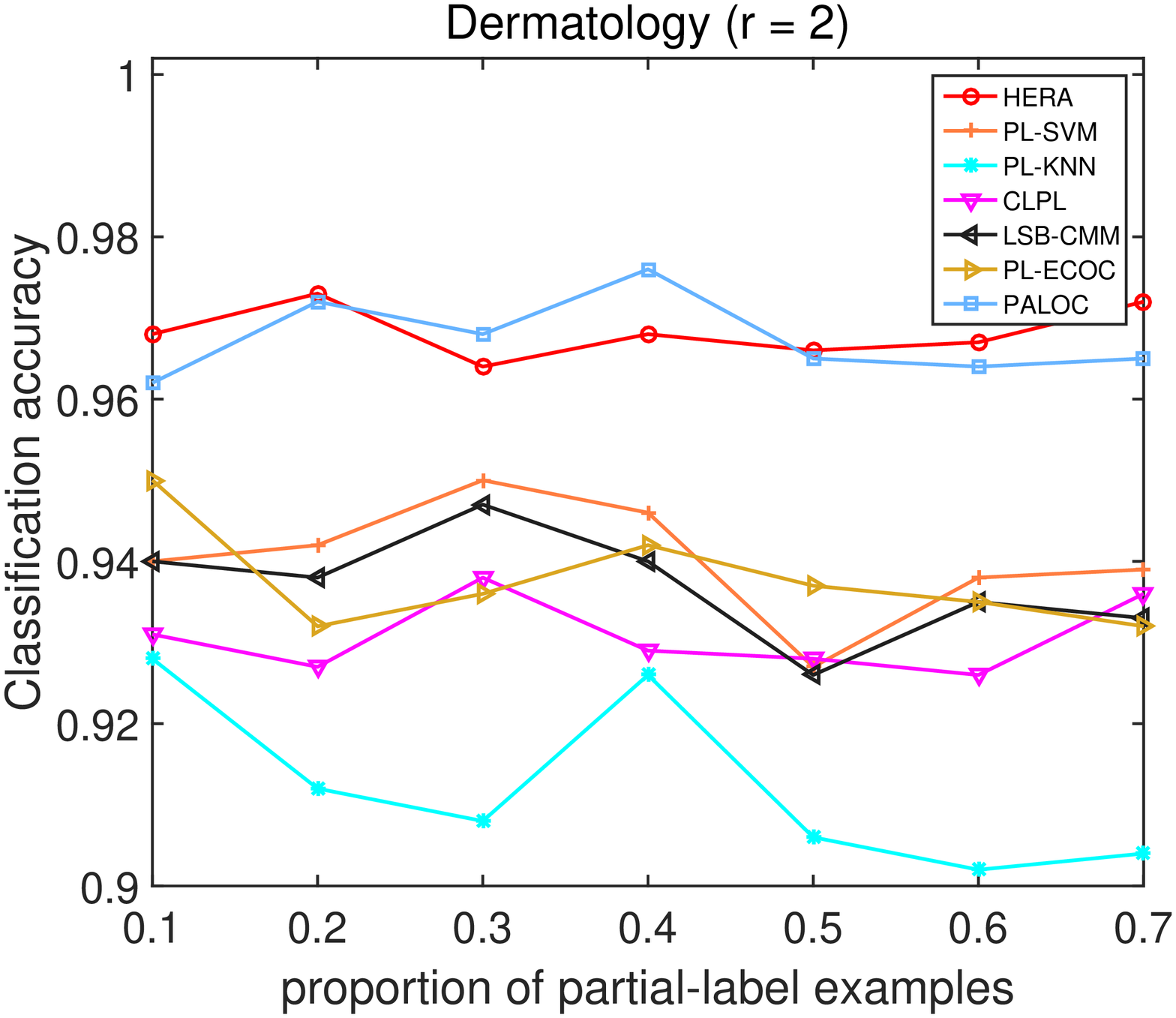}&\includegraphics[width = 1.77in,height=1.3in]{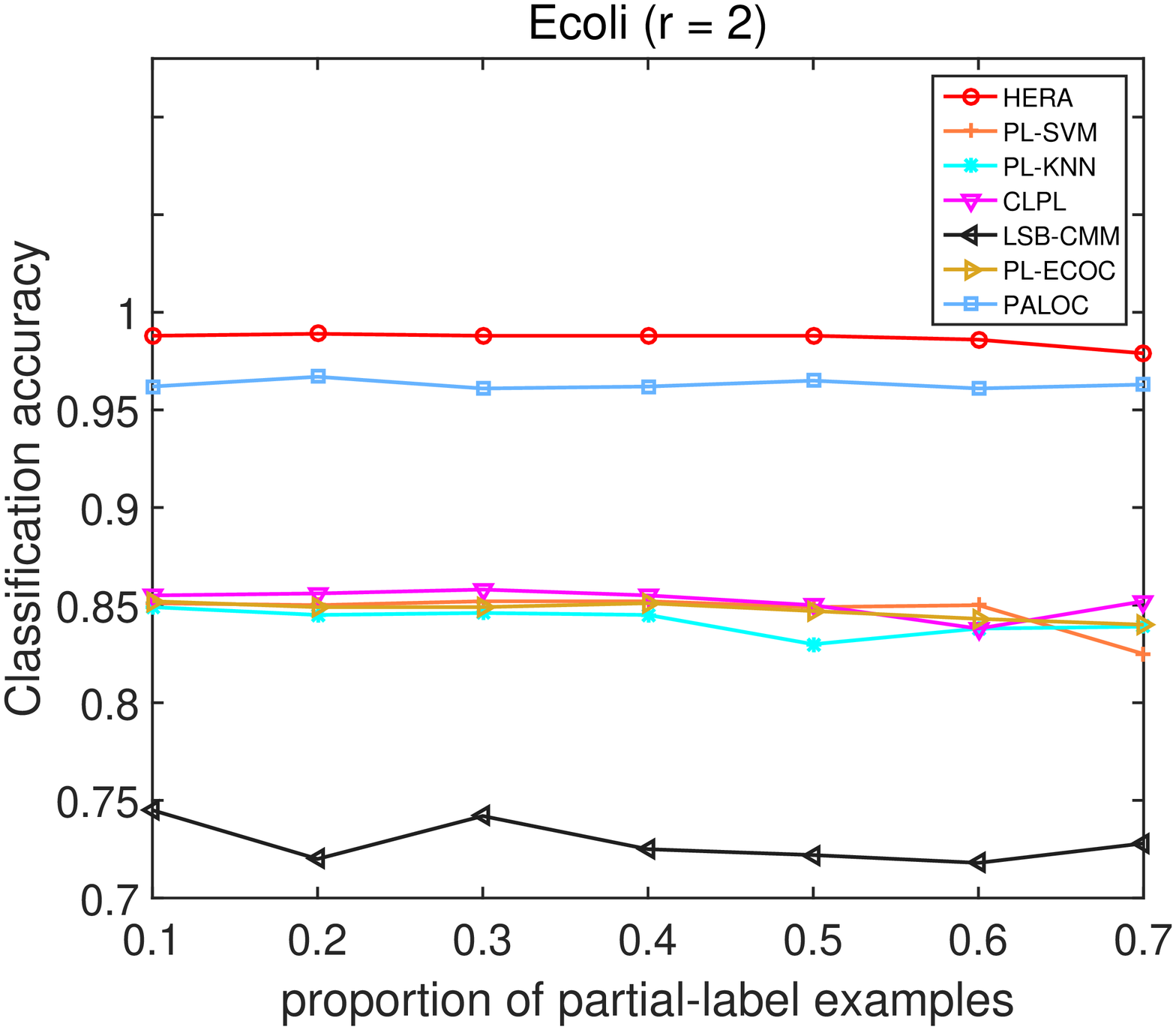}\\
\includegraphics[width = 1.77in,height=1.3in]{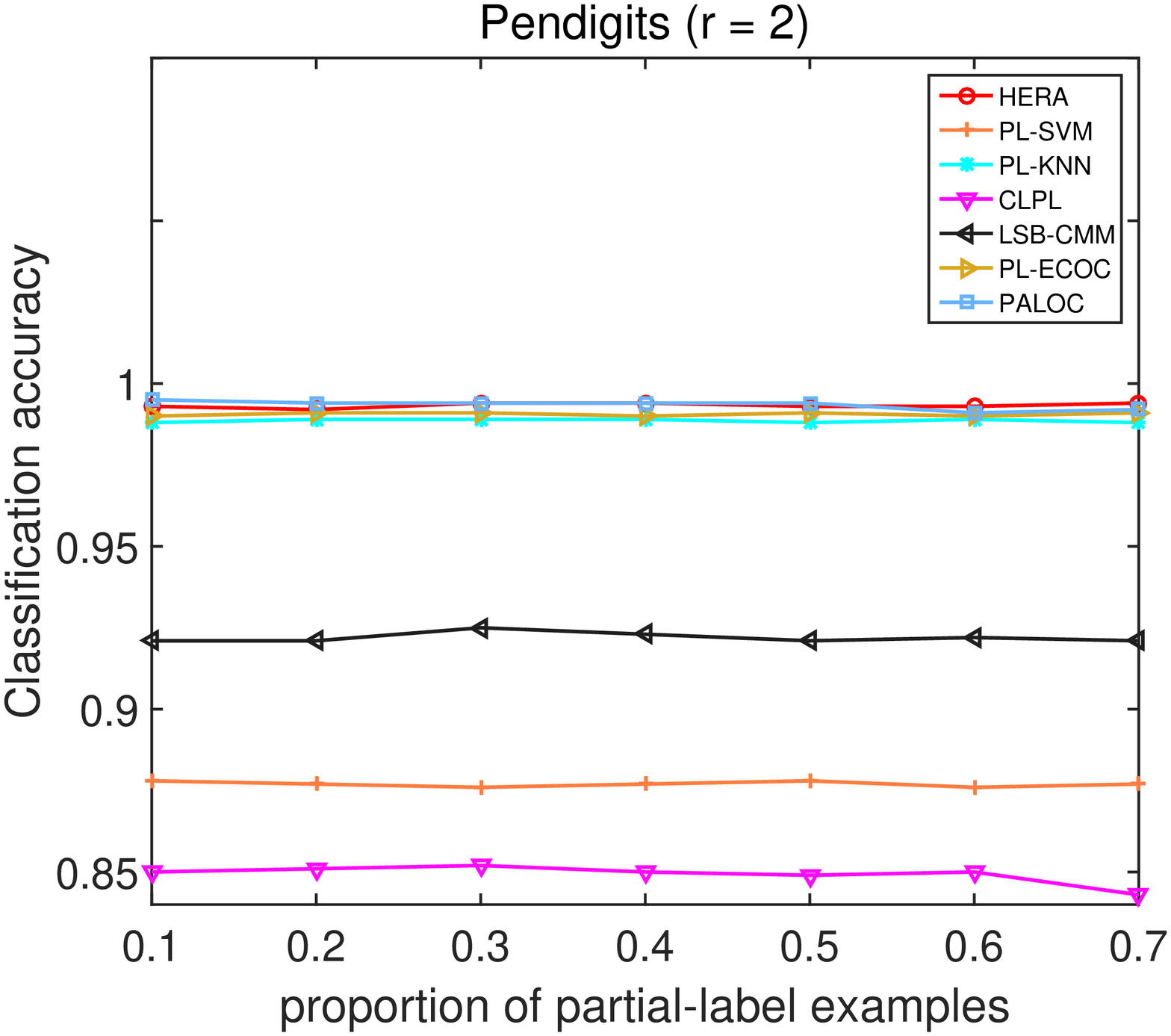}&\includegraphics[width = 1.77in,height=1.3in]{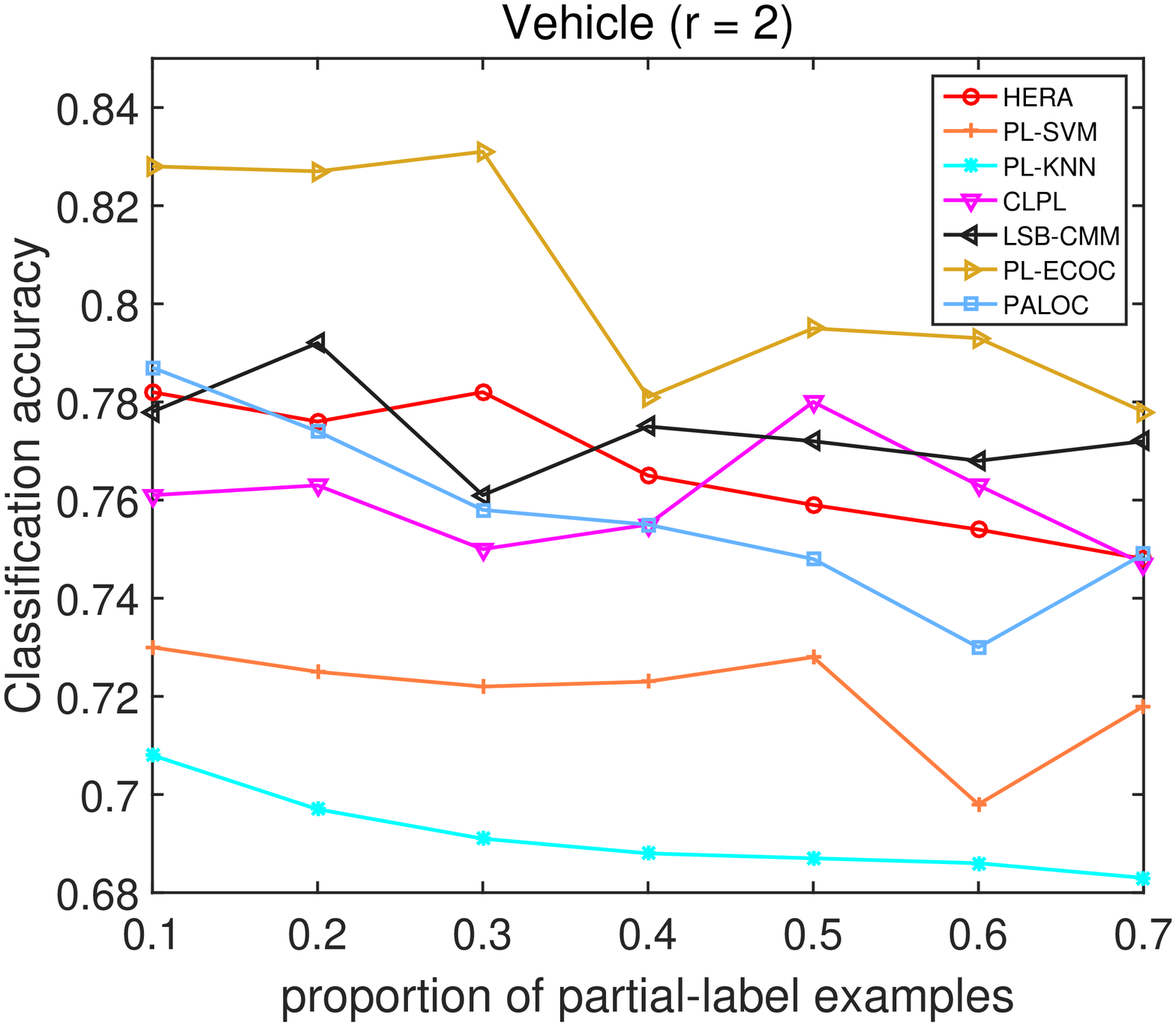}&\includegraphics[width = 1.77in,height=1.3in]{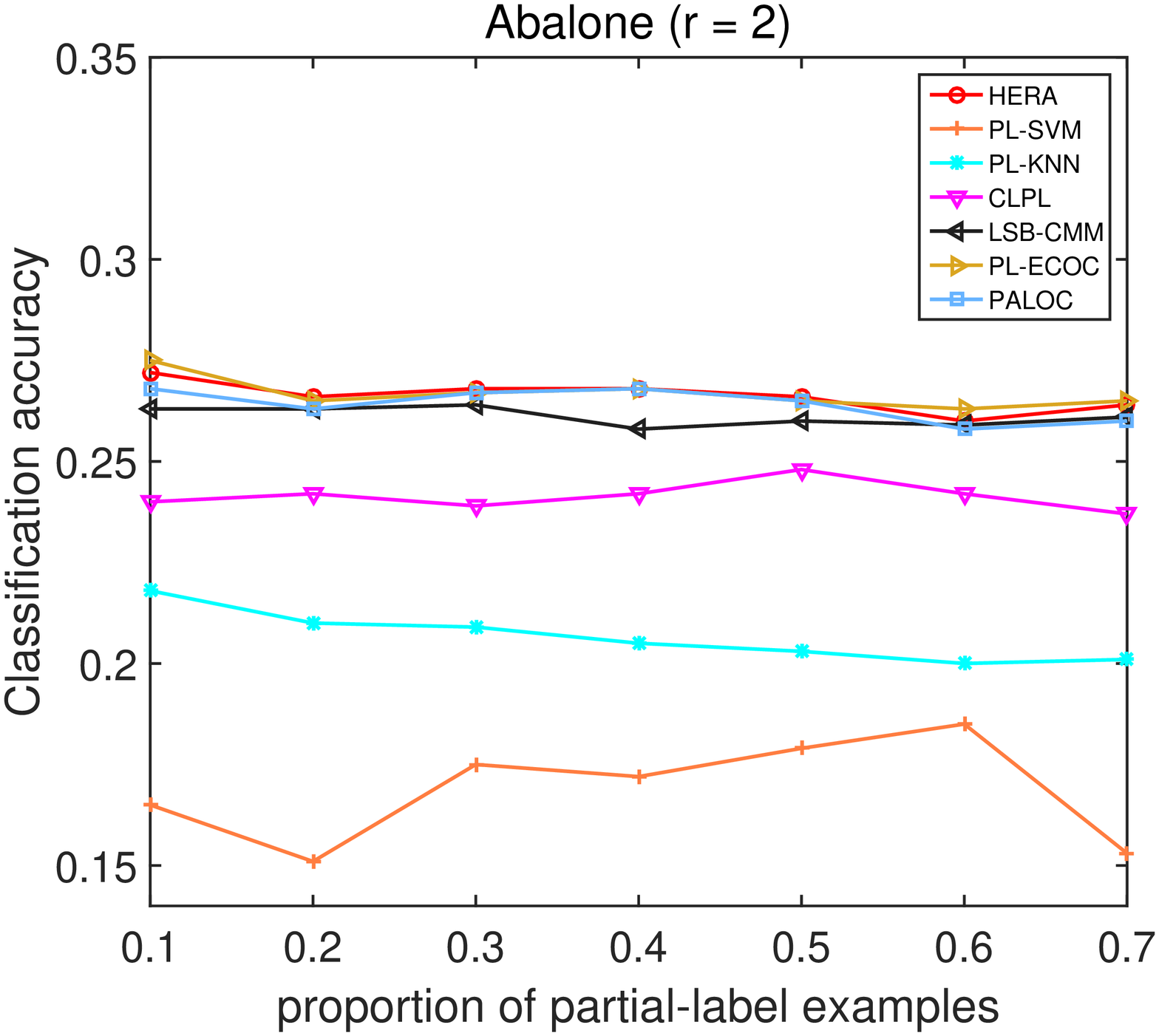}\\
\includegraphics[width = 1.77in,height=1.3in]{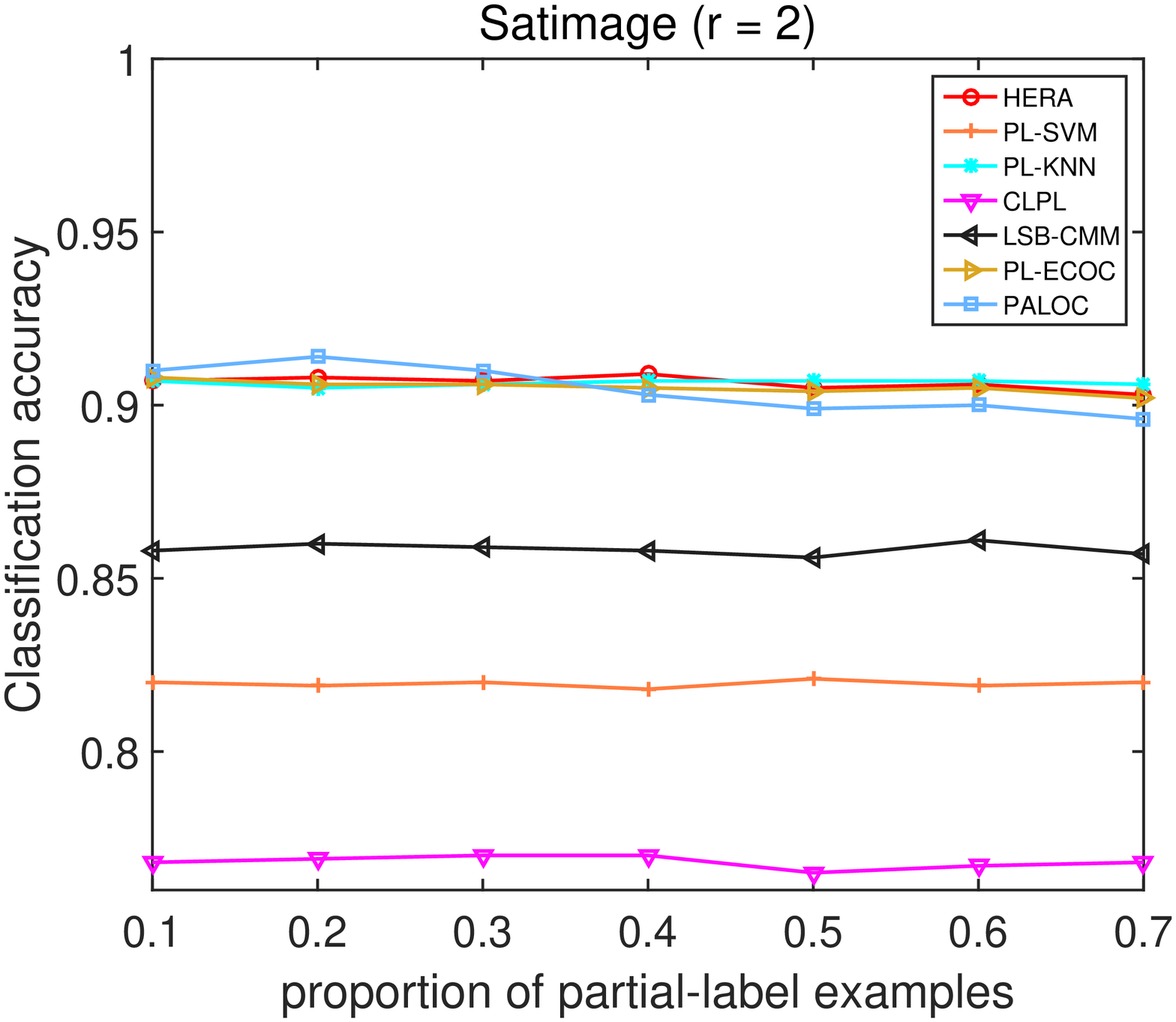}&\includegraphics[width = 1.77in,height=1.3in]{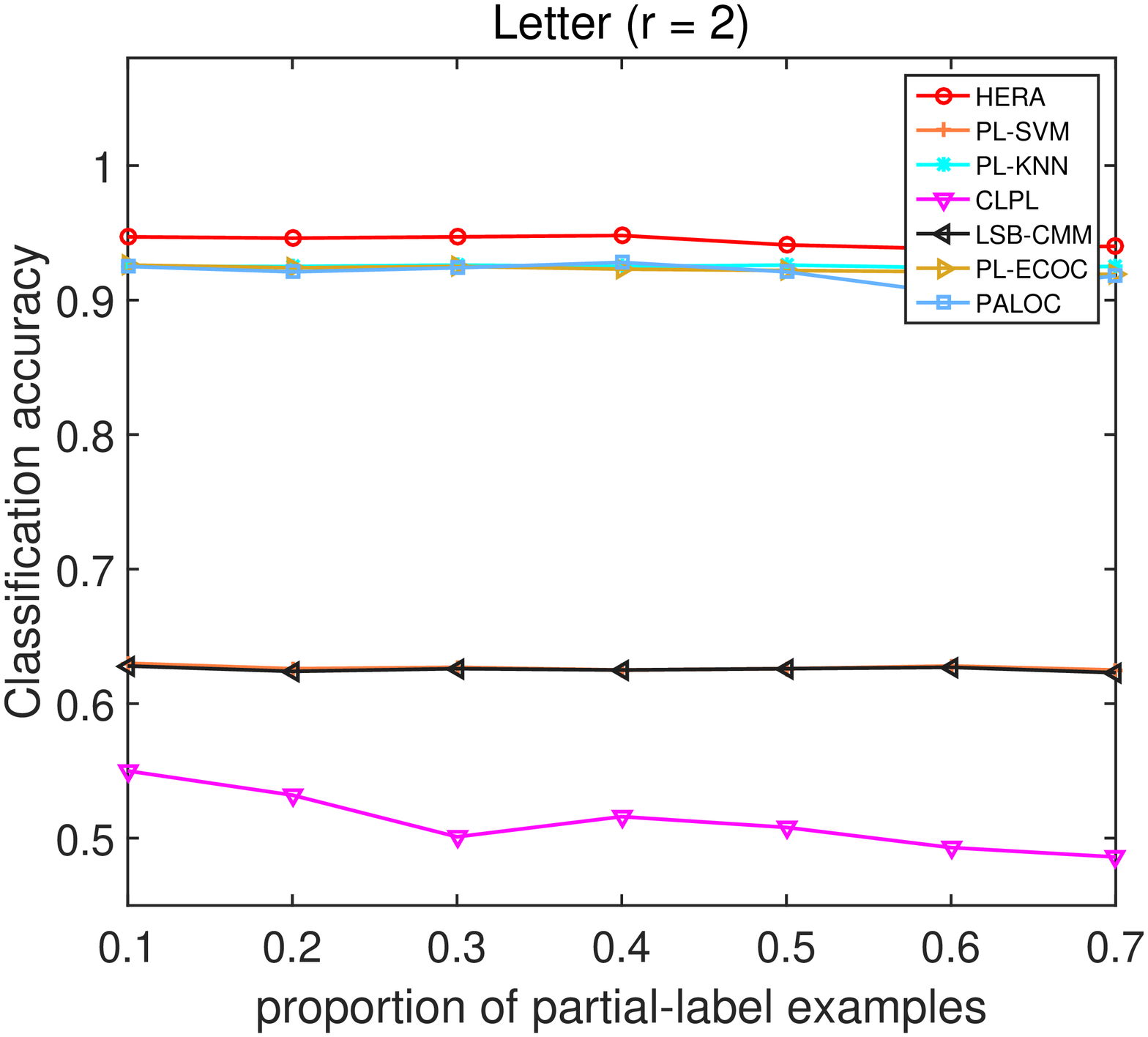}&\includegraphics[width = 1.77in,height=1.3in]{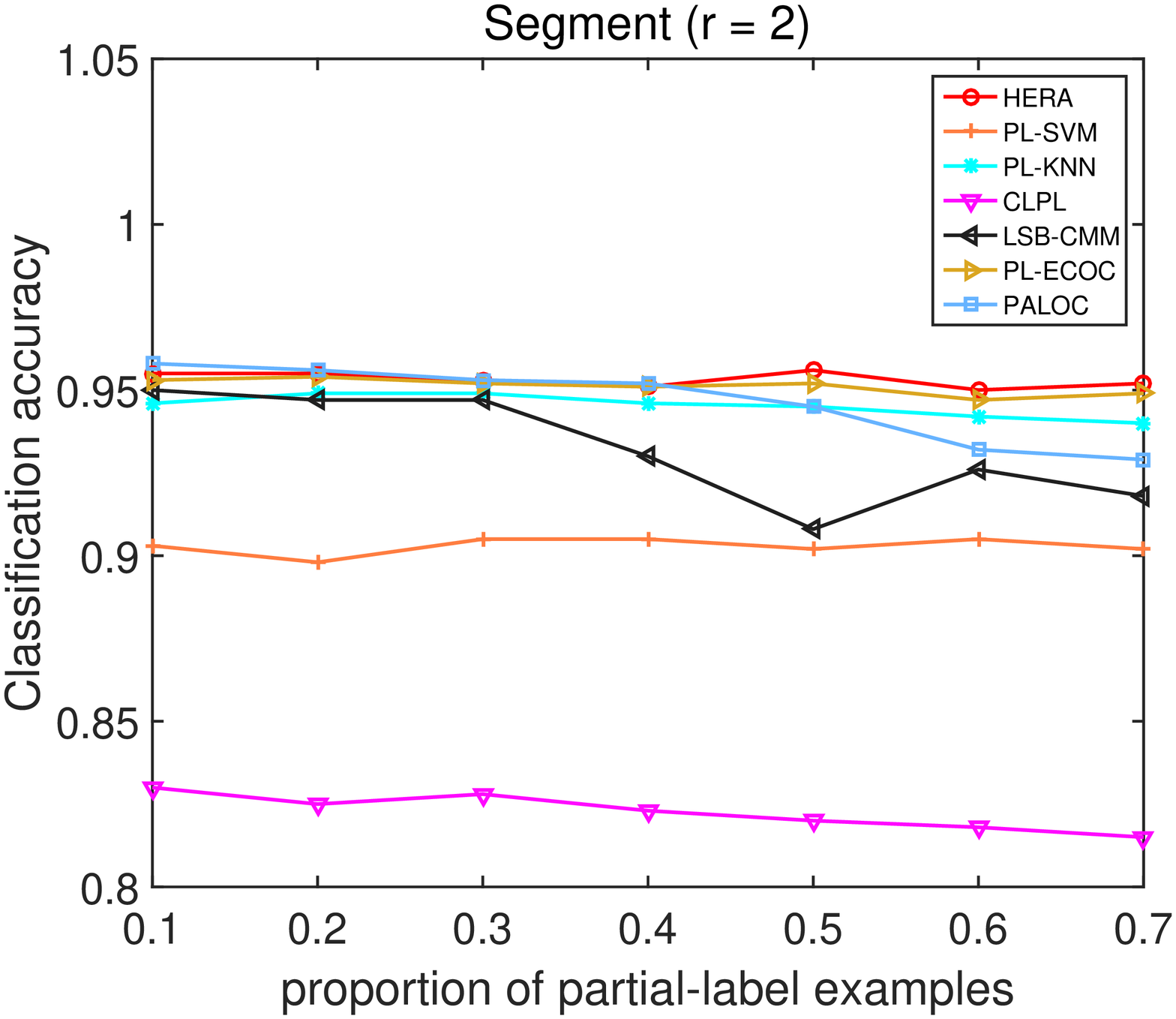}\\
\end{tabular}
\caption{The classification accuracy of several comparing methods on nine controlled UCI data sets changes as $p$ (proportion of partially labeled examples) increases from 0.1 to 0.7 (with two false candidate labels [r = 2]).}
\label{fig-uci-2}
\vspace{0mm}
\end{figure*}

\begin{figure*}
\centering
\begin{tabular}{ccc}
\includegraphics[width = 1.77in,height=1.3in]{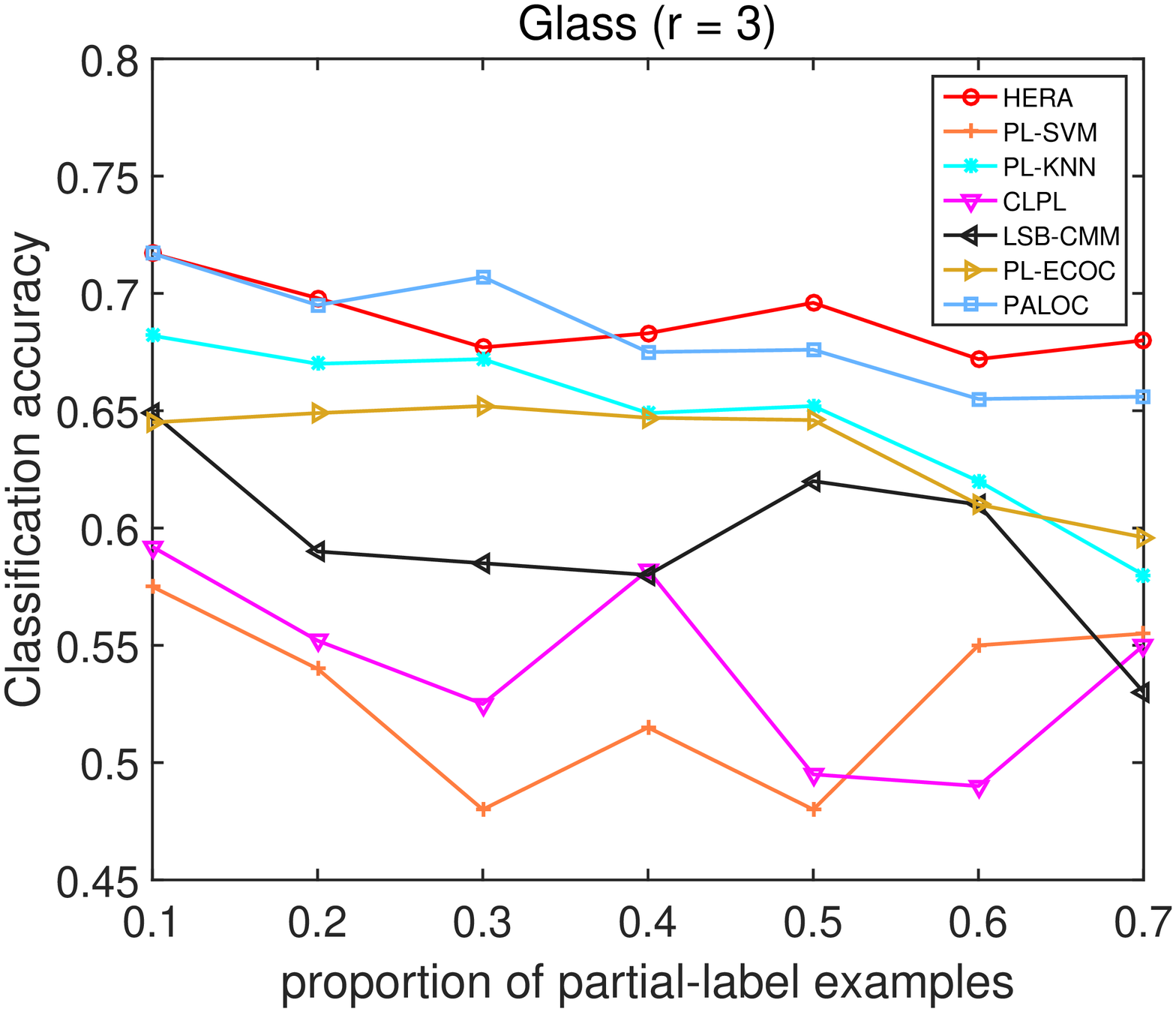}&\includegraphics[width = 1.77in,height=1.3in]{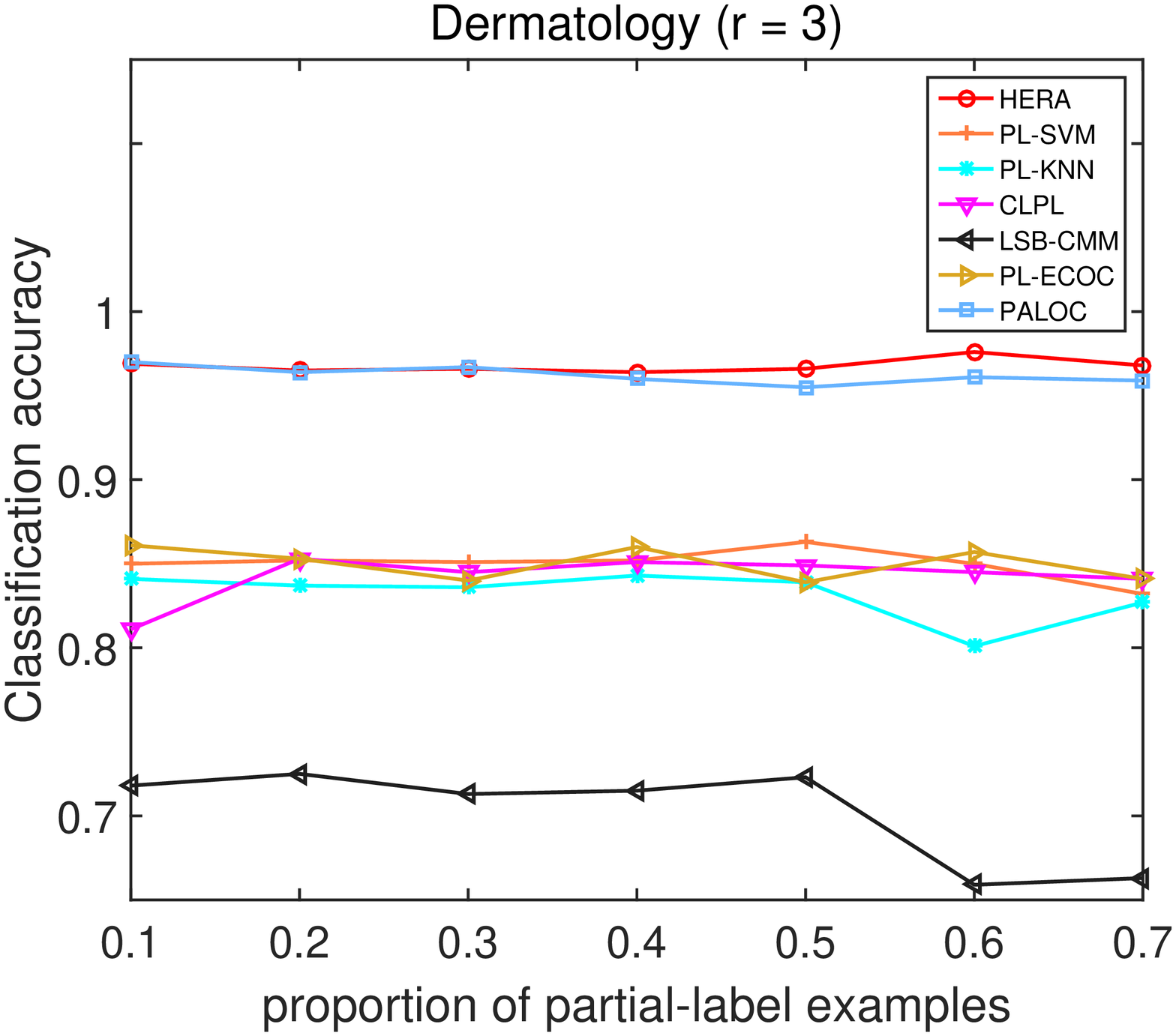}&\includegraphics[width = 1.77in,height=1.3in]{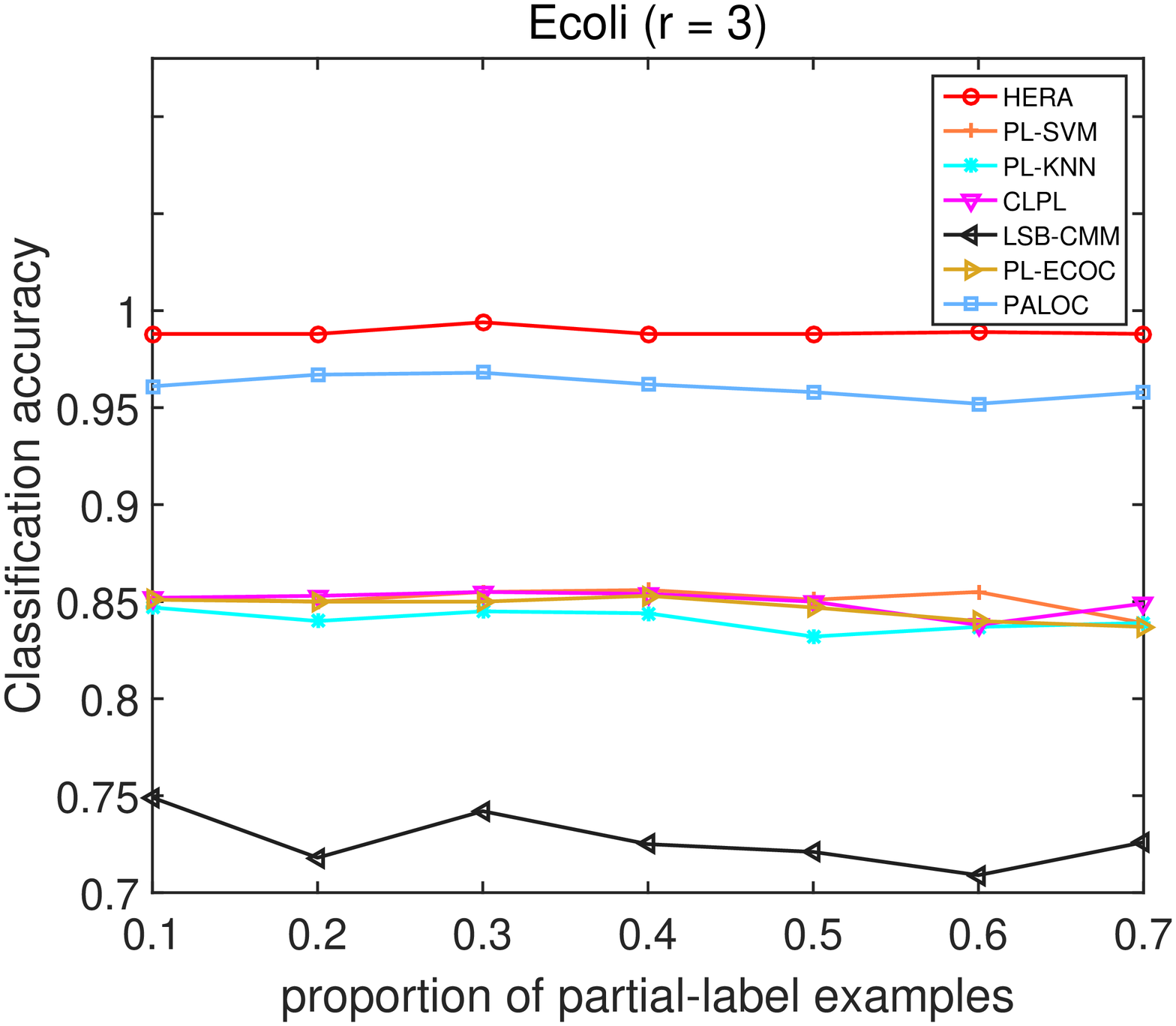}\\
\includegraphics[width = 1.77in,height=1.3in]{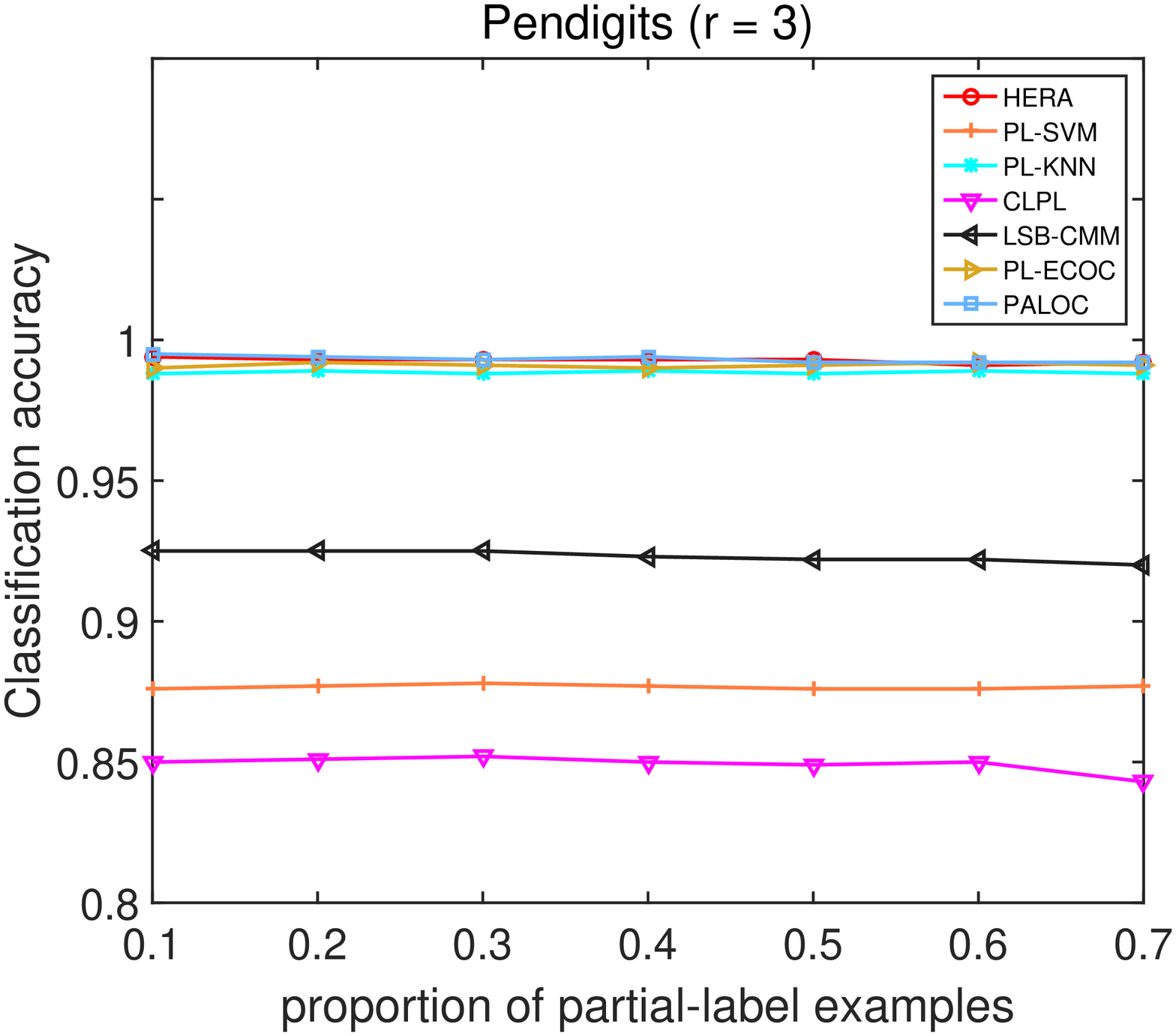}&\includegraphics[width = 1.77in,height=1.3in]{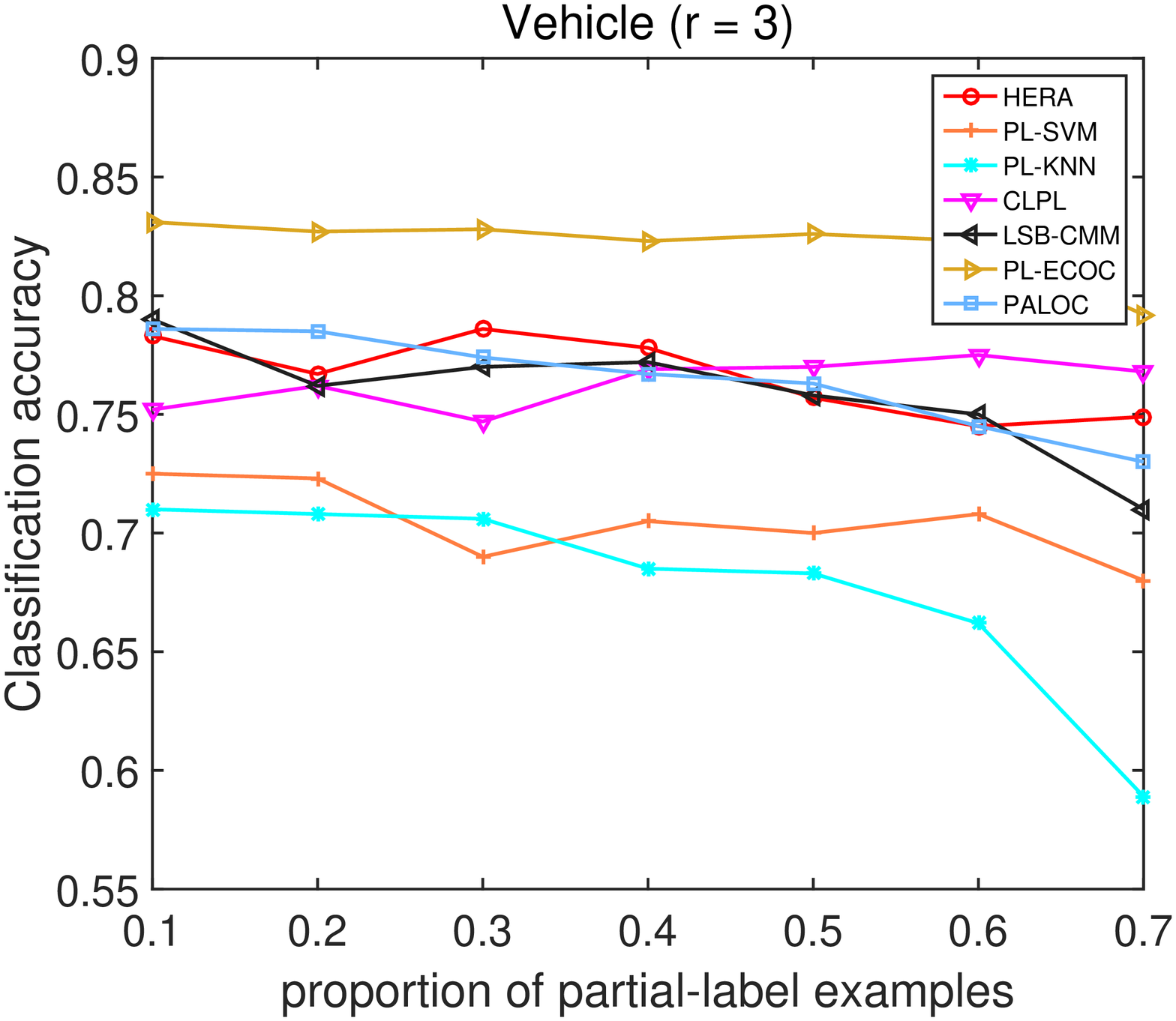}&\includegraphics[width = 1.77in,height=1.3in]{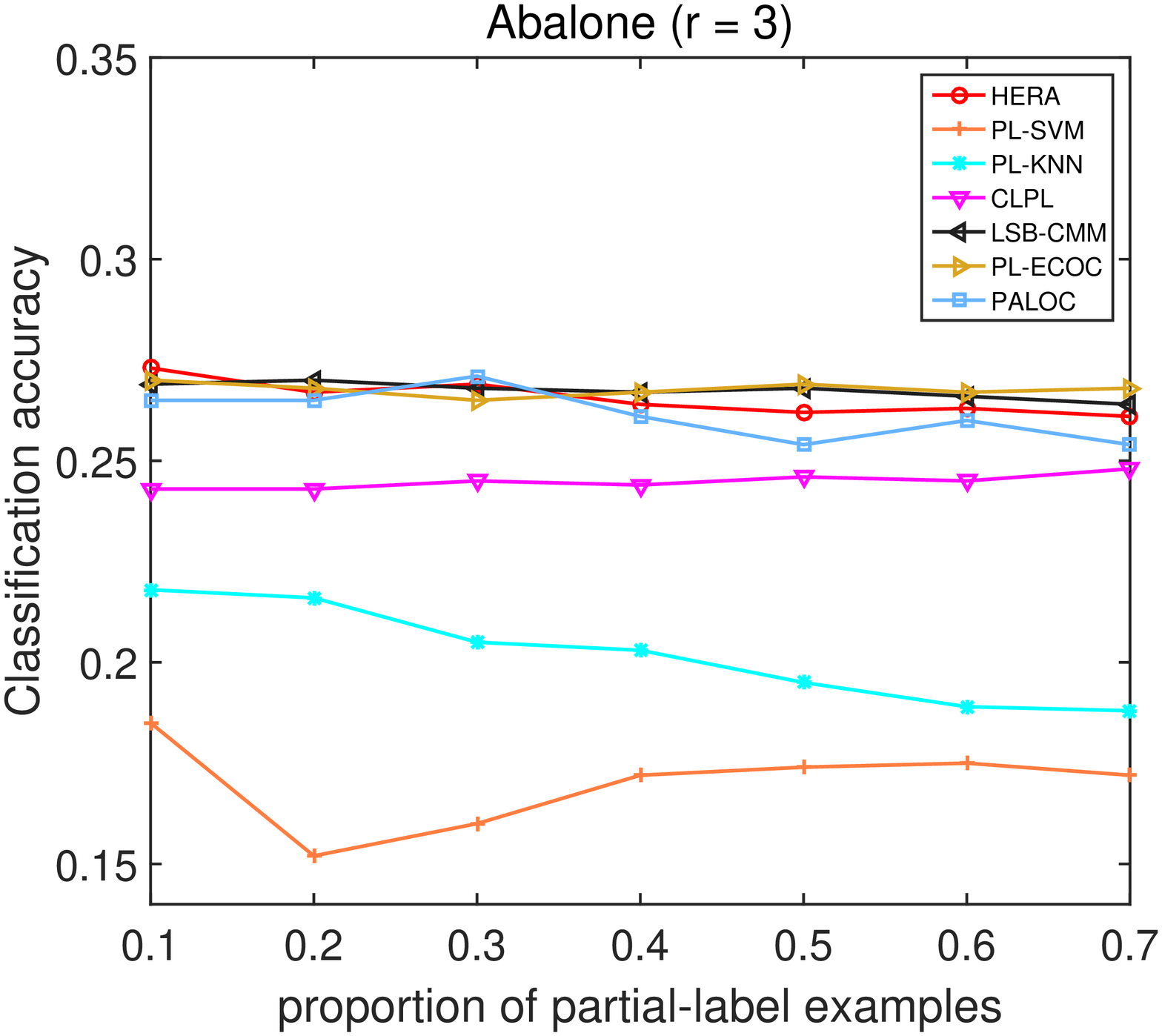}\\
\includegraphics[width = 1.77in,height=1.3in]{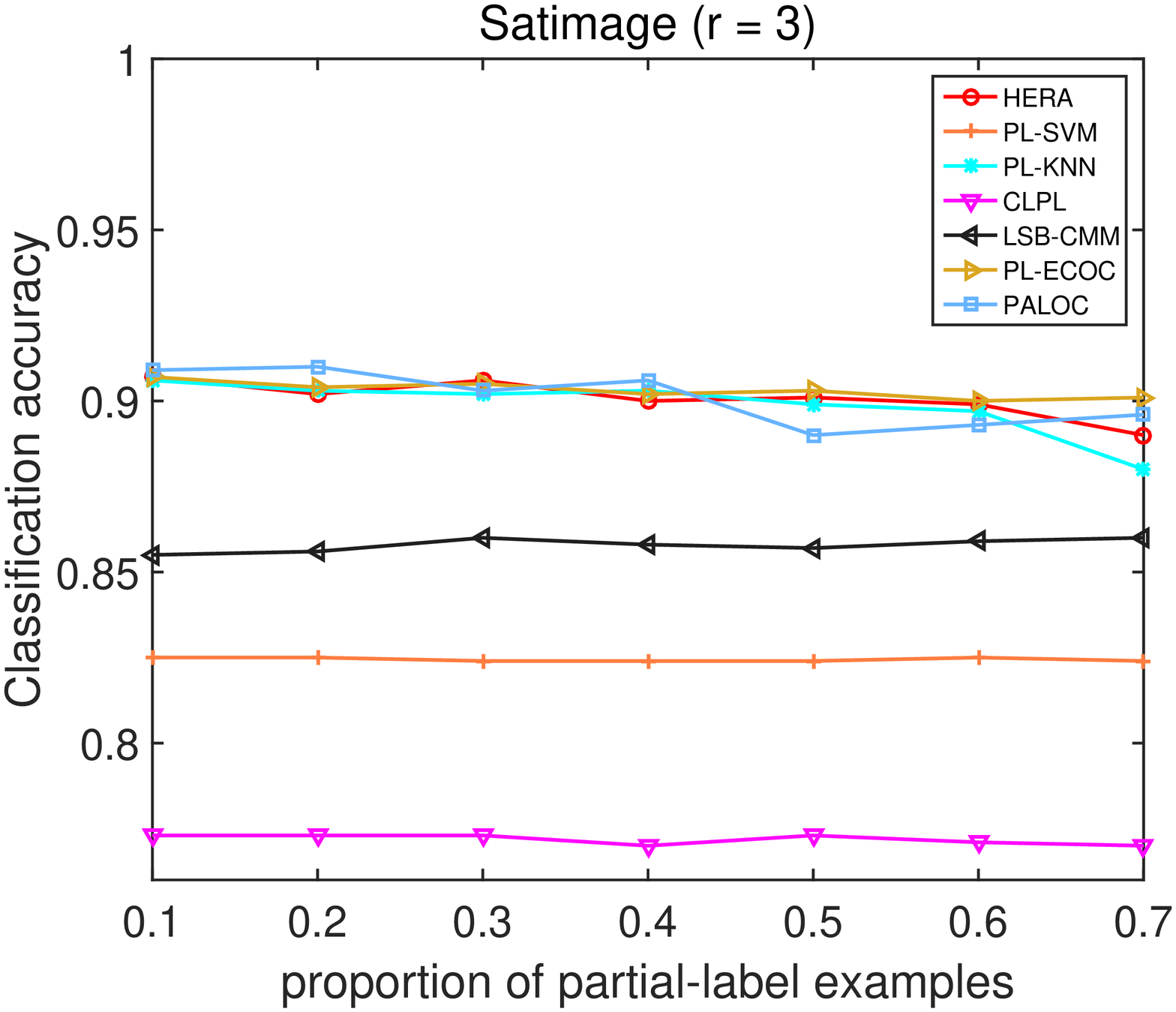}&\includegraphics[width = 1.77in,height=1.3in]{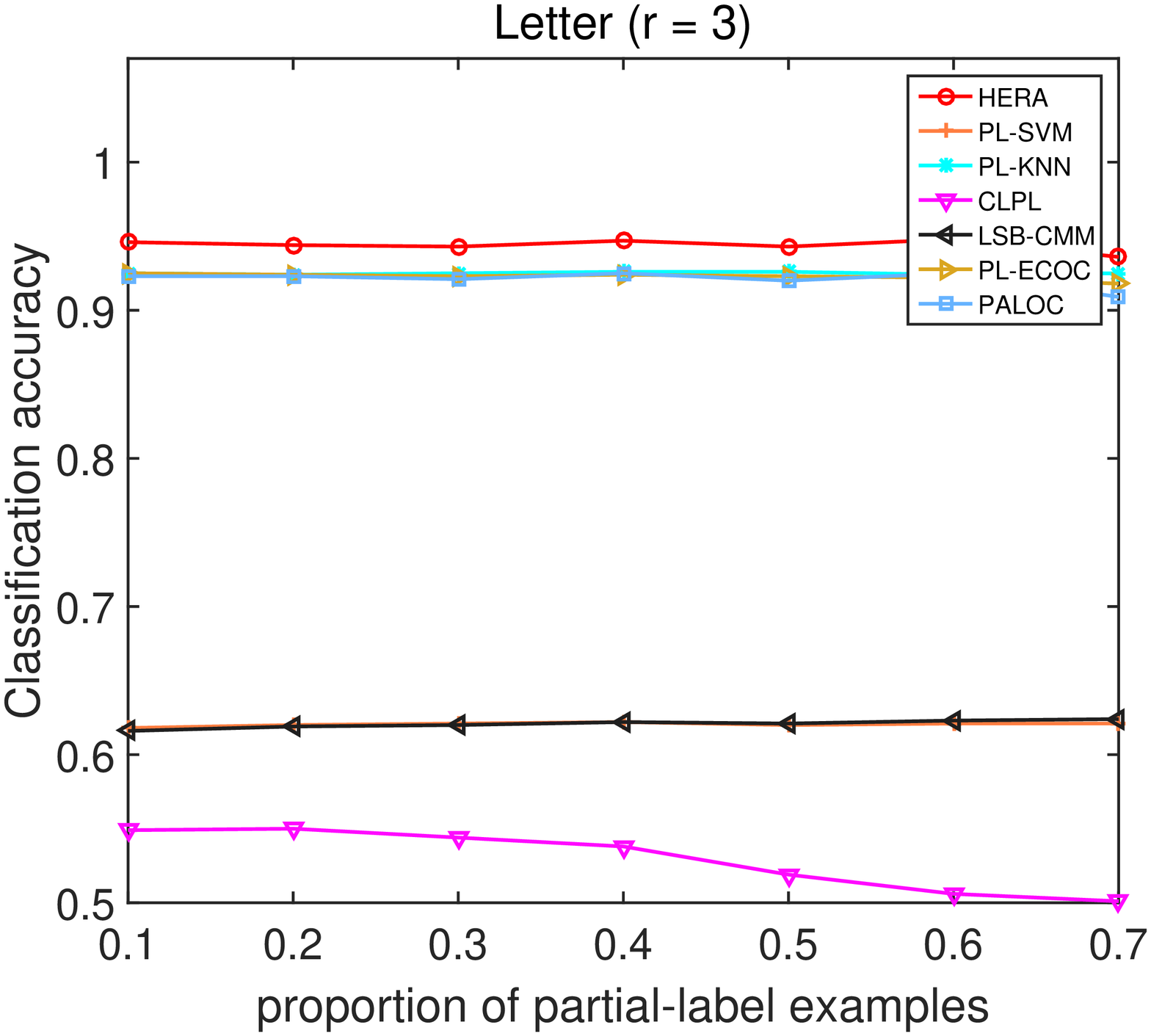}&\includegraphics[width = 1.77in,height=1.3in]{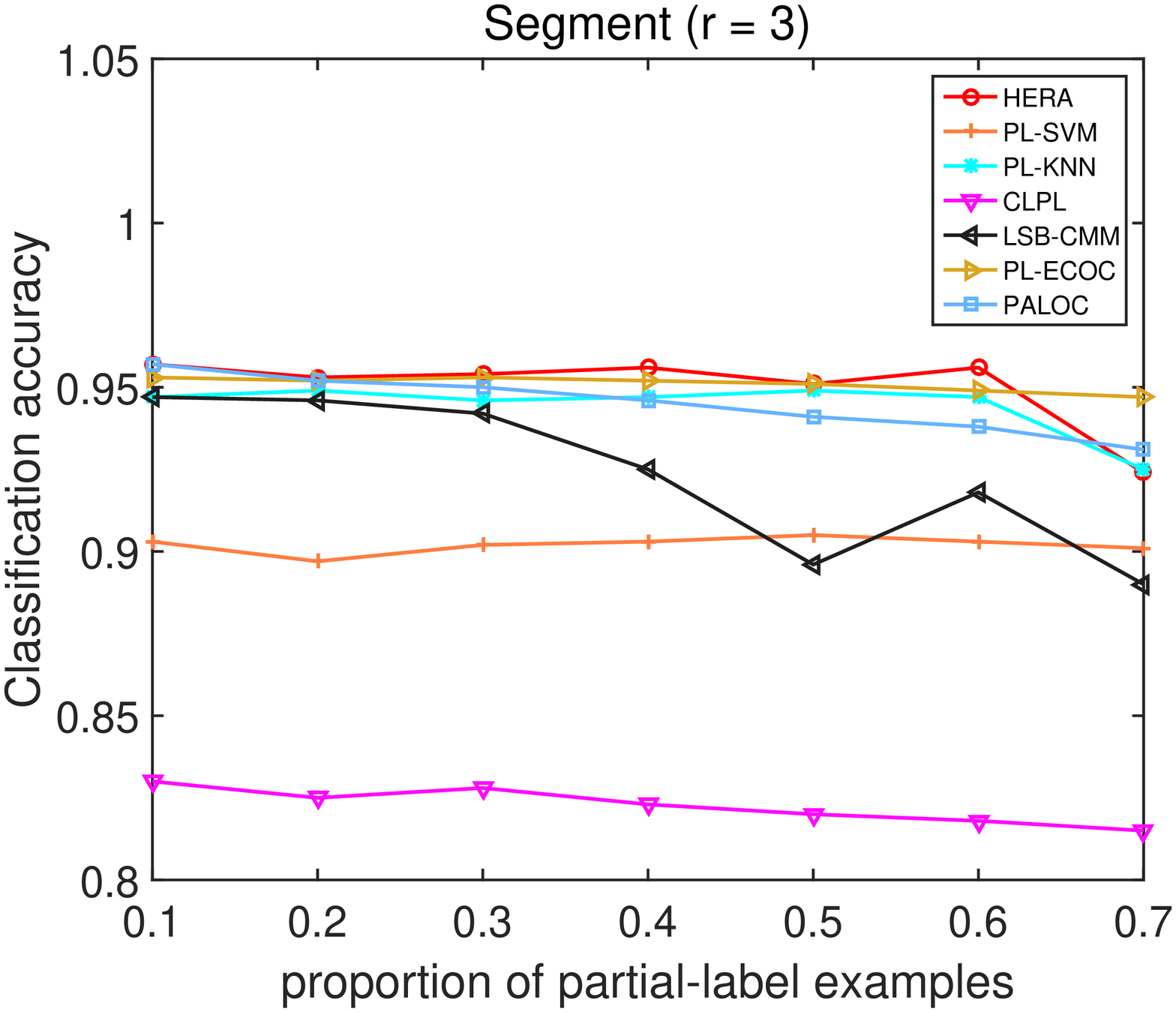}\\
\end{tabular}
\caption{The classification accuracy of several comparing methods on nine controlled UCI data sets changes as $p$ (proportion of partially labeled examples) increases from 0.1 to 0.7 (with three false candidate labels [r = 3]).}
\label{fig-uci-3}
\vspace{0mm}
\end{figure*}

\begin{figure*}
\centering
\begin{tabular}{ccc}
\includegraphics[width = 1.77in,height=1.3in]{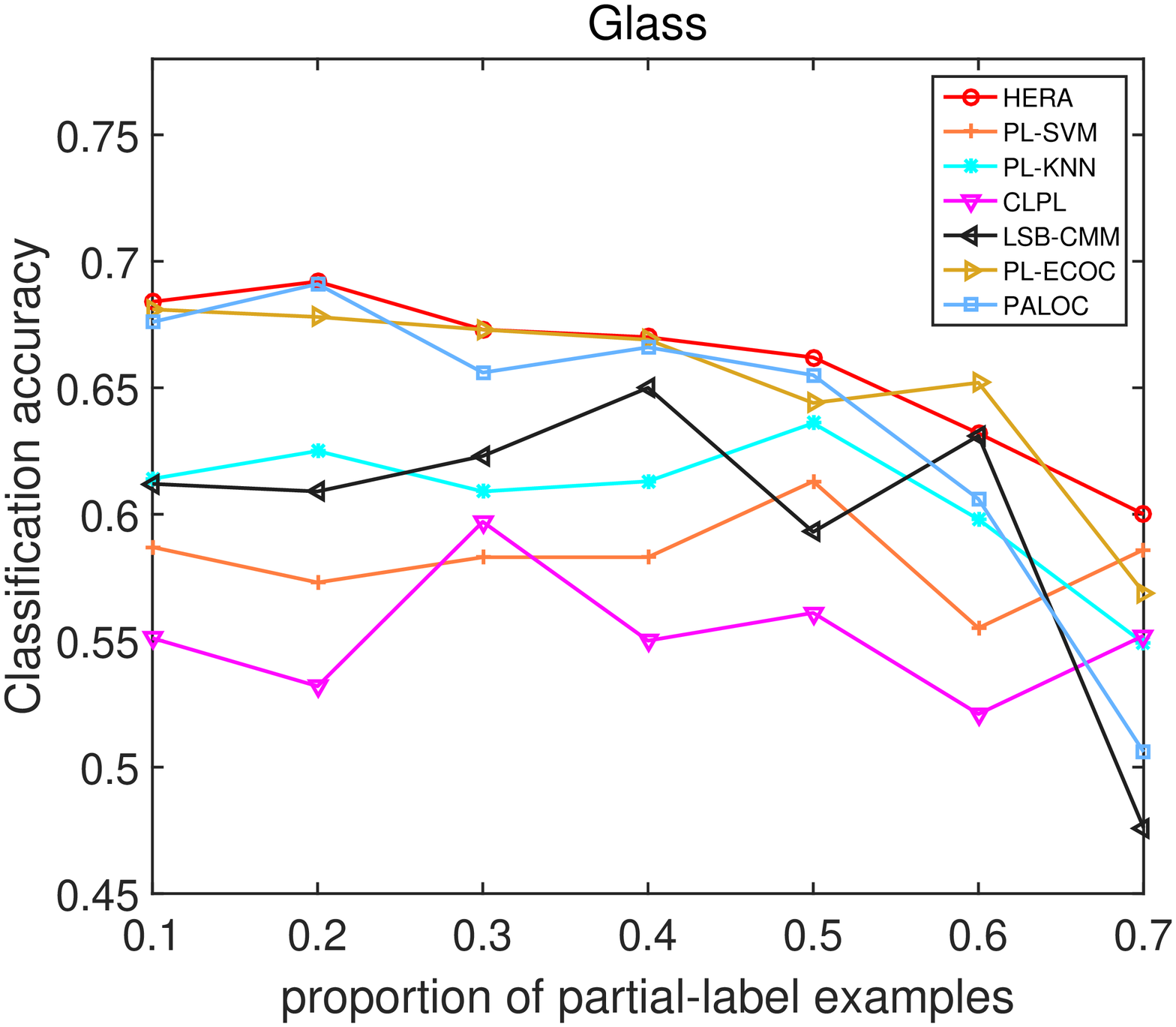}&\includegraphics[width = 1.77in,height=1.3in]{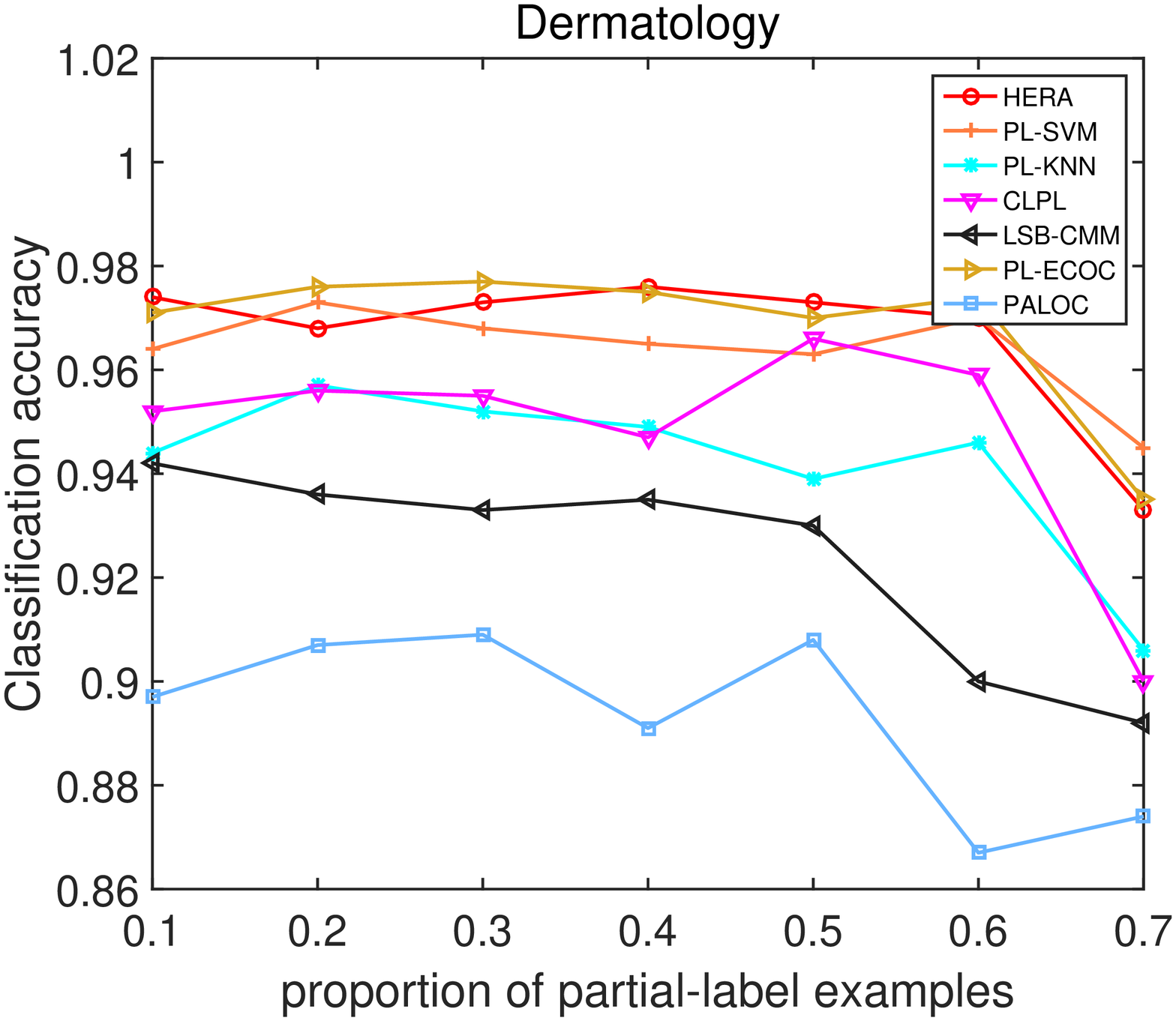}&\includegraphics[width = 1.77in,height=1.3in]{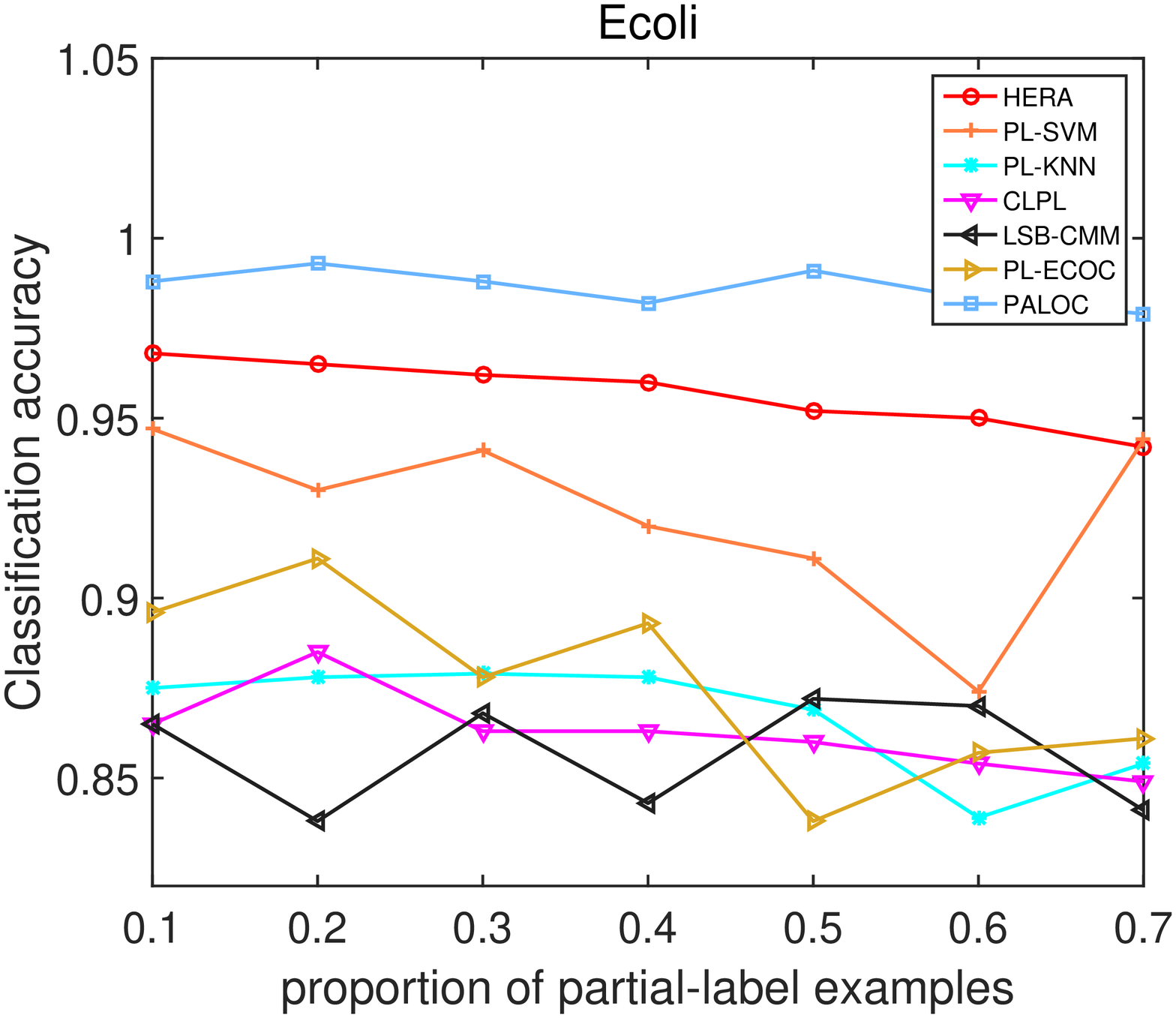}\\
\includegraphics[width = 1.77in,height=1.3in]{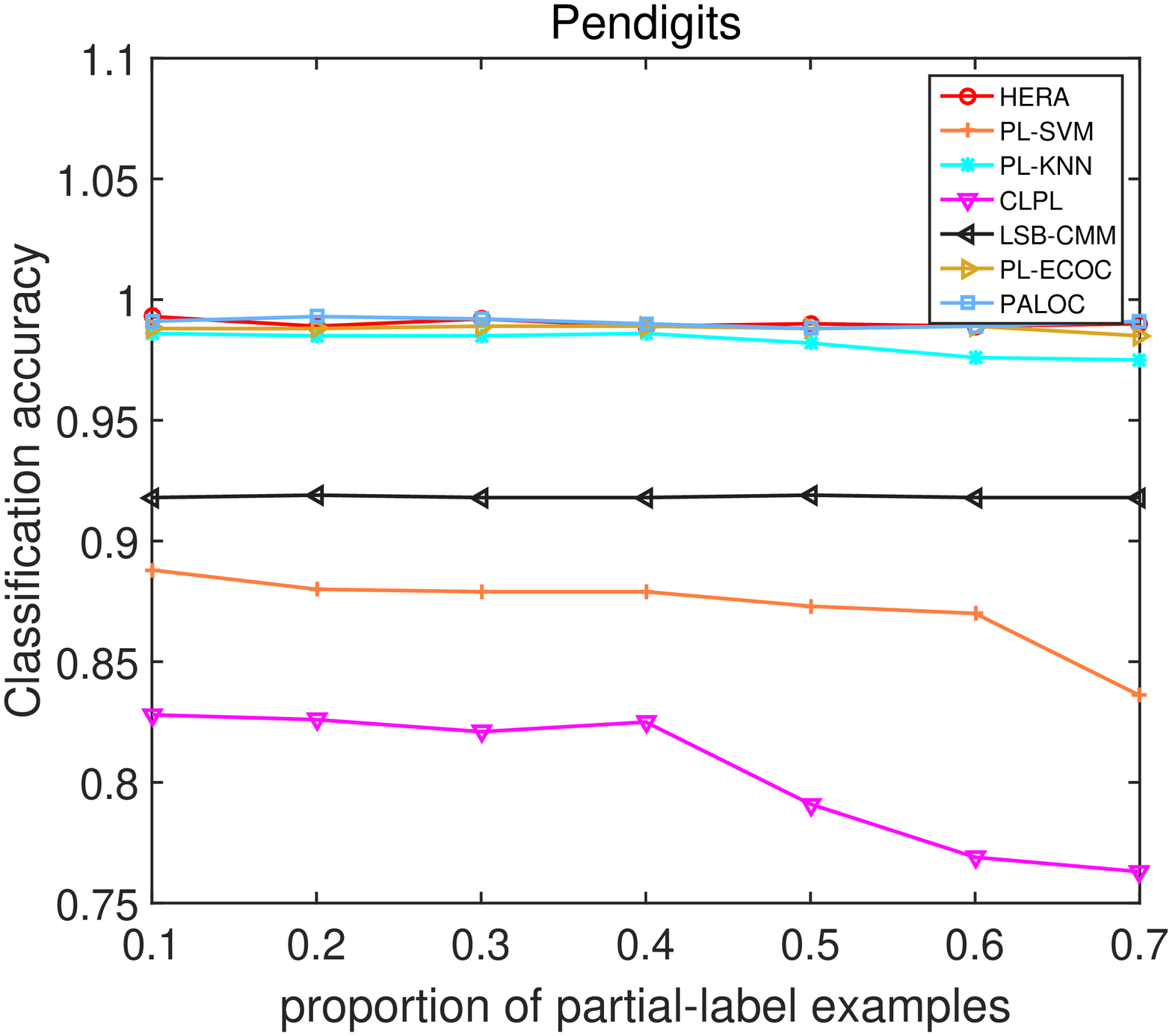}&\includegraphics[width = 1.77in,height=1.3in]{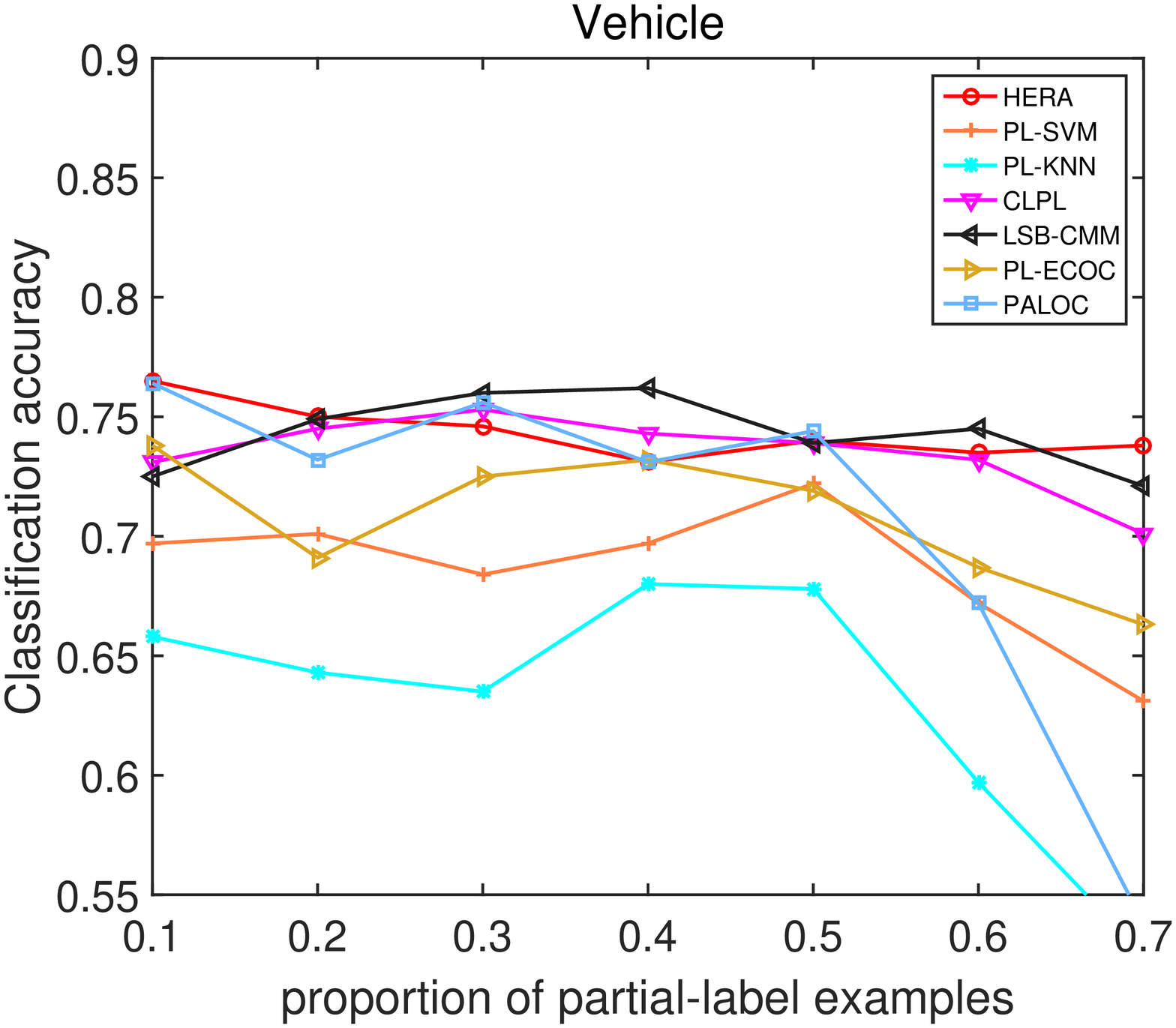}&\includegraphics[width = 1.77in,height=1.3in]{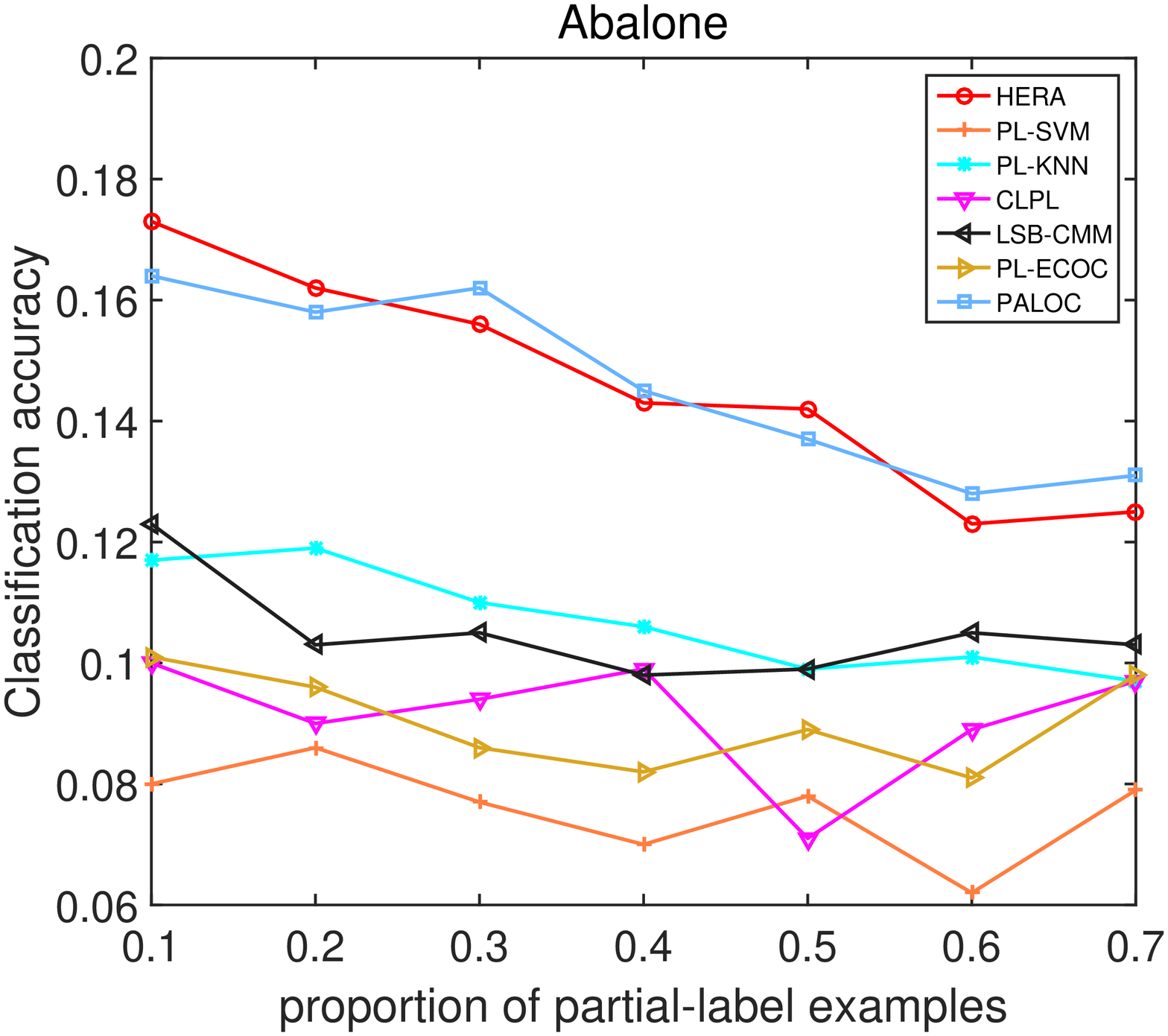}\\
\includegraphics[width = 1.77in,height=1.3in]{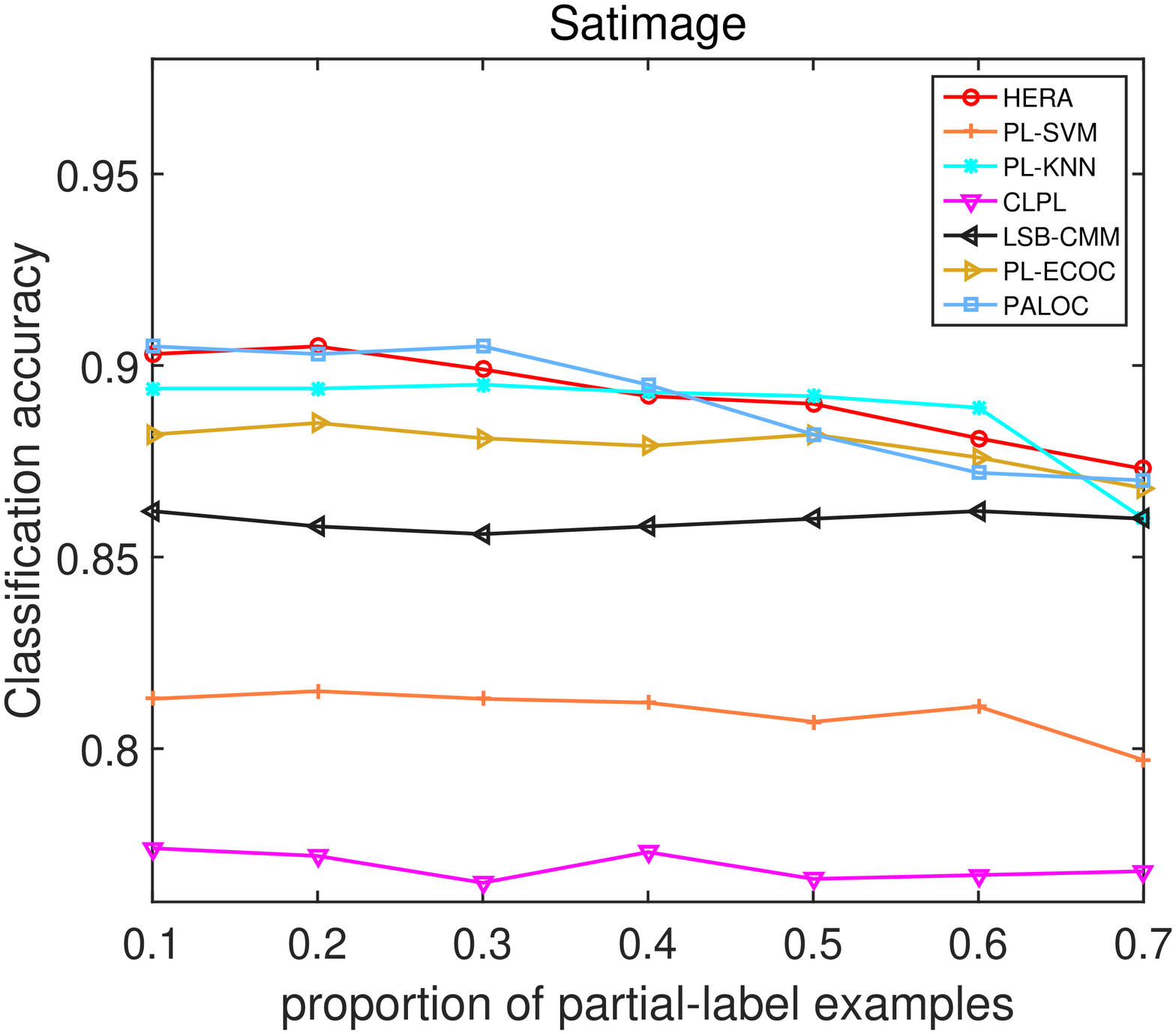}&\includegraphics[width = 1.77in,height=1.3in]{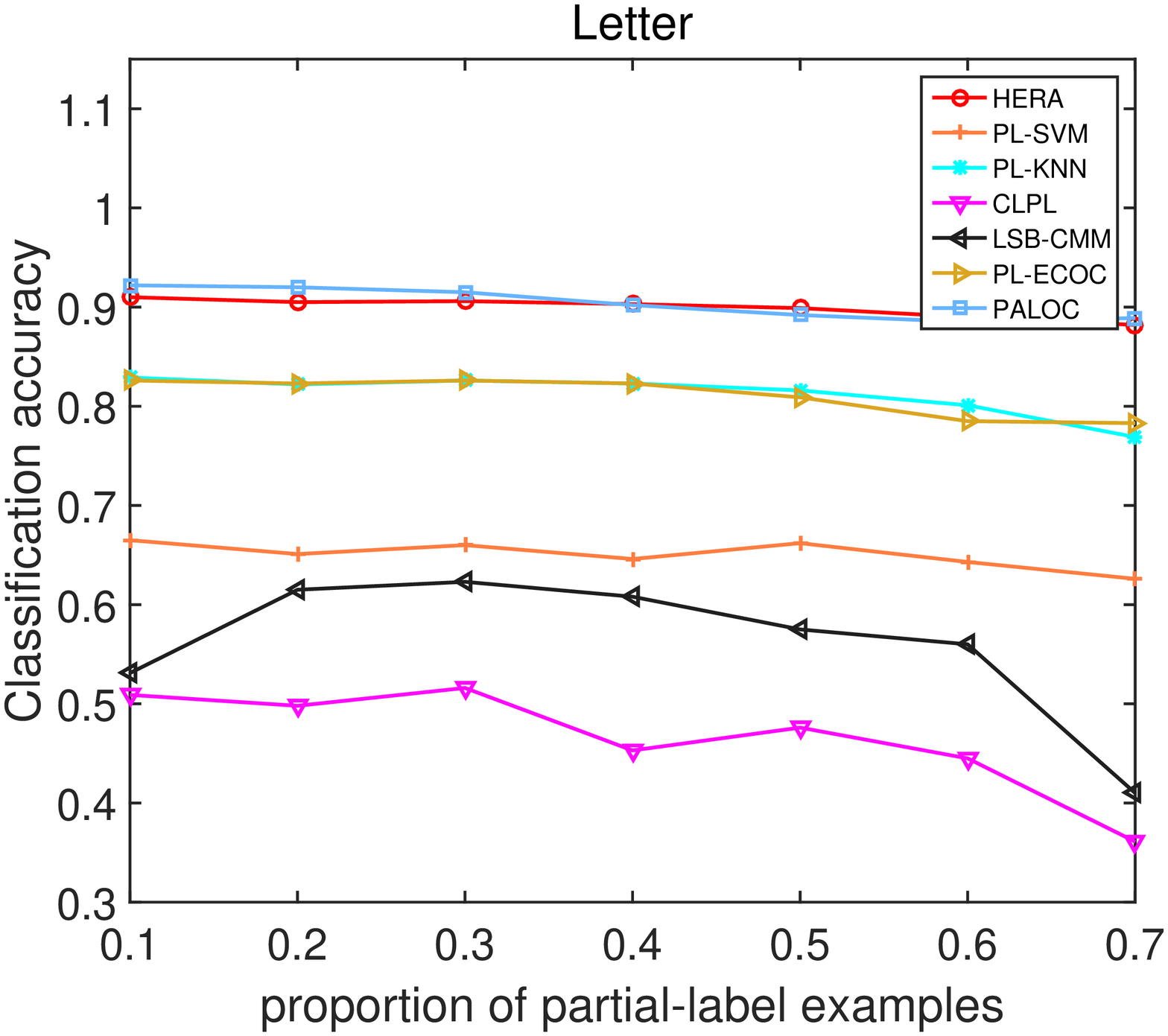}&\includegraphics[width = 1.77in,height=1.3in]{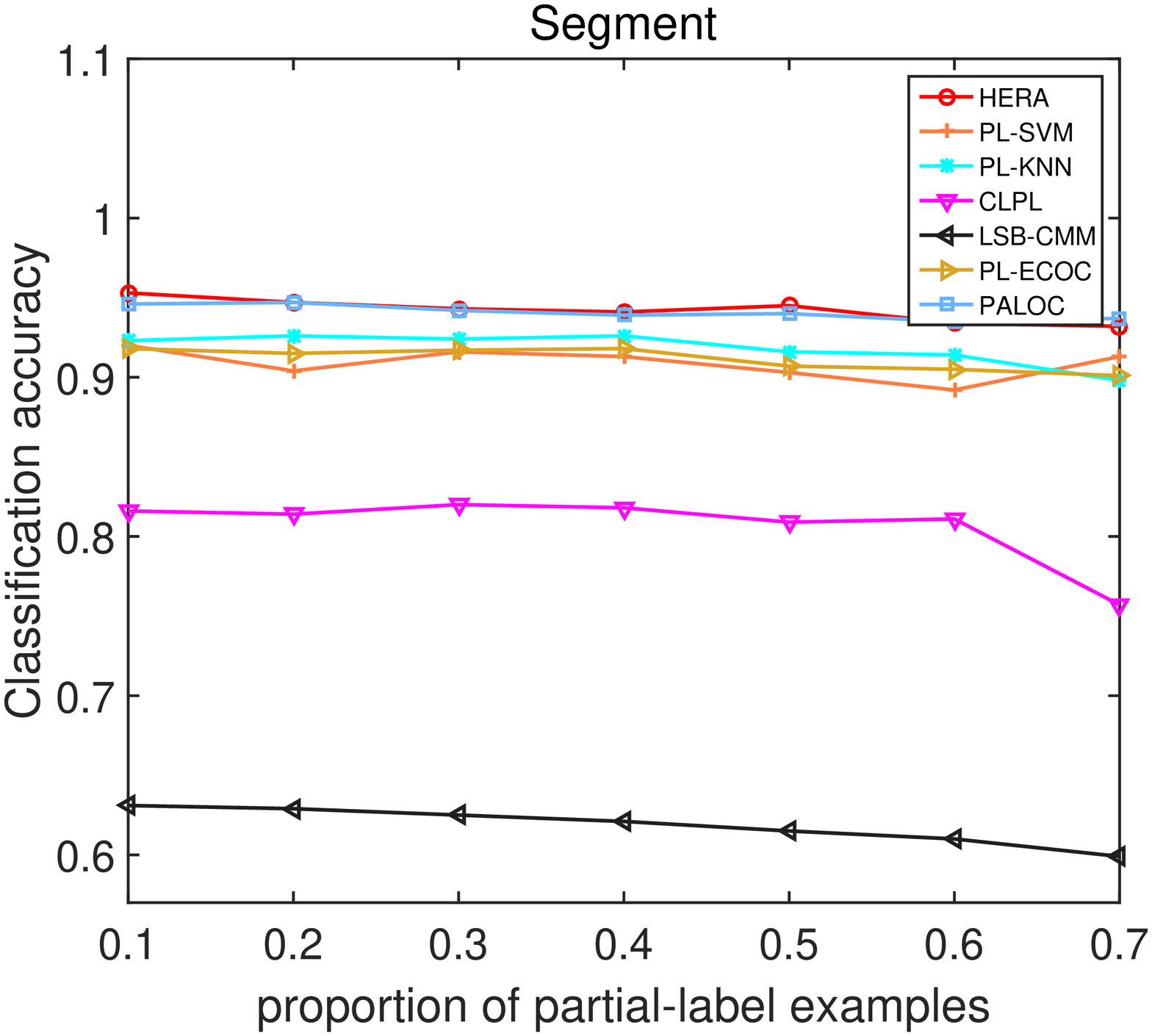}\\
\end{tabular}
\caption{The classification accuracy of several comparing methods on nine controlled UCI data sets changes as $\epsilon$ (co-occurring probability of the coupling label) increases from 0.1 to 0.7 (with 100\% partial labeled instances [p = 1] and one false positive candidate label [r = 1 ]).}
\label{fig-epsilon}
\vspace{0mm}
\end{figure*}

\subsection{Experimental Results}
In our paper, the experimental results of comparing methods are obtained by running the source codes provided by the authors, where the model parameters are configured according to the suggestions in respective literatures.

\begin{table*}[!ht]
\centering
\vspace{6mm}
\caption{Win/tie/loss counts of the HERA's classification performance against each comparing method on controlled UCI data sets (pairwise $t$-test at 0.05 significance level)}
\vspace{2mm}
\label{wintieloss}
\resizebox{14cm}{!}{
\begin{tabular}{c|cccccc|c}
\cline{1-6}
\hline \hline
Data set    &PL-KNN    &PL-SVM   &LSB-CMM  &CLPL       &PL-ECOC   &PALOC       &SUM  \\\hline
Glass       &23/3/2    &26/2/0   &25/3/0   &27/1/0     &20/8/0    &13/11/4     &134/28/6 \\
Segment     &16/12/0   &21/7/0   &27/1/0   &28/0/0     &1/26/1    &6/21/1      &99/67/2  \\
Vehicle     &28/0/0    &25/3/0   &5/14/9   &13/10/5    &2/6/20    &12/12/4     &85/45/38\\
Letter      &28/0/0    &28/0/0   &28/0/0   &28/0/0     &28/0/0    &21/7/0      &161/7/0\\
Satimage    &1/27/0    &28/0/0   &21/7/0   &28/0/0     &1/27/1    &9/14/5      &87/75/6\\
Abalone     &22/6/0    &27/1/0   &11/16/1  &25/3/0     &5/20/3    &0/24/4      &90/70/8\\
Ecoli       &28/0/0    &22/6/0   &28/0/0   &28/0/0     &28/0/0    &21/7/0      &155/13/0\\
Dermatology &21/7/0    &21/7/0   &22/6/0   &21/7/0     &21/7/0    &16/12/0     &112/46/0\\
Pendigits   &2/26/0    &28/0/0   &28/0/0   &28/0/0     &1/27/0    &0/27/1      &87/80/1\\ \hline
SUM         &169/81/2 &226/26/0  &199/47/10 &230/21/5  &100/121/25  &102/135/19    &-      \\ \hline \hline
\end{tabular}}
\vspace{1mm}
\end{table*}

\begin{table*}[!ht]
\centering
\setlength{\abovecaptionskip}{0pt}%
\setlength{\belowcaptionskip}{5pt}%
\caption{ Classification accuracy of each algorithm on real-world data sets. $\bullet/\circ$ indicates that HERA is statistically superior / inferior to the algorithm (pairwise $t$-test at 0.05 significance level). }
\label{Table-RW}
\vspace{2mm}
\resizebox{!}{1.6cm}{
\begin{tabular}{ccccccc}
\hline \hline
 & Lost  & MSRCv2 & BirdSong & SoccerPlayer & FG-NET & Yahoo! News \\
\hline
\textbf{HERA}  & 0.712$\pm$0.035   & \textbf{0.510$\pm$0.027}    & 0.695$\pm$0.021    & \textbf{0.539$\pm$0.015}          & \textbf{0.076$\pm$0.017}   &\textbf{0.646$\pm$0.005} \\ \hline
\textbf{PL-SVM}  & 0.729$\pm$0.056 $\circ$ & 0.482$\pm$0.027 $\bullet$  & 0.663$\pm$0.018 $\bullet$    & 0.497$\pm$0.004 $\bullet$        & 0.063$\pm$0.010 $\bullet$ &0.636$\pm$0.010 $\bullet$\\
\textbf{CLPL}    & \textbf{0.742$\pm$0.024} $\circ$ & 0.413$\pm$0.020 $\bullet$  & 0.632$\pm$0.009 $\bullet$    & 0.368$\pm$0.004 $\bullet$        & 0.063$\pm$0.017 $\bullet$  &0.462$\pm$0.009 $\bullet$\\
\textbf{PL-KNN}  & 0.424$\pm$0.030 $\bullet$ & 0.448$\pm$0.012 $\bullet$  & 0.614$\pm$0.009 $\bullet$    & 0.443$\pm$0.004 $\bullet$       & 0.039$\pm$0.008 $\bullet$  &0.457$\pm$0.010 $\bullet$\\
\textbf{LSB-CMM} & 0.707$\pm$0.019 $\bullet$ & 0.456$\pm$0.008 $\bullet$  & 0.717$\pm$0.015 $\circ$    & 0.525$\pm$0.006 $\circ$        & 0.059$\pm$0.008 $\bullet$  &0.642$\pm$0.007 $\bullet$\\
\textbf{PL-ECOC} & 0.703$\pm$0.052 $\bullet$ & 0.505$\pm$0.027 $\bullet$  & \textbf{0.740$\pm$0.016} $\circ$    & 0.537$\pm$0.020 $\circ$        &   0.040$\pm$0.018  $\bullet$     &0.610$\pm$0.008 $\bullet$\\
\textbf{PALOC}   & 0.629$\pm$0.056 $\bullet$ & 0.479$\pm$0.042 $\bullet$  & 0.711$\pm$0.016 $\circ$    & 0.537$\pm$0.015 $\circ$        & 0.065$\pm$0.019 $\bullet$  &0.625$\pm$0.005 $\bullet$\\
\hline \hline
\end{tabular}}
\vspace{-2mm}
\end{table*}

\subsubsection{Controlled UCI data sets}
We compare the HERA with all above comparing methods on nine controlled UCI data sets, and Figure \ref{fig-uci-1}-\ref{fig-uci-3} illustrate the classification accuracy of each comparing method as $p$ increases from 0.1 to 0.7 with the step-size of 0.1. Together with the ground-truth label, the $r$ class labels are randomly chosen from $\mathcal{Y}$ to constitute the rest of each candidate label set, where $r\in\{1, 2, 3\}$. Figure \ref{fig-epsilon} illustrates the classification accuracy of each comparing algorithm as $\epsilon$ increases from 0.1 to 0.7 with step-size 0.1 ($p = 1, r = 1$). Table \ref{wintieloss} summaries the win/tie/loss counts between HERA and the other comparing methods. Out of 252 (9 data sets $\times$ 28 configurations) statistical comparisons show that HERA achieves either superior or comparable performance against the six comparing methods:
\begin{itemize}
    \item Among these comparing methods, HERA achieves superior performance against PL-SVM, LSB-CMM and CLPL in most cases (over 80\% cases). And compared with PL-KNN, PL-ECOC and PALOC, it also achieves superior or comparable performance in 99.2\%, 86.3\%, 92.5\% cases, respectively. The experimental results demonstrate that HERA has superior capacity of disambiguation against other methods based on varying disambiguation strategies, as well as disambiguation-free strategy.
    \item Among these UCI data sets, HERA outperforms all comparing methods on \emph{Ecoli} and \emph{Letter} data set. And for other controlled UCI data sets, it is also superior or comparable against all the comparing state-of-the-art methods. Specifically, the average classification accuracy of HERA is 15.8\% higher than PL-KNN on \emph{Ecoli} data set and 12.8\% higher than LSB-CMM on \emph{Dermatology} data set. Meanwhile, for CLPL and PL-ECOC, it also separately has 14.3\% and 5.7\% higher classification accuracy on \emph{Pendigits} and \emph{Glass} data set.
    \item Overall, among the 252 statistical comparisons and 1512 experimental results, HERA outperforms other comparing methods in 1026 cases and achieves comparable performance in 431 cases (total: 96\% cases), which strongly demonstrates the effectiveness of the proposed HERA algorithm.
\end{itemize}

\begin{figure*}
\centering
\begin{tabular}{cc}
\includegraphics[width = 2.3in,height=1.6in]{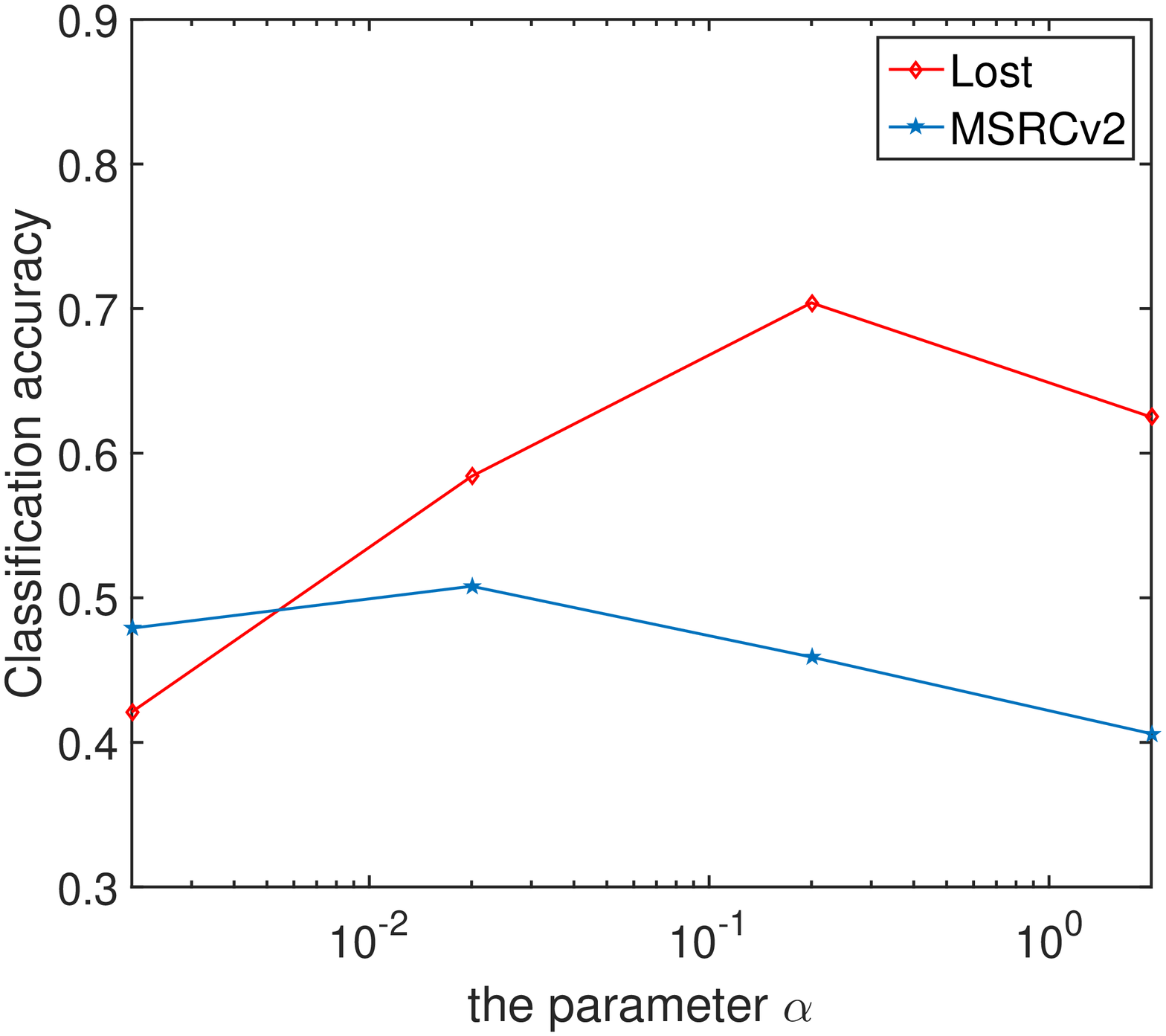}&\includegraphics[width = 2.3in,height=1.6in]{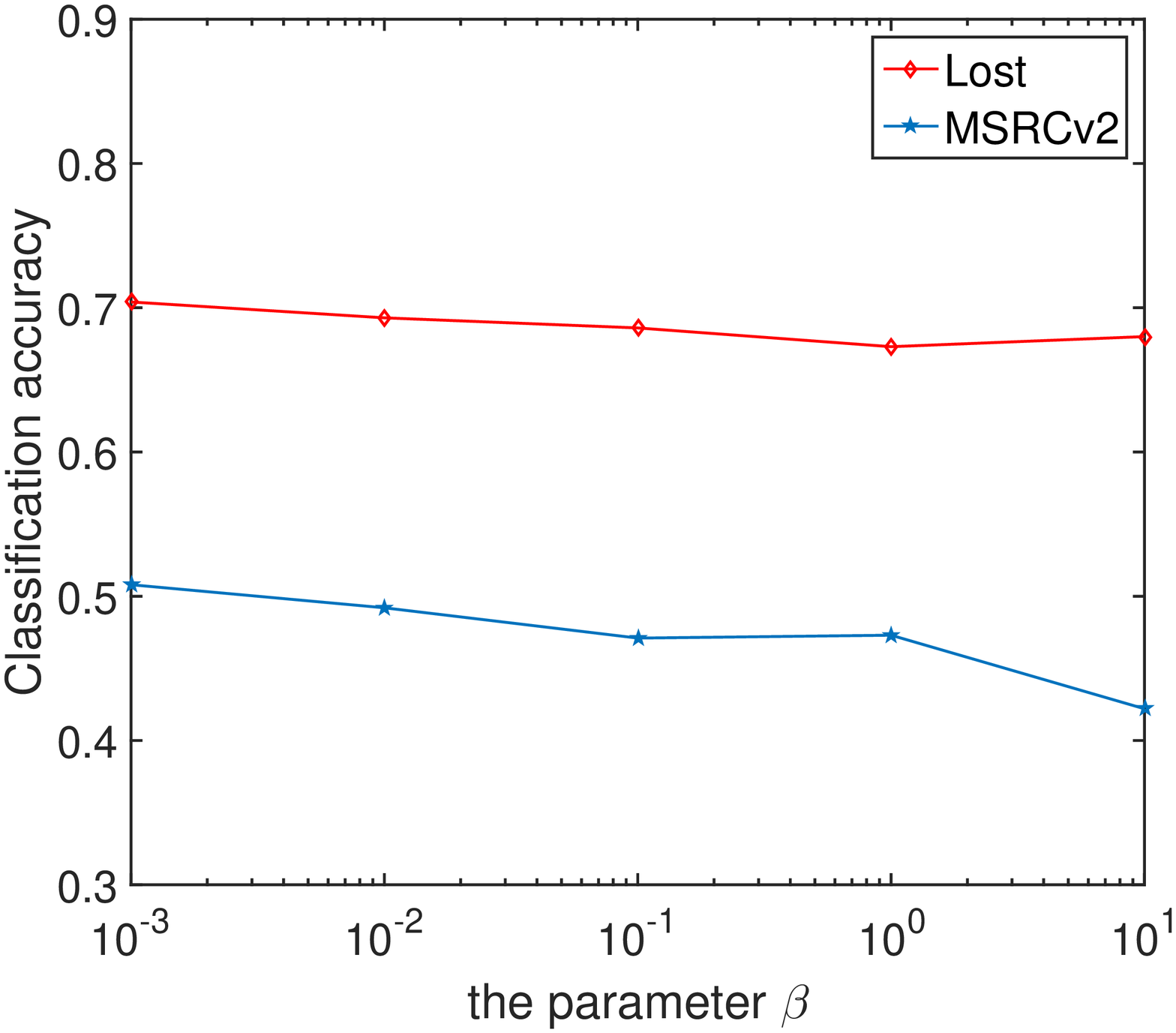}\\
\includegraphics[width = 2.3in,height=1.6in]{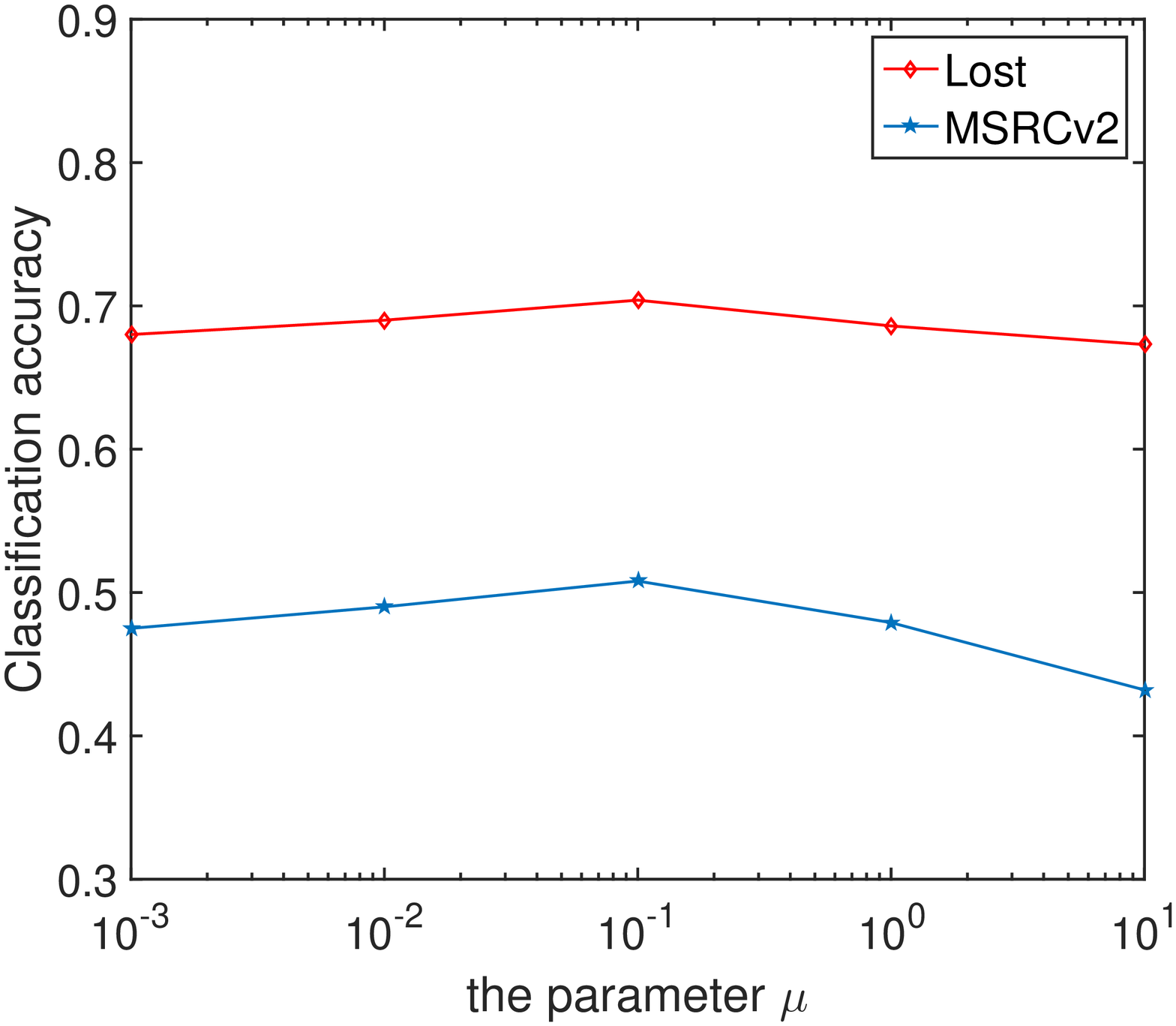}&\includegraphics[width = 2.3in,height=1.6in]{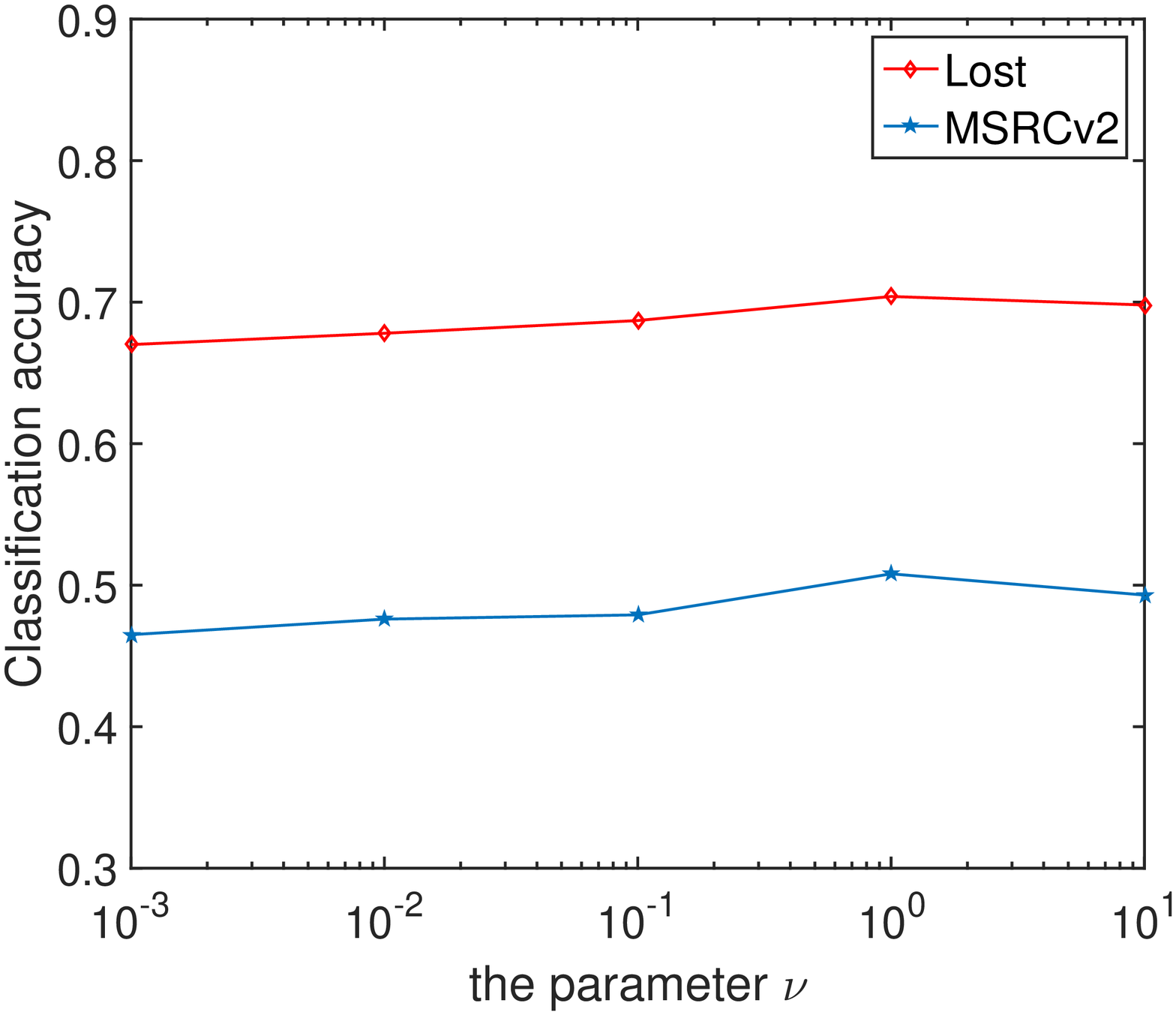}\\
\end{tabular}
\caption{Parameter analysis: the classification accuracy changes as each parameter increases with other parameters fixed.}
\label{fig-analysis}
\vspace{0mm}
\end{figure*}

\subsubsection{Real-World (RW) data sets}
We compare the HERA with all above comparing algorithms on the RW data sets. The comparison results are reported in Table \ref{Table-RW}, where the recorded results are based on ten-fold cross-validation. According to Table \ref{Table-RW}, it is clear to observe that HERA performs better than most comparing PLL algorithms on these RW data sets.
\begin{itemize}
    \item From the view of the comparing methods, HERA achieves superior or comparable performance against all comparing state-of-the-art methods. Especially, compared with the ADS-based methods, the classification accuracy of HERA is 28.5\% higher than PL-KNN on \emph{Lost} data set and 9.5\% higher than CLPL on \emph{MSRCv2} data set. And compared with the IDS-based methods, HERA gets 2.2\% higher classification accuracy against PL-SVM on \emph{BirdSong} data set. Besides, when compared with DFS-based methods PL-ECOC and PALOC, it also can achieve 3.6\% and 1.1\% higher performance on \emph{Yahoo! News} and \emph{FG-NET} data set, respectively.
    \item From the view of the employed data sets, HERA also performs good disambiguation ability on all RW data sets. Specifically, the classification accuracies of HERA are totally higher than that of all other comparing methods on \emph{MSRCv2}, \emph{FG-NET}, \emph{SoccerPlayer} and \emph{Yahoo! News} data sets. For \emph{Lost} data sets, it also outperforms the comparing methods on 4/6 data sets. Besides, even on the \emph{BirdSong} data set, it can also outperform half of comparing methods.
    \item Note that, although the proposed HERA does not achieve the best performance on all Real World data sets, it has outperformed each comparing methods on at least 4/6 data sets and the improvement is also relatively significant. Therefore, the effectiveness of our proposed algorithm is demonstrated.
\end{itemize}

\subsubsection{Summary}
The two series of experiments mentioned above powerfully demonstrate the effectiveness of HERA, and we attribute the success to the superiority of heterogeneous loss and the special exploit of sparse and low-rank scheme, i.e. simultaneously integrate the pairwise ranking loss and pointwise reconstruction loss into a unified framework, and regularize the ground-truth label matrix with sparse structure and the noisy label matrix with low-rank constraint. In summary, during the learning stage, the proposed HERA not only utilizes the different contribution of each candidate label but also takes the candidate label relevance of the whole training data into consideration, which jointly improves the ability of modeling disambiguation. And as expected, the experimental results demonstrate the effectiveness of our method.

\begin{figure*}
\centering
\begin{tabular}{ccc}
\includegraphics[width = 1.77in,height=1.3in]{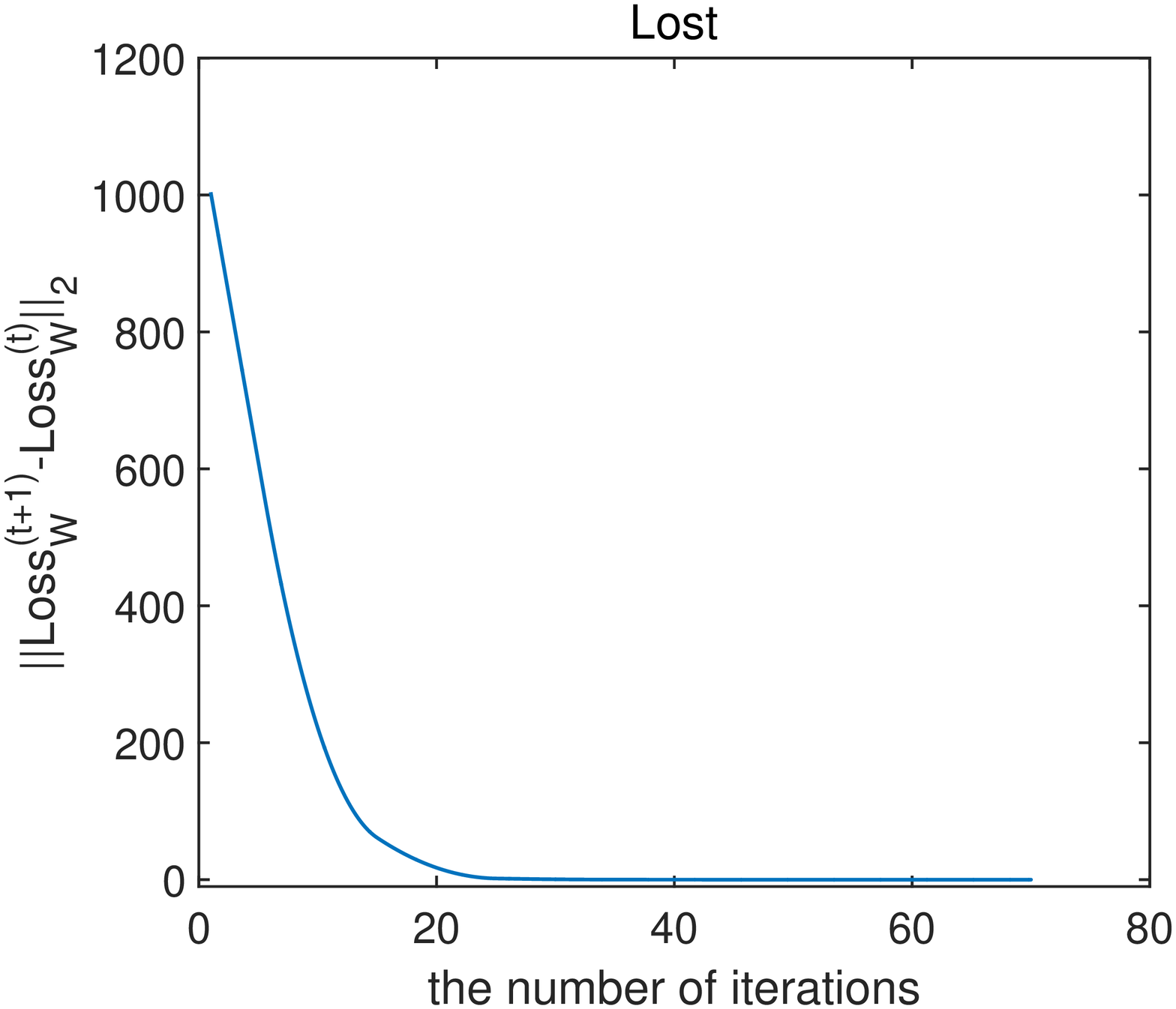}&\includegraphics[width = 1.77in,height=1.3in]{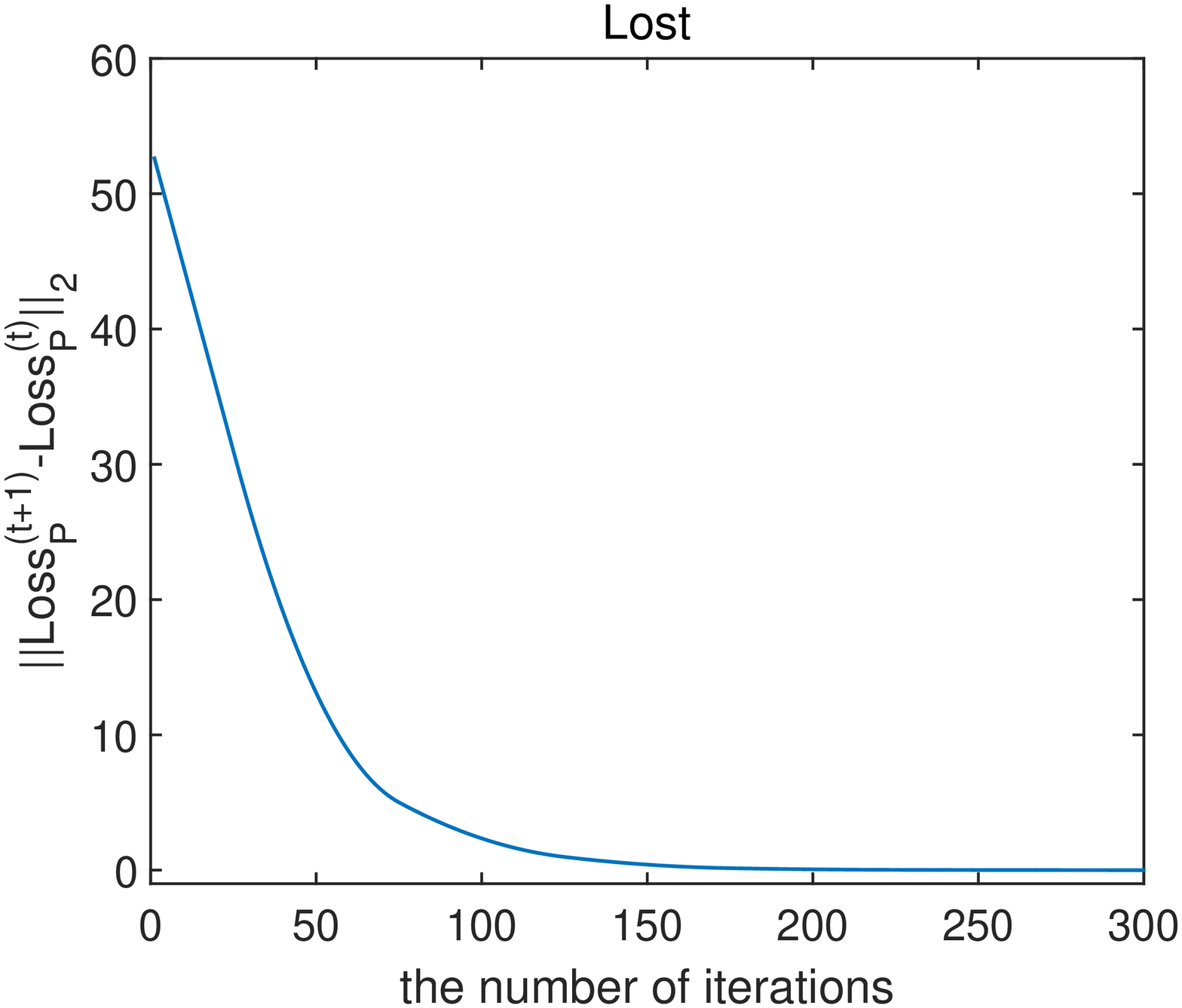}&\includegraphics[width = 1.77in,height=1.3in]{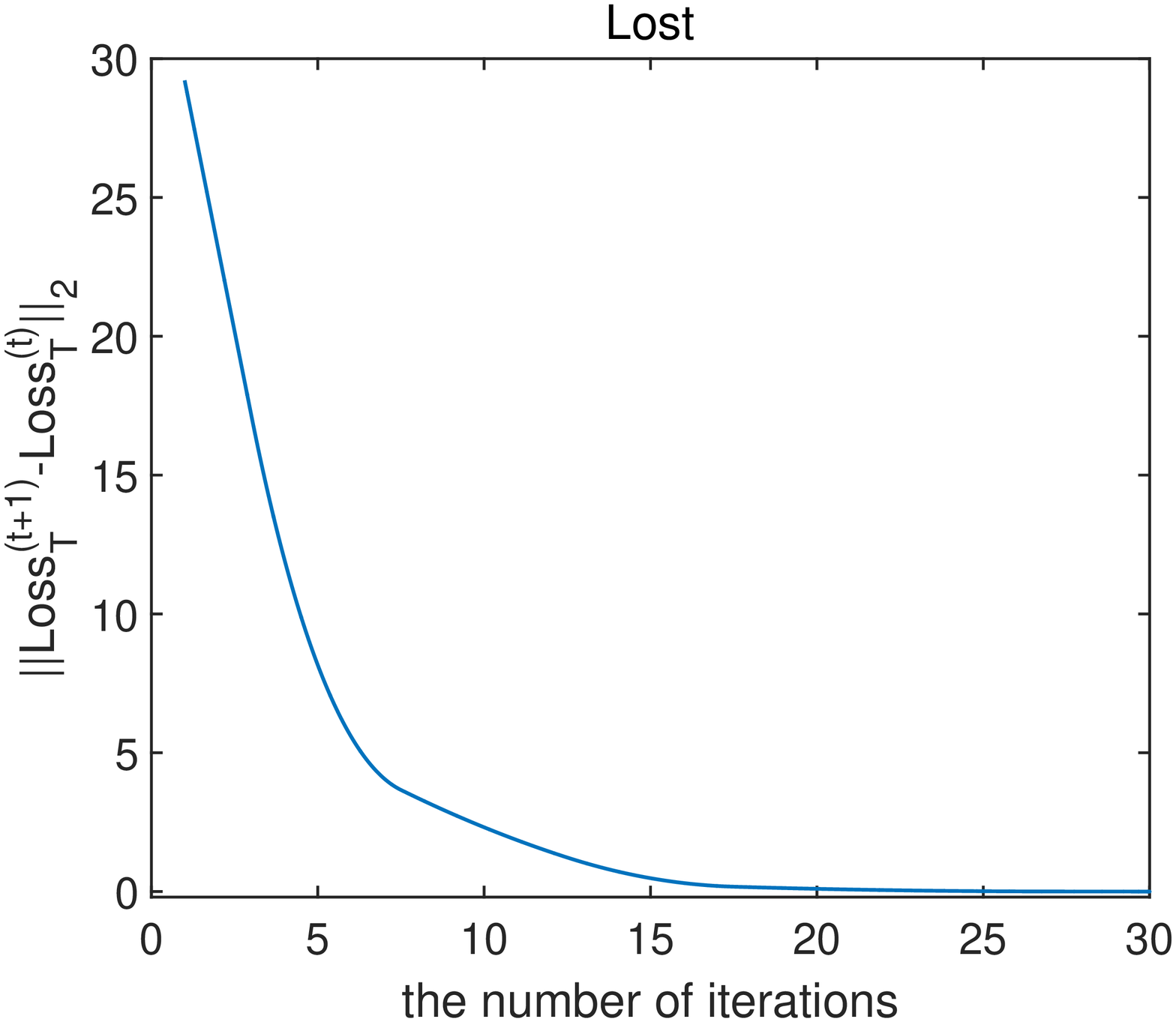}\\
\includegraphics[width = 1.77in,height=1.3in]{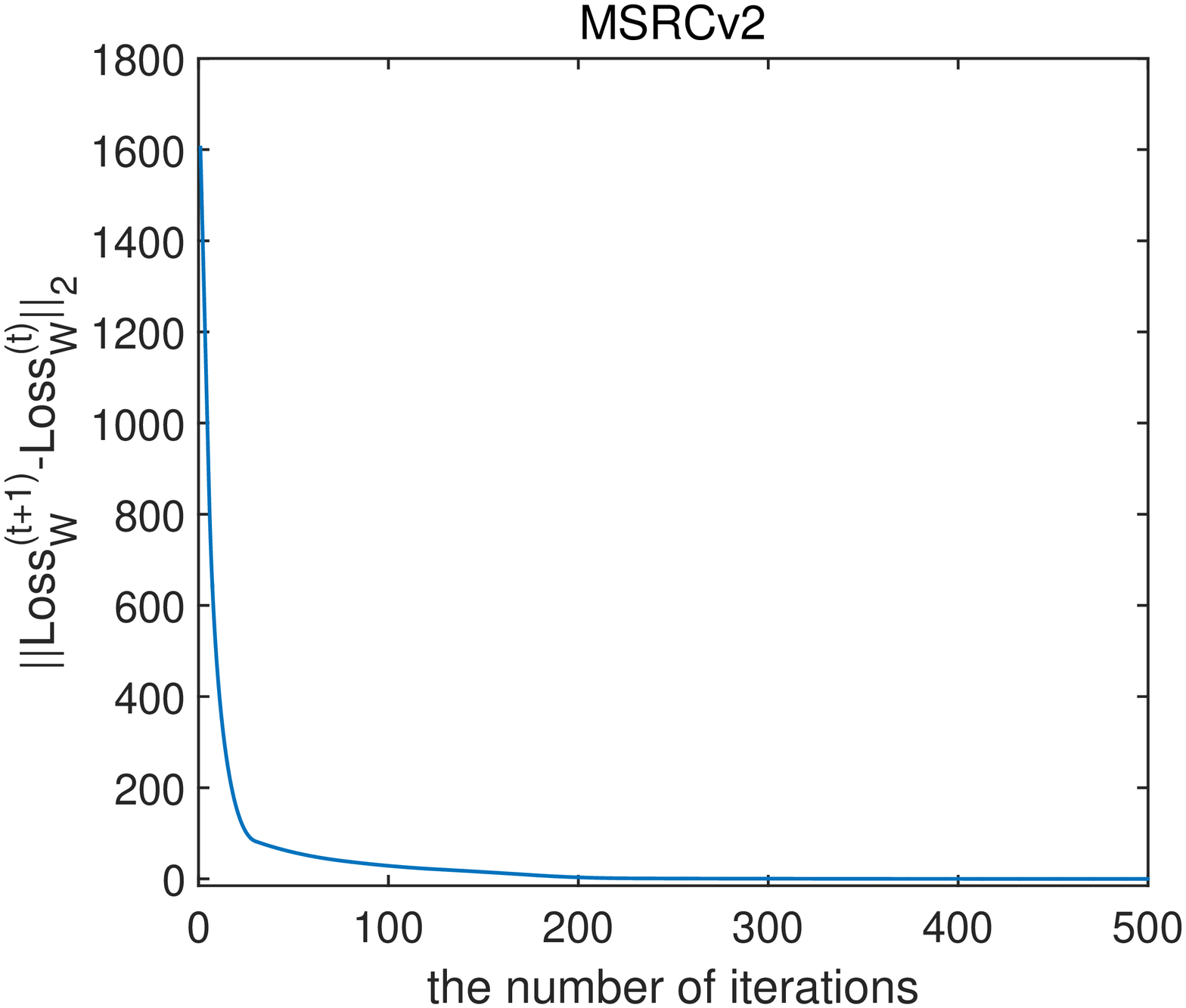}&\includegraphics[width = 1.77in,height=1.3in]{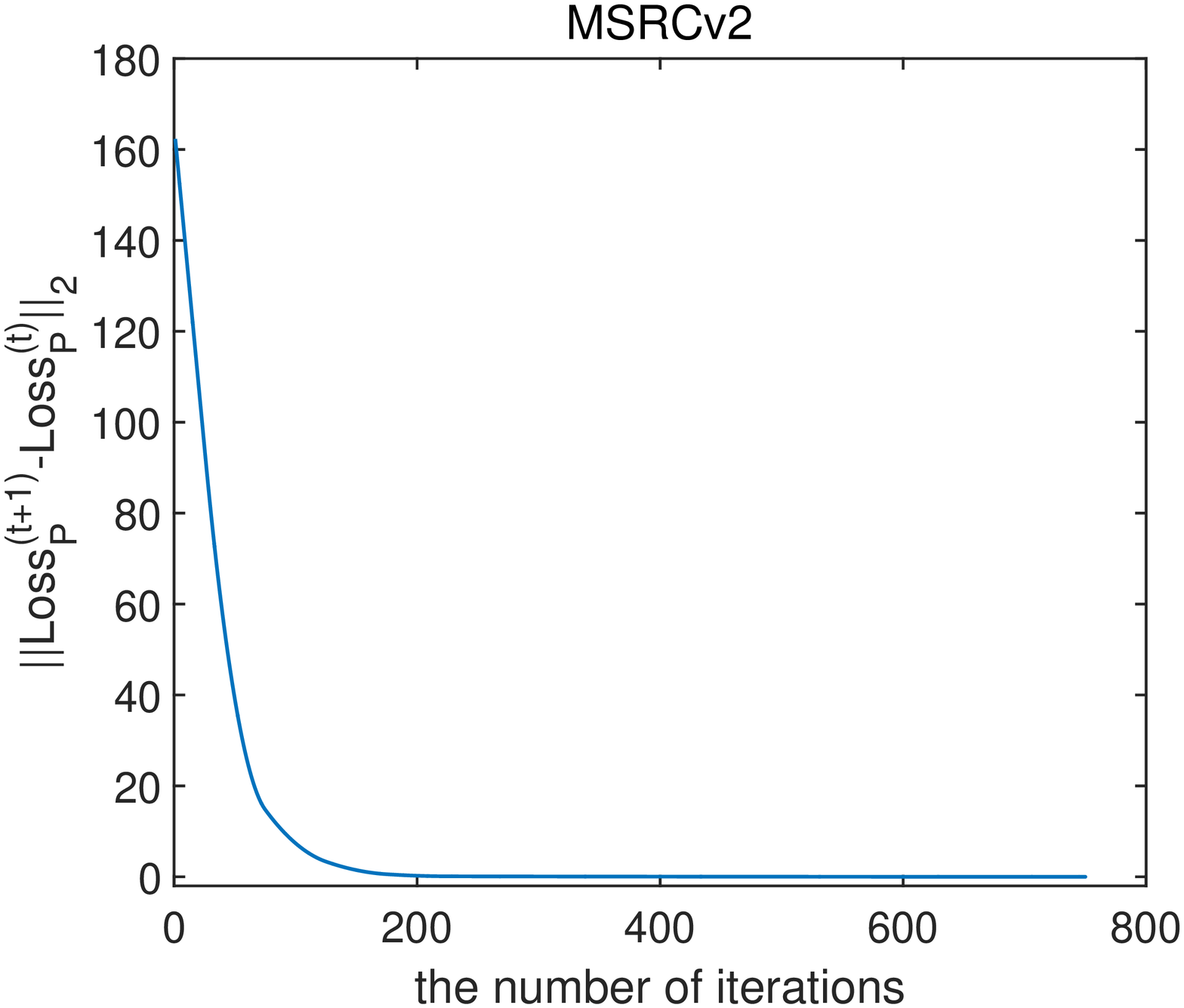}&\includegraphics[width = 1.77in,height=1.3in]{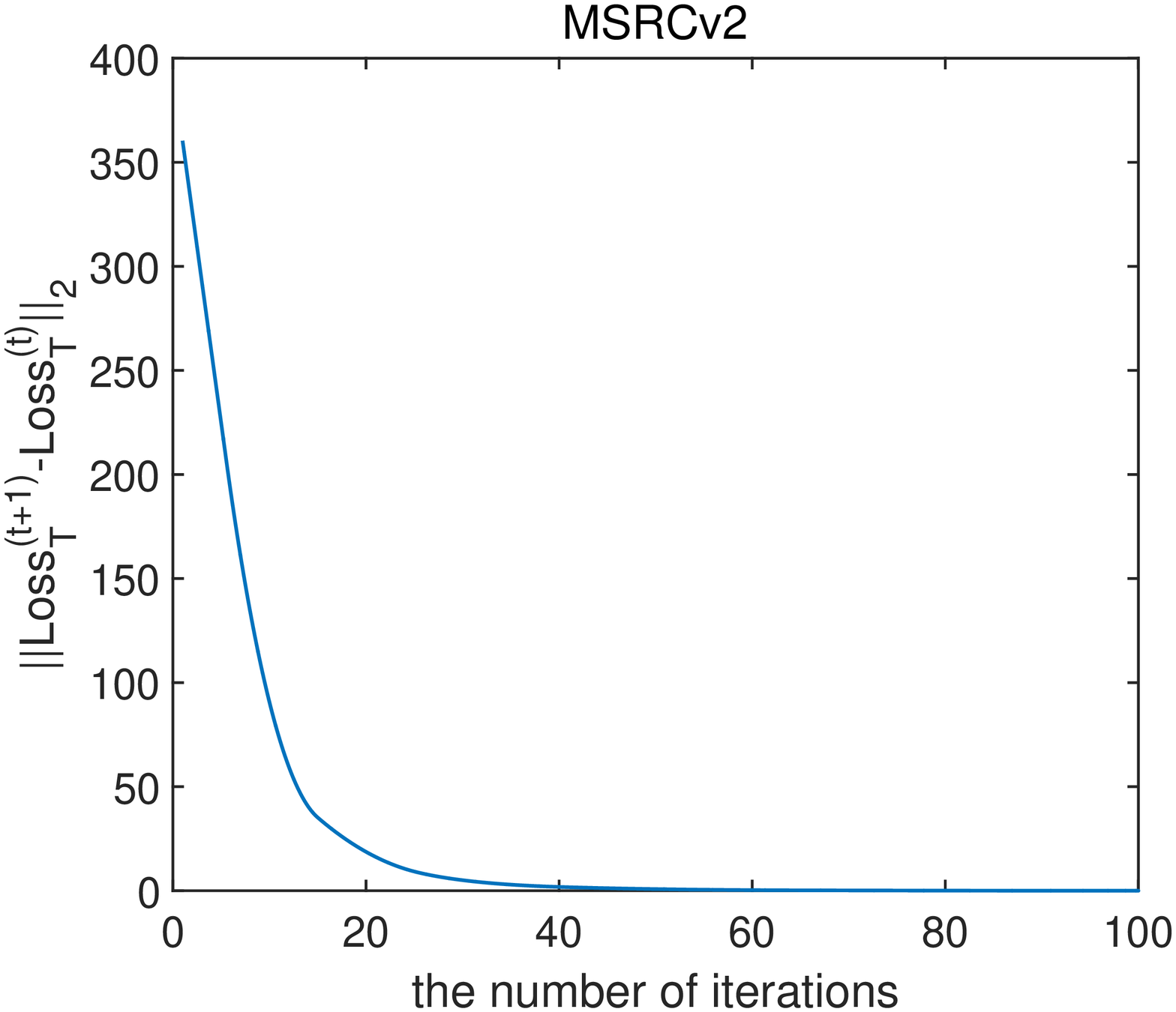}\\
\end{tabular}
\caption{The convergence curves of HREA on \emph{Lost} and \emph{MSRCV2} data sets with increasing number of iterations.}
\label{fig-covergence}
\vspace{0mm}
\end{figure*}

\subsection{Further Analysis}

\subsubsection{Sensitivity Analysis}
We study the sensitivity of HERA with respect to its four employed parameters $\alpha, \beta, \mu, \nu$. Figure \ref{fig-analysis} shows the performance of HERA under different parameter configurations on \emph{Lost} and \emph{MSRCv2} data sets. From Figure \ref{fig-analysis}, we can easily find that the value of $\alpha$ usually has great influence on the performance of the proposed framework. Faced with different data sets, we set the parameter $\alpha$ among $\{2*{10}^{-3}, \ldots, 2*{10}^{-1}\}$ via cross-validation. Besides, other parameters often follow the optimal configurations ($\beta = 10^{-3}, \mu = 0.1, \nu = 1$) but vary with minor adjustments on different data sets.
\subsubsection{Convergence Analysis}

We conduct the convergence analysis of HERA on \emph{Lost} and \emph{MSRCv2} data sets. Specifically, in Figure \ref{fig-covergence}, each group of subfigures separately illustrate the convergence curves of the sub-optimization process of model parameter \textbf{W} (left), sub-optimization process of confidence matrix \textbf{P} (center), and the whole optimization process of HERA. We can easily observe that each $\|Loss^{(t+1)}-Loss^{(t)}\|$ gradually decreases to 0 as $t$ increases. Therefore, the convergence of HERA is demonstrated.

\section{Conclusion}
In this paper, we have proposed a novel PLL method \textbf{HERA}, which simultaneously integrated the strengths of both pairwise ranking loss and pointwise reconstruction loss to provide informative labeling identification information for desired model learning. Meanwhile, the sparse and low-rank scheme is also embedded into the proposed framework to exploit the labeling confidence information of the other false candidate labels and avoid the optimization problem falling into trivial solution. Extensive experiments have empirically demonstrated our proposed method can achieve superior or comparable performance against other comparing state-of-the-art methods.

\bibliographystyle{ACM-Reference-Format}
\bibliography{ijcai18}


\begin{thebibliography}{40}


\ifx \showCODEN    \undefined \def \showCODEN     #1{\unskip}     \fi
\ifx \showDOI      \undefined \def \showDOI       #1{#1}\fi
\ifx \showISBNx    \undefined \def \showISBNx     #1{\unskip}     \fi
\ifx \showISBNxiii \undefined \def \showISBNxiii  #1{\unskip}     \fi
\ifx \showISSN     \undefined \def \showISSN      #1{\unskip}     \fi
\ifx \showLCCN     \undefined \def \showLCCN      #1{\unskip}     \fi
\ifx \shownote     \undefined \def \shownote      #1{#1}          \fi
\ifx \showarticletitle \undefined \def \showarticletitle #1{#1}   \fi
\ifx \showURL      \undefined \def \showURL       {\relax}        \fi
\providecommand\bibfield[2]{#2}
\providecommand\bibinfo[2]{#2}
\providecommand\natexlab[1]{#1}
\providecommand\showeprint[2][]{arXiv:#2}

\bibitem[\protect\citeauthoryear{Briggs, Fern, and Raich}{Briggs
  et~al\mbox{.}}{2012}]%
        {Briggs:rlsimfmia-KDDM2012}
\bibfield{author}{\bibinfo{person}{F. Briggs}, \bibinfo{person}{X. Fern}, {and}
  \bibinfo{person}{R. Raich}.} \bibinfo{year}{2012}\natexlab{}.
\newblock \showarticletitle{Rank-loss support instance machines for MIML
  instance annotation}. In \bibinfo{booktitle}{\emph{ACM SIGKDD international
  conference on Knowledge discovery and data mining}}.
  \bibinfo{pages}{534--542}.
\newblock


\bibitem[\protect\citeauthoryear{Chen, Patel, and Chellappa}{Chen
  et~al\mbox{.}}{2015}]%
        {Chen2015Matrix}
\bibfield{author}{\bibinfo{person}{C. Chen}, \bibinfo{person}{V. Patel}, {and}
  \bibinfo{person}{R. Chellappa}.} \bibinfo{year}{2015}\natexlab{}.
\newblock \showarticletitle{Matrix completion for resolving label ambiguity}.
  In \bibinfo{booktitle}{\emph{Computer Vision and Pattern Recognition}}.
  \bibinfo{pages}{4110--4118}.
\newblock


\bibitem[\protect\citeauthoryear{Chen, Patel, and Chellappa}{Chen
  et~al\mbox{.}}{2018}]%
        {chen2018learning}
\bibfield{author}{\bibinfo{person}{C. Chen}, \bibinfo{person}{V. Patel}, {and}
  \bibinfo{person}{R. Chellappa}.} \bibinfo{year}{2018}\natexlab{}.
\newblock \showarticletitle{Learning from ambiguously labeled face images}.
\newblock \bibinfo{journal}{\emph{IEEE Transactions on Pattern Analysis and
  Machine Intelligence}} (\bibinfo{year}{2018}), \bibinfo{pages}{1653--1667}.
\newblock


\bibitem[\protect\citeauthoryear{Chen, Liu, Tang, Jian, Jie, and Tao}{Chen
  et~al\mbox{.}}{2017}]%
        {Chen2017A}
\bibfield{author}{\bibinfo{person}{G. Chen}, \bibinfo{person}{T. Liu},
  \bibinfo{person}{Y. Tang}, \bibinfo{person}{Y. Jian}, \bibinfo{person}{Y.
  Jie}, {and} \bibinfo{person}{D. Tao}.} \bibinfo{year}{2017}\natexlab{}.
\newblock \showarticletitle{A Regularization Approach for Instance-Based
  Superset Label Learning}.
\newblock \bibinfo{journal}{\emph{IEEE Transactions on Cybernetics}}
  (\bibinfo{year}{2017}), \bibinfo{pages}{1--12}.
\newblock


\bibitem[\protect\citeauthoryear{Chen, Patel, Chellappa, and Phillips}{Chen
  et~al\mbox{.}}{2014}]%
        {Chen:IEEET2014}
\bibfield{author}{\bibinfo{person}{Y. Chen}, \bibinfo{person}{V. Patel},
  \bibinfo{person}{R. Chellappa}, {and} \bibinfo{person}{P. Phillips}.}
  \bibinfo{year}{2014}\natexlab{}.
\newblock \showarticletitle{Ambiguously Labeled Learning Using Dictionaries}.
\newblock \bibinfo{journal}{\emph{IEEE Transactions on Information Forensics
  and Security}} (\bibinfo{year}{2014}), \bibinfo{pages}{2076--2088}.
\newblock


\bibitem[\protect\citeauthoryear{Chiang, Dhillon, and Hsieh}{Chiang
  et~al\mbox{.}}{2018}]%
        {chiang2018using}
\bibfield{author}{\bibinfo{person}{K. Chiang}, \bibinfo{person}{I. Dhillon},
  {and} \bibinfo{person}{C. Hsieh}.} \bibinfo{year}{2018}\natexlab{}.
\newblock \showarticletitle{Using side information to reliably learn low-rank
  matrices from missing and corrupted observations}.
\newblock \bibinfo{journal}{\emph{The Journal of Machine Learning Research}}
  (\bibinfo{year}{2018}), \bibinfo{pages}{3005--3039}.
\newblock


\bibitem[\protect\citeauthoryear{Cour, Sapp, and Taskar}{Cour
  et~al\mbox{.}}{2011}]%
        {Cour:lfpl-JMLR2011}
\bibfield{author}{\bibinfo{person}{T. Cour}, \bibinfo{person}{B. Sapp}, {and}
  \bibinfo{person}{B. Taskar}.} \bibinfo{year}{2011}\natexlab{}.
\newblock \showarticletitle{Learning from Partial Labels}.
\newblock \bibinfo{journal}{\emph{IEEE Transactions on Knowledge and Data
  Engineering}} (\bibinfo{year}{2011}), \bibinfo{pages}{1501--1536}.
\newblock


\bibitem[\protect\citeauthoryear{Dietterich and Bakiri}{Dietterich and
  Bakiri}{1994}]%
        {d1994solving}
\bibfield{author}{\bibinfo{person}{T. Dietterich} {and} \bibinfo{person}{G.
  Bakiri}.} \bibinfo{year}{1994}\natexlab{}.
\newblock \showarticletitle{Solving multiclass learning problems via
  error-correcting output codes}.
\newblock \bibinfo{journal}{\emph{Journal of Artificial Intelligence Research}}
  (\bibinfo{year}{1994}), \bibinfo{pages}{263--286}.
\newblock


\bibitem[\protect\citeauthoryear{Feng and An}{Feng and An}{2018}]%
        {feng2018leveraging}
\bibfield{author}{\bibinfo{person}{L. Feng} {and} \bibinfo{person}{B. An}.}
  \bibinfo{year}{2018}\natexlab{}.
\newblock \showarticletitle{Leveraging Latent Label Distributions for Partial
  Label Learning.}. In \bibinfo{booktitle}{\emph{International Joint Conference
  on Artificial Intelligence}}. \bibinfo{pages}{2107--2113}.
\newblock


\bibitem[\protect\citeauthoryear{Feng and An}{Feng and An}{2019a}]%
        {Feng:IJCAI2019}
\bibfield{author}{\bibinfo{person}{L. Feng} {and} \bibinfo{person}{B. An}.}
  \bibinfo{year}{2019}\natexlab{a}.
\newblock \showarticletitle{Partial Label Learning by Semantic Difference
  Maximization}. In \bibinfo{booktitle}{\emph{International Joint Conference on
  Artificial Intelligence}}. \bibinfo{pages}{in press}.
\newblock


\bibitem[\protect\citeauthoryear{Feng and An}{Feng and An}{2019b}]%
        {feng2019partial}
\bibfield{author}{\bibinfo{person}{L. Feng} {and} \bibinfo{person}{B. An}.}
  \bibinfo{year}{2019}\natexlab{b}.
\newblock \showarticletitle{Partial Label Learning with Self-Guided
  Retraining}. In \bibinfo{booktitle}{\emph{AAAI Conference on Artificial
  Intelligence}}. \bibinfo{pages}{in press}.
\newblock


\bibitem[\protect\citeauthoryear{Grandvalet and Bengio}{Grandvalet and
  Bengio}{2004}]%
        {Grandvalet:CWP2004}
\bibfield{author}{\bibinfo{person}{Y. Grandvalet} {and} \bibinfo{person}{Y.
  Bengio}.} \bibinfo{year}{2004}\natexlab{}.
\newblock \showarticletitle{Learning from Partial Labels with Minimum Entropy}.
\newblock \bibinfo{journal}{\emph{Cirano Working Papers}}
  (\bibinfo{year}{2004}), \bibinfo{pages}{512--517}.
\newblock


\bibitem[\protect\citeauthoryear{Guillaumin, Verbeek, and Schmid}{Guillaumin
  et~al\mbox{.}}{2010}]%
        {Guill:mimlfalbof-ECCV2010}
\bibfield{author}{\bibinfo{person}{M. Guillaumin}, \bibinfo{person}{J.
  Verbeek}, {and} \bibinfo{person}{C. Schmid}.}
  \bibinfo{year}{2010}\natexlab{}.
\newblock \showarticletitle{Multiple instance metric learning from
  automatically labeled bags of faces}. In \bibinfo{booktitle}{\emph{European
  Conference on Computer Vision}}. \bibinfo{pages}{634--647}.
\newblock


\bibitem[\protect\citeauthoryear{Hullermeier and Beringer}{Hullermeier and
  Beringer}{2005}]%
        {Huller:LNCS2005}
\bibfield{author}{\bibinfo{person}{E. Hullermeier} {and} \bibinfo{person}{J.
  Beringer}.} \bibinfo{year}{2005}\natexlab{}.
\newblock \showarticletitle{Learning from Ambiguously Labeled Examples}.
\newblock \bibinfo{journal}{\emph{International Symposium on Intelligent Data
  Analysis}} (\bibinfo{year}{2005}), \bibinfo{pages}{168--179}.
\newblock


\bibitem[\protect\citeauthoryear{Jin and Ghahramani}{Jin and
  Ghahramani}{2003}]%
        {jin2003learning}
\bibfield{author}{\bibinfo{person}{R. Jin} {and} \bibinfo{person}{Z.
  Ghahramani}.} \bibinfo{year}{2003}\natexlab{}.
\newblock \showarticletitle{Learning with multiple labels}. In
  \bibinfo{booktitle}{\emph{Advances in Neural Information Processing
  Systems}}. \bibinfo{pages}{921--928}.
\newblock


\bibitem[\protect\citeauthoryear{Liu, Lin, Yan, Sun, Yu, and Ma}{Liu
  et~al\mbox{.}}{2013}]%
        {liurrsslrr2013}
\bibfield{author}{\bibinfo{person}{G. Liu}, \bibinfo{person}{C. Lin},
  \bibinfo{person}{S. Yan}, \bibinfo{person}{J. Sun}, \bibinfo{person}{J. Yu},
  {and} \bibinfo{person}{Y. Ma}.} \bibinfo{year}{2013}\natexlab{}.
\newblock \showarticletitle{Robust Recovery of Subspace Structures by Low-Rank
  Representation}.
\newblock \bibinfo{journal}{\emph{IEEE Transactions on Pattern Analysis and
  Machine Intelligence}} (\bibinfo{year}{2013}), \bibinfo{pages}{171--184}.
\newblock


\bibitem[\protect\citeauthoryear{Liu and Dietterich}{Liu and
  Dietterich}{2012}]%
        {Liu:acmmmfsll-NIPS2012}
\bibfield{author}{\bibinfo{person}{L. Liu} {and} \bibinfo{person}{T.
  Dietterich}.} \bibinfo{year}{2012}\natexlab{}.
\newblock \showarticletitle{A conditional multinomial mixture model for
  superset label learning}. In \bibinfo{booktitle}{\emph{Advances in Neural
  Information Processing Systems}}. \bibinfo{pages}{548--556}.
\newblock


\bibitem[\protect\citeauthoryear{Liu and Dietterich}{Liu and
  Dietterich}{2014}]%
        {Liu:ICML2014}
\bibfield{author}{\bibinfo{person}{L. Liu} {and} \bibinfo{person}{T.
  Dietterich}.} \bibinfo{year}{2014}\natexlab{}.
\newblock \showarticletitle{Learnability of the superset label learning
  problem}. In \bibinfo{booktitle}{\emph{International Conference on Machine
  Learning}}. \bibinfo{pages}{1629--1637}.
\newblock


\bibitem[\protect\citeauthoryear{Luo and Orabona}{Luo and Orabona}{2010}]%
        {Luo:lfcls-NIPS2010}
\bibfield{author}{\bibinfo{person}{J. Luo} {and} \bibinfo{person}{F. Orabona}.}
  \bibinfo{year}{2010}\natexlab{}.
\newblock \showarticletitle{Learning from candidate labeling sets}. In
  \bibinfo{booktitle}{\emph{Advances in Neural Information Processing
  Systems}}. \bibinfo{pages}{1504--1512}.
\newblock


\bibitem[\protect\citeauthoryear{Nguyen and Caruana}{Nguyen and
  Caruana}{2008}]%
        {Nguyen:KDDM2008}
\bibfield{author}{\bibinfo{person}{N. Nguyen} {and} \bibinfo{person}{R.
  Caruana}.} \bibinfo{year}{2008}\natexlab{}.
\newblock \showarticletitle{Classification with partial labels}. In
  \bibinfo{booktitle}{\emph{ACM SIGKDD International Conference on Knowledge
  Discovery and Data Mining}}. \bibinfo{pages}{551--559}.
\newblock


\bibitem[\protect\citeauthoryear{Oukhellou, Denux, and Aknin}{Oukhellou
  et~al\mbox{.}}{2009}]%
        {Oukhellou2009Learning}
\bibfield{author}{\bibinfo{person}{L. Oukhellou}, \bibinfo{person}{T. Denux},
  {and} \bibinfo{person}{P. Aknin}.} \bibinfo{year}{2009}\natexlab{}.
\newblock \showarticletitle{Learning from partially supervised data using
  mixture models and belief functions}.
\newblock \bibinfo{journal}{\emph{Pattern Recognition}} (\bibinfo{year}{2009}),
  \bibinfo{pages}{334--348}.
\newblock


\bibitem[\protect\citeauthoryear{Panis and Lanitis}{Panis and Lanitis}{2014}]%
        {Panis:FG-NET-JAH2015}
\bibfield{author}{\bibinfo{person}{G. Panis} {and} \bibinfo{person}{A.
  Lanitis}.} \bibinfo{year}{2014}\natexlab{}.
\newblock \showarticletitle{An overview of research activities in facial age
  estimation using the FG-NET aging database}. In
  \bibinfo{booktitle}{\emph{European Conference on Computer Vision}}.
  \bibinfo{pages}{737--750}.
\newblock


\bibitem[\protect\citeauthoryear{Tang and Zhang}{Tang and Zhang}{2017}]%
        {tang:AAAI2017}
\bibfield{author}{\bibinfo{person}{C. Tang} {and} \bibinfo{person}{M. Zhang}.}
  \bibinfo{year}{2017}\natexlab{}.
\newblock \showarticletitle{Confidence-Rated Discriminative Partial Label
  Learning}. In \bibinfo{booktitle}{\emph{AAAI Conference on Artificial
  Intelligence}}. \bibinfo{pages}{2611--2617}.
\newblock


\bibitem[\protect\citeauthoryear{Wang and Zhang}{Wang and Zhang}{2019}]%
        {wang2019adaptive}
\bibfield{author}{\bibinfo{person}{D. Wang} {and} \bibinfo{person}{M. Zhang}.}
  \bibinfo{year}{2019}\natexlab{}.
\newblock \showarticletitle{Adaptive graph guided disambiguation for partial
  label learning}. In \bibinfo{booktitle}{\emph{ACM SIGKDD International
  Conference on Knowledge Discovery and Data Mining}}. \bibinfo{pages}{in
  press}.
\newblock


\bibitem[\protect\citeauthoryear{Wang and Zhang}{Wang and Zhang}{2018}]%
        {wang2018towards}
\bibfield{author}{\bibinfo{person}{J. Wang} {and} \bibinfo{person}{M. Zhang}.}
  \bibinfo{year}{2018}\natexlab{}.
\newblock \showarticletitle{Towards mitigating the class-imbalance problem for
  partial label learning}. In \bibinfo{booktitle}{\emph{ACM SIGKDD
  International Conference on Knowledge Discovery and Data Mining}}.
  \bibinfo{pages}{2427--2436}.
\newblock


\bibitem[\protect\citeauthoryear{Wang, Li, and Zhou}{Wang
  et~al\mbox{.}}{2019}]%
        {Wang:IJCAI2019}
\bibfield{author}{\bibinfo{person}{Q. Wang}, \bibinfo{person}{Y. Li}, {and}
  \bibinfo{person}{Z. Zhou}.} \bibinfo{year}{2019}\natexlab{}.
\newblock \showarticletitle{Partial Label Learning with Unlabeled Data}. In
  \bibinfo{booktitle}{\emph{International Joint Conference on Artificial
  Intelligence}}. \bibinfo{pages}{in press}.
\newblock


\bibitem[\protect\citeauthoryear{Wu and Zhang}{Wu and Zhang}{2019}]%
        {wu2019disambiguation}
\bibfield{author}{\bibinfo{person}{J. Wu} {and} \bibinfo{person}{M. Zhang}.}
  \bibinfo{year}{2019}\natexlab{}.
\newblock \showarticletitle{Disambiguation enabled linear discriminant analysis
  for partial label dimensionality reduction}. In \bibinfo{booktitle}{\emph{ACM
  SIGKDD International Conference on Knowledge Discovery and Data Mining}}.
  \bibinfo{pages}{in press}.
\newblock


\bibitem[\protect\citeauthoryear{Xu, Jin, and Zhou}{Xu et~al\mbox{.}}{2013}]%
        {xu2013speedup}
\bibfield{author}{\bibinfo{person}{M. Xu}, \bibinfo{person}{R. Jin}, {and}
  \bibinfo{person}{Z. Zhou}.} \bibinfo{year}{2013}\natexlab{}.
\newblock \showarticletitle{Speedup matrix completion with side information:
  Application to multi-label learning}. In \bibinfo{booktitle}{\emph{Advances
  in Neural Information Processing Systems}}. \bibinfo{pages}{2301--2309}.
\newblock


\bibitem[\protect\citeauthoryear{Xu, Lv, and Geng}{Xu et~al\mbox{.}}{2019}]%
        {xu2019partial}
\bibfield{author}{\bibinfo{person}{N. Xu}, \bibinfo{person}{J. Lv}, {and}
  \bibinfo{person}{X. Geng}.} \bibinfo{year}{2019}\natexlab{}.
\newblock \showarticletitle{Partial Label Learning via Label Enhancement}. In
  \bibinfo{booktitle}{\emph{AAAI Conference on Artificial Intelligence}}.
  \bibinfo{pages}{in press}.
\newblock


\bibitem[\protect\citeauthoryear{Xuan and Zhang}{Xuan and Zhang}{2018}]%
        {TEBDfPLL-IJCAI2018}
\bibfield{author}{\bibinfo{person}{W. Xuan} {and} \bibinfo{person}{M. Zhang}.}
  \bibinfo{year}{2018}\natexlab{}.
\newblock \showarticletitle{Towards Enabling Binary Decomposition for Partial
  Label Learning}. In \bibinfo{booktitle}{\emph{International Joint Conference
  on Artificial Intelligence}}. \bibinfo{pages}{2427--2434}.
\newblock


\bibitem[\protect\citeauthoryear{Yu and Zhang}{Yu and Zhang}{2015}]%
        {Yu2015ACML}
\bibfield{author}{\bibinfo{person}{F. Yu} {and} \bibinfo{person}{M. Zhang}.}
  \bibinfo{year}{2015}\natexlab{}.
\newblock \showarticletitle{Maximum margin partial label learning}. In
  \bibinfo{booktitle}{\emph{Asian Conference on Machine Learning}}.
  \bibinfo{pages}{96--111}.
\newblock


\bibitem[\protect\citeauthoryear{Yu and Zhang}{Yu and Zhang}{2017}]%
        {Yu:ML2015}
\bibfield{author}{\bibinfo{person}{F. Yu} {and} \bibinfo{person}{M. Zhang}.}
  \bibinfo{year}{2017}\natexlab{}.
\newblock \showarticletitle{Maximum margin partial label learning}.
\newblock \bibinfo{journal}{\emph{Machine Learning}} (\bibinfo{year}{2017}),
  \bibinfo{pages}{573--593}.
\newblock


\bibitem[\protect\citeauthoryear{Yu, Jain, Kar, and Dhillon}{Yu
  et~al\mbox{.}}{2014}]%
        {yu2014large}
\bibfield{author}{\bibinfo{person}{H. Yu}, \bibinfo{person}{P. Jain},
  \bibinfo{person}{P. Kar}, {and} \bibinfo{person}{I. Dhillon}.}
  \bibinfo{year}{2014}\natexlab{}.
\newblock \showarticletitle{Large-scale multi-label learning with missing
  labels}. In \bibinfo{booktitle}{\emph{International Conference on Machine
  Learning}}. \bibinfo{pages}{593--601}.
\newblock


\bibitem[\protect\citeauthoryear{Zeng, Xiao, Jia, Chan, Gao, Xu, and Ma}{Zeng
  et~al\mbox{.}}{2013}]%
        {Zeng:lbaali-CVPR2013}
\bibfield{author}{\bibinfo{person}{Z. Zeng}, \bibinfo{person}{S. Xiao},
  \bibinfo{person}{K. Jia}, \bibinfo{person}{T. Chan}, \bibinfo{person}{S.
  Gao}, \bibinfo{person}{D. Xu}, {and} \bibinfo{person}{Y. Ma}.}
  \bibinfo{year}{2013}\natexlab{}.
\newblock \showarticletitle{Learning by associating ambiguously labeled
  images}. In \bibinfo{booktitle}{\emph{IEEE Conference on Computer Vision and
  Pattern Recognition}}. \bibinfo{pages}{708--715}.
\newblock


\bibitem[\protect\citeauthoryear{Zhang and Yu}{Zhang and Yu}{2015}]%
        {Zhang:IJCAI2015}
\bibfield{author}{\bibinfo{person}{M. Zhang} {and} \bibinfo{person}{F. Yu}.}
  \bibinfo{year}{2015}\natexlab{}.
\newblock \showarticletitle{Solving the partial label learning problem: an
  instance-based approach}. In \bibinfo{booktitle}{\emph{International Joint
  Conference on Artificial Intelligence}}. \bibinfo{pages}{4048--4054}.
\newblock


\bibitem[\protect\citeauthoryear{Zhang, Yu, and Tang}{Zhang
  et~al\mbox{.}}{2017}]%
        {zhang:IEEET2017}
\bibfield{author}{\bibinfo{person}{M. Zhang}, \bibinfo{person}{F. Yu}, {and}
  \bibinfo{person}{C. Tang}.} \bibinfo{year}{2017}\natexlab{}.
\newblock \showarticletitle{Disambiguation-free partial label learning}.
\newblock \bibinfo{journal}{\emph{IEEE Transactions on Knowledge and Data
  Engineering}} (\bibinfo{year}{2017}), \bibinfo{pages}{2155--2167}.
\newblock


\bibitem[\protect\citeauthoryear{Zhang, Zhou, and Liu}{Zhang
  et~al\mbox{.}}{2016}]%
        {zhang2016partial}
\bibfield{author}{\bibinfo{person}{M. Zhang}, \bibinfo{person}{B. Zhou}, {and}
  \bibinfo{person}{X. Liu}.} \bibinfo{year}{2016}\natexlab{}.
\newblock \showarticletitle{Partial label learning via feature-aware
  disambiguation}. In \bibinfo{booktitle}{\emph{ACM SIGKDD International
  Conference on Knowledge Discovery and Data Mining}}.
  \bibinfo{pages}{1335--1344}.
\newblock


\bibitem[\protect\citeauthoryear{Zhou and Gu}{Zhou and Gu}{2017}]%
        {Yu2017Geometric}
\bibfield{author}{\bibinfo{person}{Y. Zhou} {and} \bibinfo{person}{H. Gu}.}
  \bibinfo{year}{2017}\natexlab{}.
\newblock \showarticletitle{Geometric Mean Metric Learning for Partial Label
  Data}.
\newblock \bibinfo{journal}{\emph{Neurocomputing}} (\bibinfo{year}{2017}),
  \bibinfo{pages}{394--402}.
\newblock


\bibitem[\protect\citeauthoryear{Zhou, He, and Gu}{Zhou et~al\mbox{.}}{2016}]%
        {Zhou2016Partial}
\bibfield{author}{\bibinfo{person}{Y. Zhou}, \bibinfo{person}{J. He}, {and}
  \bibinfo{person}{H. Gu}.} \bibinfo{year}{2016}\natexlab{}.
\newblock \showarticletitle{Partial Label Learning via Gaussian Processes}.
\newblock \bibinfo{journal}{\emph{IEEE Transactions on Cybernetics}}
  (\bibinfo{year}{2016}), \bibinfo{pages}{4443--4450}.
\newblock


\bibitem[\protect\citeauthoryear{Zhu, Yan, and Ma}{Zhu et~al\mbox{.}}{2010}]%
        {zhu2010image}
\bibfield{author}{\bibinfo{person}{G. Zhu}, \bibinfo{person}{S. Yan}, {and}
  \bibinfo{person}{Y. Ma}.} \bibinfo{year}{2010}\natexlab{}.
\newblock \showarticletitle{Image tag refinement towards low-rank, content-tag
  prior and error sparsity}. In \bibinfo{booktitle}{\emph{ACM International
  Conference on Multimedia}}. \bibinfo{pages}{461--470}.
\newblock


\end{thebibliography}

\end{document}